\documentclass[10pt,twocolumn,letterpaper]{article}

\usepackage{iccv}
\usepackage{times}
\usepackage{epsfig}
\usepackage{graphicx}
\usepackage{amsmath}
\usepackage{amssymb}

\usepackage{booktabs}
\usepackage{multirow}
\usepackage{xspace}
\usepackage{pifont}
\usepackage{xcolor}
\usepackage[accsupp]{axessibility}  %

\usepackage[utf8]{inputenc} %
\usepackage[T1]{fontenc}    %

\usepackage{algorithm}
\usepackage{amsfonts}       %
\usepackage{amsmath}
\usepackage{amssymb}
\usepackage{amsthm}
\usepackage{array}
\usepackage{bm}
\usepackage{booktabs}       %
\usepackage{caption}
\usepackage{colortbl}
\usepackage{empheq}
\usepackage{enumitem}
\usepackage{etoolbox}
\usepackage{float}
\usepackage{makecell}
\usepackage{mathrsfs}
\usepackage{mdframed}
\usepackage{microtype}      %
\usepackage{multirow}
\usepackage{nccmath}
\usepackage{nicefrac}       %
\usepackage[noend]{algorithmic}
\usepackage{pifont}
\usepackage{ragged2e}
\usepackage{rotating}
\usepackage{subcaption}
\usepackage{tabularx}
\usepackage{tikz}
\usepackage{times}
\usepackage{url}            %
\usepackage{wrapfig}
\usepackage{xcolor, color, colortbl}
\usepackage{xspace}
\usepackage{thm-restate}

\newcommand{\citep}{\cite}
\newcommand{\citet}{\cite}

\newcounter{rowcntr}[table]
\renewcommand{\therowcntr}{\arabic{chapter}.\the\numexpr\arabic{table}+1.\arabic{rowcntr}}

\AtBeginEnvironment{tabular}{\setcounter{rowcntr}{0}}
\newcolumntype{H}{>{\setbox0=\hbox\bgroup}c<{\egroup}@{}}

\newcommand{\R}{{\mathbb{R}}}

\newcommand{\comment}[1]{}

\def\pp{{\boldsymbol p}}

\def\vx{{\bm{x}}}

\newcommand{\RRC}{\mbox{\emph{RRC}}}
\newcommand{\RRCMixing}{\mbox{\emph{RRC$+$Mixing}}}
\newcommand{\RRCRARE}{\mbox{\emph{RRC$+$RA/RE}}}
\newcommand{\RRCMsRs}{\mbox{\emph{RRC$+$M$^\ast$ $+$R$^\ast$}}}
\newcommand{\Mixing}{\mbox{\emph{Mixing}}}
\newcommand{\RARE}{\mbox{\emph{RA/RE}}}
\newcommand{\MsRs}{\mbox{\emph{M$^\ast$$+$R$^\ast$}}}

\definecolor{colorYes}{RGB}{51,160,44}
\definecolor{colorNo}{RGB}{228,26,28} %

\newcommand{\cmark}{\textcolor{colorYes}{\ding{51}}}%
\newcommand{\xmark}{\textcolor{colorNo}{\ding{55}}}%

\newcommand{\IN}{ImageNet}
\newcommand{\CIFAR}{CIFAR-100}
\newcommand{\Flowers}{Flowers-102}
\newcommand{\Food}{Food-101}
\newcommand{\INp}{ImageNet$^+$}
\newcommand{\CIFARp}{CIFAR-100$^+$}
\newcommand{\Flowersp}{Flowers-102$^+$}
\newcommand{\Foodp}{Food-101$^+$}

\usepackage[pagebackref=true,breaklinks=true,letterpaper=true,colorlinks,bookmarks=false]{hyperref}

\usepackage[capitalize]{cleveref}
\crefname{section}{Sec.}{Secs.}
\Crefname{section}{Section}{Sections}
\Crefname{table}{Table}{Tables}
\crefname{table}{Tab.}{Tabs.}

\iccvfinalcopy %

\def\iccvPaperID{10289} %

\ificcvfinal\fi

\begin{document}

\title{Reinforce Data, Multiply Impact: Improved Model Accuracy and Robustness 
with Dataset Reinforcement}

\author{{Fartash Faghri\thanks{Correspondence to 
\href{mailto:fartash@apple.com}{fartash@apple.com}.}} ,
Hadi Pouransari, Sachin Mehta, Mehrdad Farajtabar,\\
Ali Farhadi, Mohammad Rastegari, Oncel Tuzel\\
Apple\\
}
\maketitle
\ificcvfinal\thispagestyle{empty}\fi

\begin{abstract}

We propose \emph{Dataset Reinforcement}, a strategy to improve a dataset once 
such that the accuracy of any model architecture trained on the reinforced 
dataset is improved at no additional training cost for users.
We propose a Dataset Reinforcement strategy based on data augmentation and 
knowledge distillation.
Our generic strategy is designed based on extensive analysis across CNN- and 
transformer-based models and performing large-scale study of distillation with 
state-of-the-art models with various data augmentations.
We create a reinforced version of the \IN{} training dataset, called 
\INp{}, as well as reinforced datasets \CIFARp{}, \Flowersp{}, and 
\Foodp{}.
Models trained with \INp{} are more accurate, robust, and calibrated, and 
transfer well to downstream tasks (e.g., segmentation and detection).
As an example, the accuracy of ResNet-50 improves by 1.7\% on the \IN{} 
validation set, 3.5\% on \IN{}V2, and 10.0\% on \IN{}-R.
Expected Calibration Error (ECE) on the \IN{} validation set is also reduced by 9.9\%.
Using this backbone with Mask-RCNN for 
object detection on MS-COCO, the mean average precision improves by 0.8\%. We 
reach similar gains for MobileNets, ViTs, and Swin-Transformers. For 
MobileNetV3 and Swin-Tiny, we observe significant improvements on \IN{}-R/A/C 
of \textbf{up to 20\% improved robustness}.
Models pretrained on \INp{} and fine-tuned on \CIFARp{}, \Flowersp{}, and 
\Foodp{}, reach up to 3.4\% improved accuracy.
The code, datasets, and pretrained models are available at
\href{https://github.com/apple/ml-dr}{https://github.com/apple/ml-dr}.

\end{abstract}

\section{Introduction}\label{sec:intro}
\begin{figure}[t]
    \centering
    \includegraphics[width=0.95\linewidth]{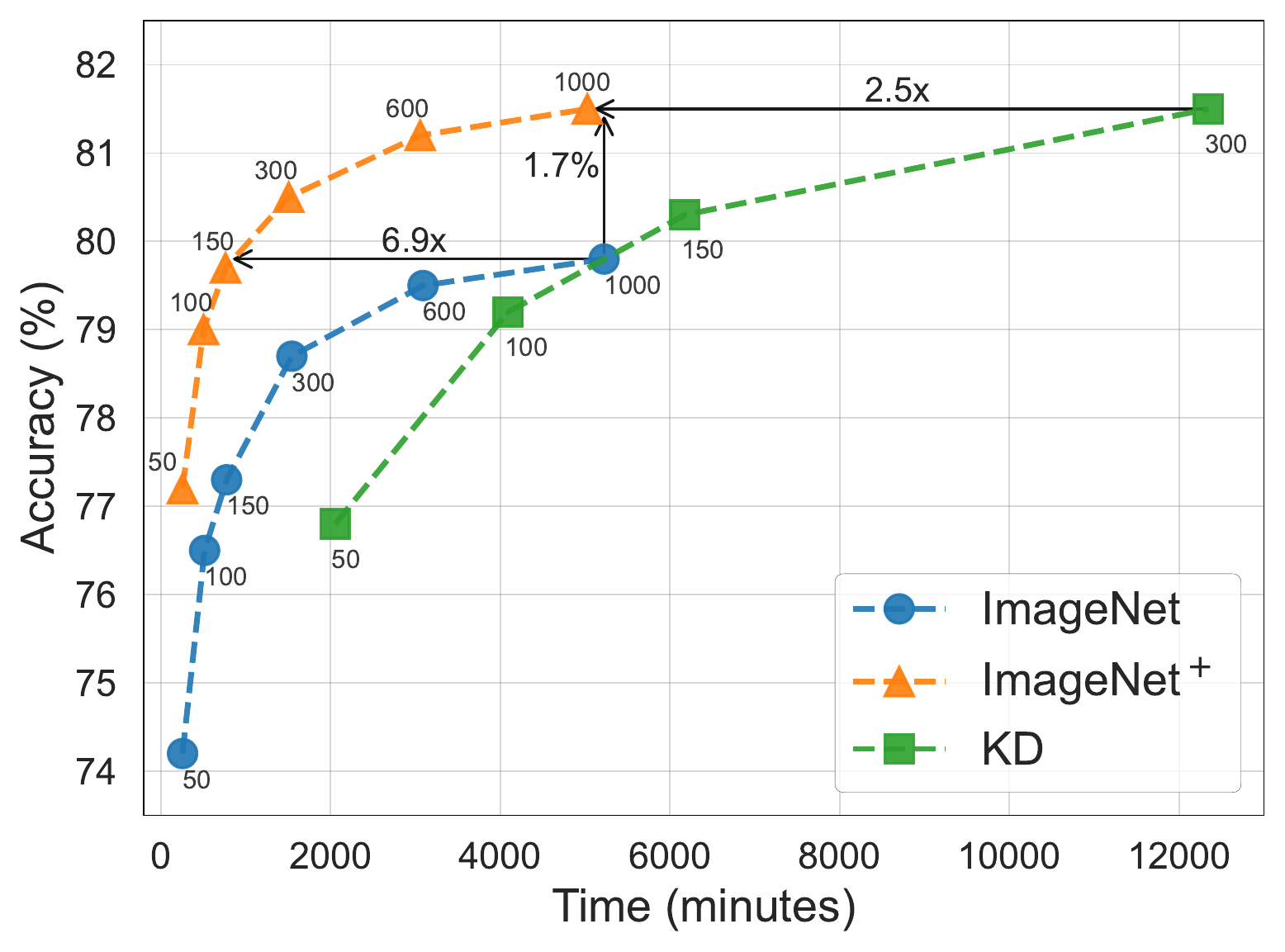}
       \vspace{-4mm}
    \caption{\textbf{Reinforced \IN{}, \INp{}, improves accuracy at 
    similar iterations/wall-clock.}  \IN{} validation accuracy of ResNet-50 
    is shown as a function of training duration with (1) \IN{} dataset, (2) 
    knowledge distillation (KD), and (3) \INp{} dataset (ours).  Each point 
    is a full training with epochs varying from 50-1000.
    An epoch has the same number of iterations for \IN{}/\INp{}.
    }
    \label{fig:wall_clock}
    \vspace{-4mm}
\end{figure}
\begin{table}[t]
    \centering
    \resizebox{1.0\columnwidth}{!}{
        \begin{tabular}{lcccccc}
            \toprule[1.5pt]
            \multirow{2}{*}{\textbf{Model}}
            & \textbf{+Data}
            & \textbf{+Reinforced}
            & \multirow{2}{*}{\textbf{\IN{}}} 
            & \multirow{2}{*}{\textbf{\CIFAR{}}} 
            & \multirow{2}{*}{\textbf{\Flowers{}}} 
            & \multirow{2}{*}{\textbf{\Food{}}} \\
            & \textbf{Augmentation}
            & \textbf{Dataset(s)}\\
             \midrule[1.25pt]
             \multirow{2}{*}{MobileNetV3-Large}
             & \xmark & \xmark & 75.8 & 84.4 & 92.5 & 86.1\\
             & \xmark & \cmark & \textbf{77.9} & \textbf{87.5} & \textbf{95.3} & \textbf{89.5}\\
             \midrule
             \multirow{4}{*}{ResNet-50}
             & RandAugment & \xmark & 80.4 & 88.4 & 93.6 & 90.0\\
             & AutoAugment & \xmark & 80.2 & 87.9 & 95.1 & 89.0\\
             & TrivialAugWide & \xmark & 80.4 & 87.9 & 94.8 & 89.3\\
             & \xmark & \cmark & \textbf{82.0} & \textbf{89.8} & \textbf{96.3} & \textbf{92.1}\\
             \midrule
             \multirow{2}{*}{SwinTransformer-Tiny}
             & RandAugment & \xmark & 81.3 & 90.7 & 96.3 & 92.3 \\
             & \xmark & \cmark & \textbf{84.0} & \textbf{91.2} & \textbf{97.0} & \textbf{92.9}\\
            \bottomrule[1.5pt]
        \end{tabular}
    }
    \caption{\textbf{Training/fine-tuning on reinforced datasets improve 
    accuracy for a variety of architectures.} We reinforce each dataset 
    \emph{once} and train multiple models with similar cost as training on the 
    original dataset. %
    For datasets other than \IN{}, we fine-tune \IN{}/\INp{} 
    pretrained models.
    Dataset reinforcement significantly benefits from efficiently reusing the 
    knowledge of a teacher.
         }
    \label{tab:teaser_results}
    \vspace{-4mm}
\end{table}

\begin{figure*}[t]
    \centering
    \includegraphics[width=1.0\textwidth]{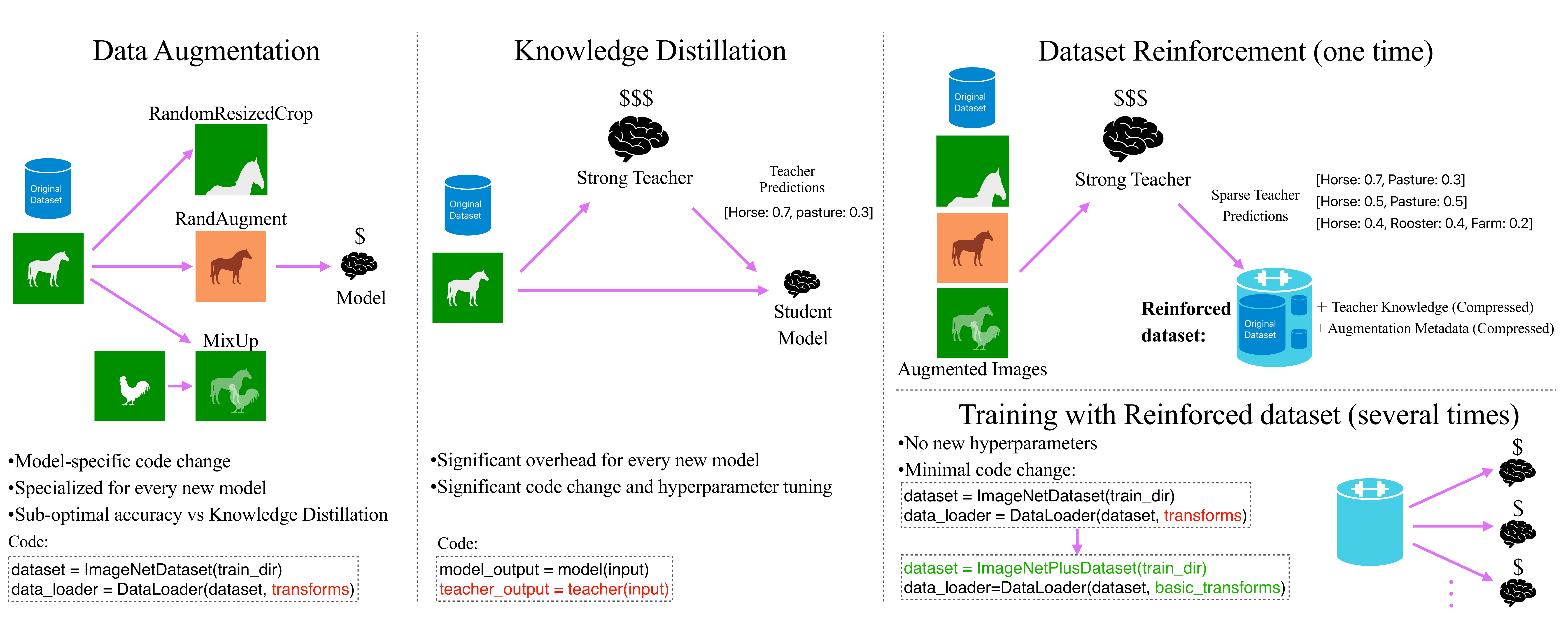}
       \vspace{-8mm}
    \caption{\textbf{Illustration of Dataset Reinforcement.}
    Data augmentation and knowledge distillation are common approaches to 
    improving accuracy.
    Dataset reinforcement combines the benefits of both by bringing
    the advantages of large models trained on large datasets to other datasets 
    and models.
    Training of new models with a reinforced dataset is as fast as 
    training on the original dataset for the same total iterations.
    Creating a reinforced dataset is a one-time process (e.g., \IN{} to \INp{})
    the cost of which is amortized over repeated uses.
    }\label{fig:illustration_wide}
    \vspace{-4mm}
\end{figure*}

With the advent of the CLIP~\citep{radford2021learning}, the machine 
learning community got increasingly interested in massive datasets whereby 
the models are trained on hundreds of millions of samples,
which is orders of magnitude larger than the conventional 
\IN{}~\citep{deng2009imagenet} with 1.2M samples.
At the same time, models have gradually grown larger in multiple domains~\citep{alabdulmohsin2022revisiting}.
In computer vision, the state-of-the-art models
have upwards of $300$M parameters according to the 
{Timm}~\citep{rw2019timm} library
(e.g., {BEiT}~\citep{beit}, {DeiT III}~\citep{Touvron2022DeiTIR}, 
{ConvNeXt}~\citep{liu2022convnet}) and process inputs at up to $800\times 800$ 
resolution (e.g., {EfficientNet-L2-NS}~\citep{xie2020self}).
Recent multi-modal vision-language models have up to 1.9B parameters (e.g., 
{BeiT-3}~\citep{wang2022image}).

On the other side, there is a significant demand for small models that satisfy 
stringent hardware requirements.
Additionally, there are plenty of tasks with small datasets that are 
challenging to scale because of the high cost associated with  collecting and annotating new data.
We seek to bridge this gap and bring the benefits of large models to any large, 
medium, or small dataset.
We use knowledge from large 
models~\citep{radford2021learning,dosovitskiy2020image,bommasani2021opportunities} 
to enhance the training of new models.

In this paper, we introduce {\it Dataset Reinforcement (DR)\/} as a strategy 
that improves the accuracy of models through reinforcing the training dataset.  
Compared to the original training data, a method for dataset reinforcement 
should satisfy the following desiderata:
\begin{itemize}[leftmargin=*]
\itemsep0em
\item {\bf No overhead for users}: Minimal increase in the computational cost 
    of training a new model for similar total iterations (e.g., similar 
    wall-clock time and CPU/GPU utilization).
\item {\bf Minimal changes in user code and model}: Zero or minimal 
    modification to the training code and model architecture for the users of 
    the reinforced dataset (e.g., only the dataset path and the data loader need to 
    change).
\item {\bf Architecture independence}: Improve the test accuracy across variety of model architectures.
\end{itemize}

To understand the importance of the DR desiderata, let us discuss two common 
methods for performance improvements: data augmentation and knowledge 
distillation. Illustration in \cref{fig:illustration_wide} compares these 
methods and our strategy for dataset reinforcement.

Data augmentation is crucial to the improved performance of machine learning 
models. Many state-of-the-art vision models~\citep{he2016deep, 
huang2017densely,howard2019searching} use the standard Inception-style 
augmentation \cite{szegedy2015going} (i.e., random resized crop and random 
horizontal flipping) for training.
In addition to these standard augmentation methods, recent models~\citep{touvron2021training,liu2021swin} also incorporate mixing augmentations (e.g., MixUp~\citep{zhang2017mixup} and 
CutMix~\citep{yun2019cutmix}) and automatic augmentation methods (e.g., RandAugment~\citep{cubuk2020randaugment} and
AutoAugment~\citep{cubuk2018autoaugment}) to generate new data. However, data 
augmentation fails to satisfy all the desiderata as it does not provide 
architecture independent generalization. For example, light-weight CNNs perform 
best with standard Inception-style augmentations~\citep{howard2019searching} 
while vision transformers~\citep{touvron2021training,liu2021swin} prefer 
a combination of standard as well as advanced augmentation methods.

Knowledge distillation (KD) refers to the training of a student model by 
matching the output of a teacher model~\citep{hinton2015distilling}. KD has 
consistently been shown to improve the accuracy of new models independent of 
their architecture significantly more than data 
augmentations~\citep{touvron2021training}.
However, knowledge distillation is expensive as it requires performing the 
inference (forward-pass) of an often significantly large teacher model at every 
training iteration.
KD also requires modifying the training code to perform two forward passes on 
both the teacher and the student. As such, KD fails to satisfy minimal overhead 
and code change desiderata.

This paper proposes a dataset reinforcement strategy that exploits the 
advantages of both knowledge distillation and data augmentation by removing the 
training overhead of KD and finding generalizable data augmentations.
Specifically, we introduce the \emph{\INp{}} dataset that provides 
a balanced trade-off between accuracies on a variety of models and has the same 
wall-clock as training on \IN{} for the same number of iterations 
(\cref{fig:wall_clock,tab:teaser_results}).
To train models using the \INp{} dataset, one only needs to change a few 
lines of the user code to use a modified data loader that reinforces every 
sample loaded from the training set.

\vspace{-3mm}
\paragraph{Summary of contributions:}
\begin{itemize}[leftmargin=*]
        \vspace{-2mm}
    \itemsep0em
    \item We present a comprehensive large scale study of knowledge 
        distillation from 80 pretrained state-of-the-art models and their 
        ensembles.
        We observe that ensembles of state-of-the-art models 
        trained on massive datasets generalize across student architectures 
        (\cref{sec:good_teacher}).
    \item We reinforce \IN{} by efficiently storing the knowledge of a strong 
        teacher on a variety of augmentations. We investigate the generalizability 
        of various augmentations for dataset reinforcement and find a tradeoff 
        controlled by the reinforcement difficulty and model complexity 
        (\cref{sec:aug_difficulty}). This tradeoff can further be alleviated 
        using curriculums based on the reinforcements (\cref{sec:curriculum}).
    \item We introduce \INp{}, a reinforced version of \IN{}, that 
        provides up to 4\% improvement in accuracy for a variety of architectures 
        in short as well as long training.
        We show that \INp{} pretrained models result in 0.6-0.8 improvements in 
        mAP for detection on MS-COCO and 
        0.3-1.3\% improvement in mIoU for segmentation on ADE-20K 
        (\cref{sec:transfer}).
    \item
        Similarly, we create \CIFARp{}, \Flowersp{}, and \Foodp{}, and 
        demonstrate their effectiveness for fine-tuning 
        (\cref{sec:other_datasets}).
        \INp{} pretrained models fine-tuned on \CIFARp{}, \Flowersp{}, and 
        \Foodp{} show up to 3\% improvement in transfer learning on \CIFAR{}, 
        \Flowers{}, and \Food{}.
    \item To further investigate this emergent transferablity we study 
        robustness and calibration of the \INp{} trained models. They reach 
        up to 20\% improvement on a variety of OOD datasets,  \IN{}-(V2, A, 
        R, C, Sketch), and ObjectNet (\cref{sec:robustnes}).  We also show that 
        models trained on \INp{} are well calibrated compared to their 
        non-reinforced alternatives (\cref{sec:calibration}).
\end{itemize}

Our \INp{}, \CIFARp{}, \Flowersp{}, and \Foodp{} reinforcements along with code 
to reinforce new datasets are available at 
\href{https://github.com/apple/ml-dr}{https://github.com/apple/ml-dr}.

\section{Dataset Reinforcement}\label{sec:method}

Our proposed strategy for dataset reinforcement (DR) is efficiently combining 
knowledge distillation and data augmentation to generate an enhanced dataset.
We precompute and store the output of a strong pretrained model on multiple 
augmentations per sample as reinforcements. The stored outputs are more 
informative and useful for training compared with ground truth labels.  This 
approach is related to prior works, such as Fast Knowledge Distillation 
(FKD)~\citep{shen2021fast}) and ReLabel~\citep{yun2021re}, that aim to improve 
the labels.
Beyond these works, our goal is to find generalizable reinforcements that 
improve the accuracy of any architecture.
First we perform a comprehensive study to find a strong teacher 
(\cref{sec:good_teacher})  then find generalizable reinforcements on \IN{} 
(\cref{sec:aug_difficulty}). To demonstrate the generality of our strategy and 
findings, we further reinforce \CIFAR{}, \Flowers{}, and \Food{} 
(\cref{sec:other_datasets}).

The reinforced dataset consists of the original dataset plus the reinforcement 
meta data for all training samples.  During the reinforcement process, for each 
sample a fixed number of reinforcements is generated using parametrized 
augmentation operations and evaluating the teacher predictions. To save 
storage, instead of storing the augmented images, the augmentation parameters 
are stored alongside the sparsified output of the teacher. As a result, the 
extra storage needed is only a fraction of the original training set for large 
datasets.
Using our reinforced dataset has no computational overhead on training, 
requires no code change, and provides improvements for various architectures.

\subsection{What is a good teacher?}\label{sec:good_teacher}

\begin{figure*}[t]
\begin{center}
    \begin{subfigure}[t]{0.28\textwidth}
        \centering
        \includegraphics[width=\textwidth]{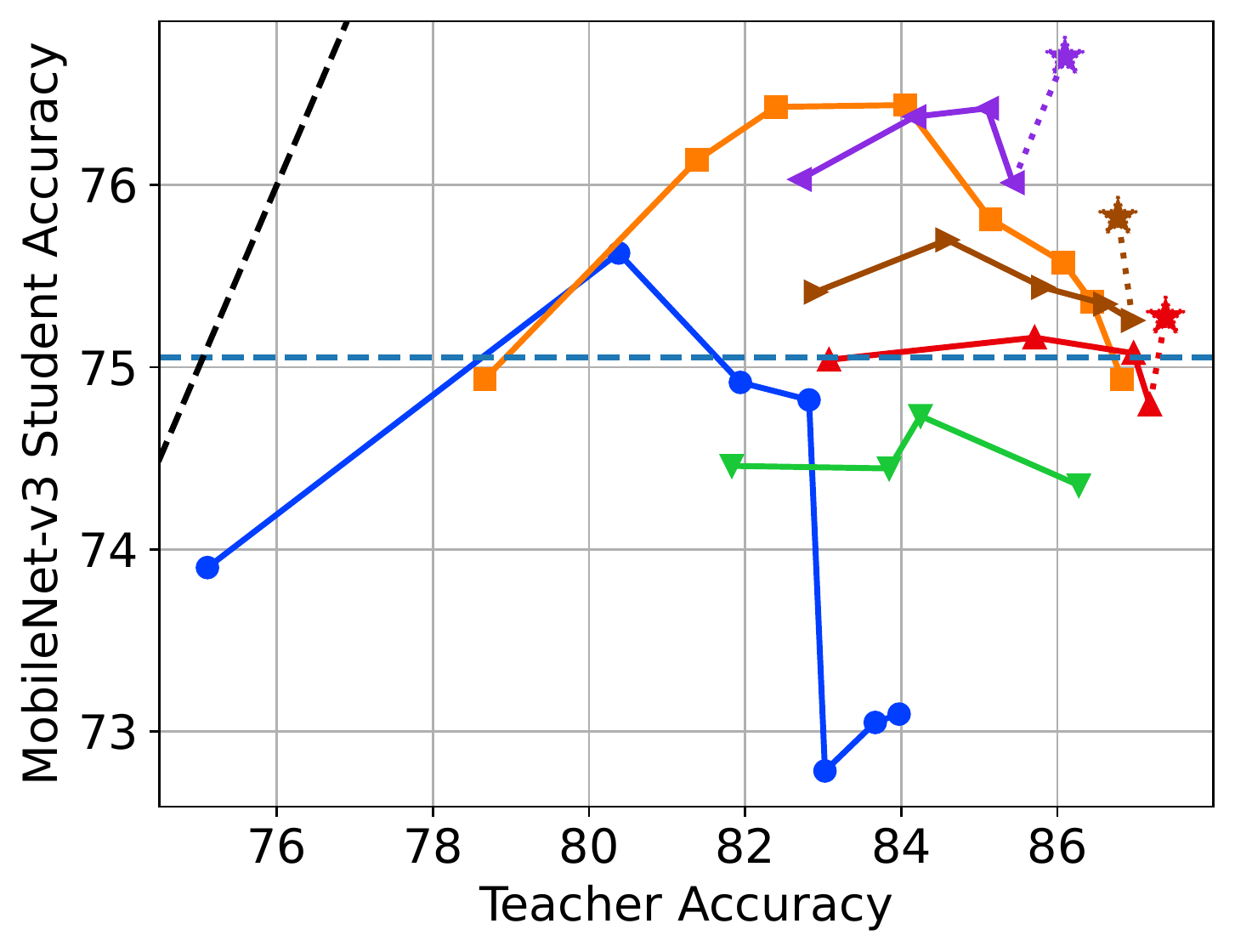}
        \caption{Light-weight CNN 
        (MobileNetV3)}\label{fig:imagenet_MobileNetv3_distill_e300}
    \end{subfigure}
    \hfill
    \begin{subfigure}[t]{0.28\textwidth}
        \centering
        \includegraphics[width=\textwidth]{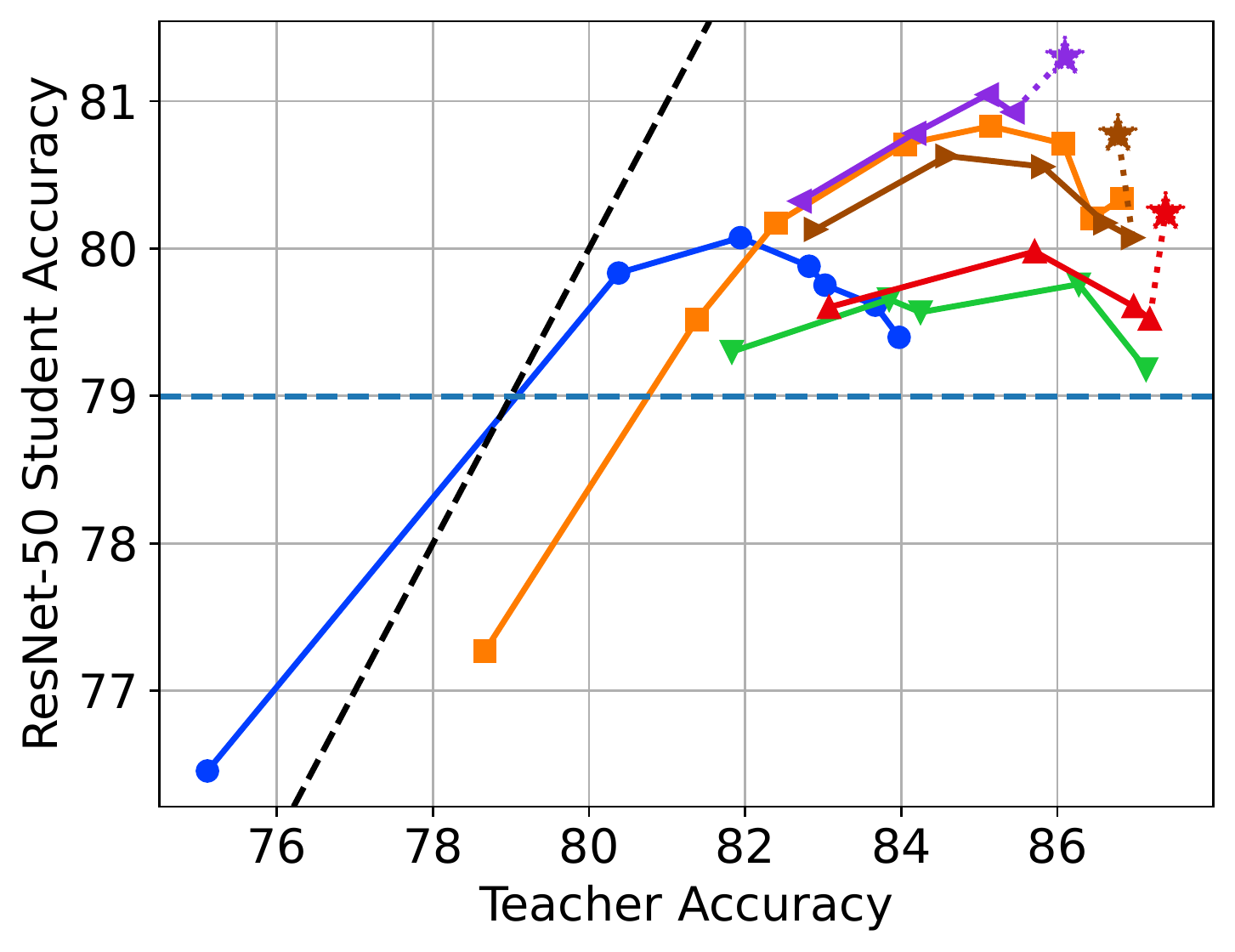}
        \caption{Heavy-weight CNN 
        (ResNet-50)}\label{fig:imagenet_R50_distill_e300}
    \end{subfigure}
    \hfill
    \begin{subfigure}[t]{0.28\textwidth}
        \centering
        \includegraphics[width=\textwidth]{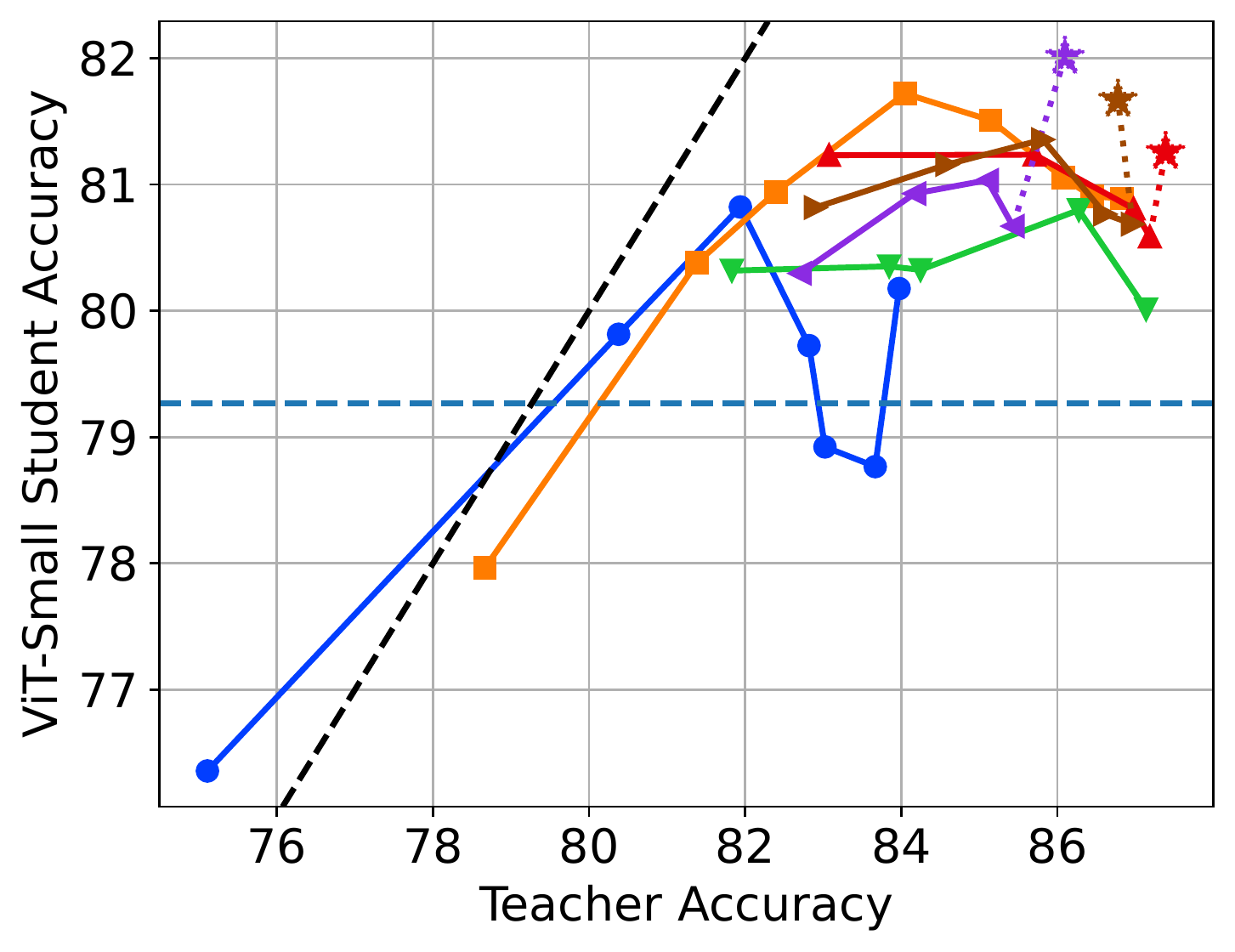}
        \caption{Transformer (ViT-Small) 
        }\label{fig:imagenet_ViTSmall_distill_e300}
    \end{subfigure}
    \hfill
    \begin{minipage}[t]{0.14\linewidth}
        \centering
        \vspace*{-3.5cm}
        \begin{subfigure}[t]{\textwidth}
            \includegraphics[width=\textwidth]{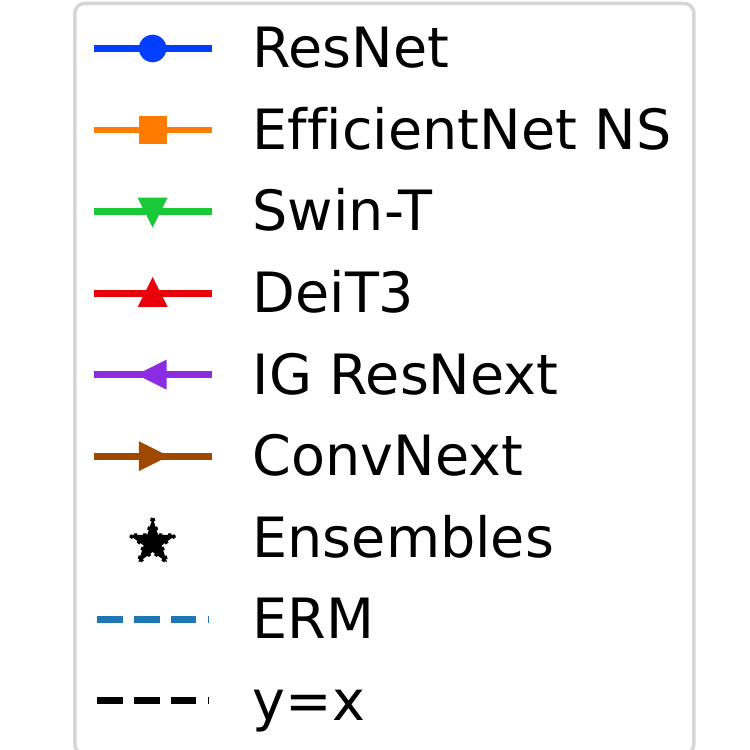}
        \end{subfigure}
    \end{minipage}
\end{center}
    \vspace{-5mm}
    \caption{\textbf{Knowledge Distillation with models and ensembles from Timm 
    library.} We observe the validation accuracy of students
    saturates or drops as the accuracy of teachers within an architecture 
    family increases.
    We also observe that ensembles (marked by asterisks) are better teachers.
    Ensemble of IG-ResNext models performs best as teachers across student 
    architectures.
    ERM (Empirical Risk Minimization) is standard training without knowledge 
    distillation.
    Similar results for 150 epoch training in \cref{fig:timm_distill_e150}.
    }\label{fig:timm_distill_e300}
    \vspace{-4mm}
\end{figure*}

Knowledge distillation (KD) refers to training a student model using the 
outputs of a teacher 
model~\citep{bucilua2006model,ba2014deep,hinton2015distilling}.  The training 
objective is as follows:
\begin{equation}
    \min_{\theta}\, \mathbb{E}_{\vx \sim \mathcal{D}, \hat{\vx} \sim 
    \mathcal{A}(\vx)} \mathcal{L}(f_{\theta}(\hat{\vx}), g(\hat{\vx}))\,,
\end{equation}
where, $\mathcal{D}$ is the training dataset, $\mathcal{A}$ is augmentation 
function, $f_{\theta}$ is the student model parameterized with $\theta$, $g$ is 
the teacher model, and $\mathcal{L}$ is the loss function between student and 
teacher outputs. Throughout this paper, we use the KL loss without 
a temperature hyperparameter and no mixing with the cross-entropy loss.
We teach the student to imitate the output of the teacher on all augmentations 
consistent with \citep{beyer2022knowledge}.

It is common to use a fixed teacher because repeating experiments and selecting 
the best teacher is expensive~\citep{beyer2022knowledge,gou2021knowledge}.
The teacher is often selected based on the state-of-the-art test accuracy of 
available pretrained models.  However, it has been observed that most accurate 
models do not necessarily appear to be the best 
teachers~\citep{cho2019efficacy,mirzadeh2020improved}.  Ensemble models on the 
other hand, have been shown to be promising teachers from the early work of 
\citet{bucilua2006model} until recent works in various 
domains~\citep{chebotar2016distilling,you2017learning,shen2020meal,stanton2021does} 
and with techniques to boost the their 
performance~\citep{shen2019meal,fakoor2020fast,malinin2019ensemble}. None of 
these works have comprehensively studied finding the best teacher along with 
the necessary augmentations that result in consistent improvements over 
multiple student architectures.

To understand what makes a good teacher to reinforce datasets, we perform 
knowledge distillation with a variety of pretrained models in the Timm 
library~\citep{rw2019timm} distilled to three representative student 
architectures MobileNetV3-large~\citep{howard2019searching}, 
ResNet-50~\citep{he2016deep}, and ViT-Small~\citep{dosovitskiy2020image}.
MobileNetV3 represents light-weight CNNs that often prefer easier training.  
ResNet-50 represents heavy-weight CNNs that can benefit from difficult training 
regimes but do not heavily rely on it because of their implicit inductive bias 
of the architecture.
ViT-small represents the transformer architectures that have less implicit bias 
compared with CNNs and learn better in the presence of complex and difficult 
datasets.
We consider various families of models as teachers including ResNets (34--152 
and type d variants)~\citep{he2016deep}, ConvNeXt family pretrained on the 
\IN{}-22K and fine-tuned on \IN{}-1K~\citep{liu2022convnet}, DeiT-3 
pretrained on the \IN{}-21K and fine-tuned on \IN{}-1K, IG-ResNext 
pretrained on the Instagram dataset~\citep{mahajan2018exploring}, EfficientNets 
with Noisy Student training~\citep{xie2020self}, and Swin-TransformersV2 
pretrained with and without \IN{}-22K and fine-tuned on 
\IN{}-1K~\citep{liu2022swin}.
This collection covers a variety of vision transformers and CNNs  pretrained on 
a wide spectrum of dataset sizes.
We train all students with $224\times 224$ inputs and follow 
\citep{beyer2022knowledge} to match the resolution of teachers optimized to 
take larger inputs by passing the large crop to the teacher and resize it to 
$224\times 224$ for the student.

We present the accuracies of students trained for 300 epochs as a function of 
the teacher accuracy in~\cref{fig:timm_distill_e300}.
Focusing first on the single (non-ensemble) networks (marked by circles),  
consistent with prior work, we observe that the most accurate models are not 
usually the best teachers~\citep{mirzadeh2020improved}.
For CNN model families (ResNets, EfficientNets, ResNexts, and ConvNeXts), the 
student accuracy is generally correlated with the teacher accuracy.
When increasing the teacher accuracy, the student first improves but then it 
starts to saturate or even drops with the most accurate member of the family.  
Vision Transformers (Swin-Transformers, and DeiT-3) as teachers do not show the 
same trend as the accuracy of the students flattens across different teachers.  
Recently,\citet{liu2022meta} suggested that temperature tuning can help in KD 
from larger teachers. We do not adopt such hyperparameter tuning strategies in 
favor of architecture-independence and generalizability of dataset 
reinforcement.

On the other side, ensembles of state-of-the-art models (marked by asterisks) 
are consistently better teachers compared with any individual member of the 
family.  We create 4-member ensembles of the best models from IG-ResNexts, 
ConvNeXts, and DeiT3 to cover CNNs, vision transformers, and extra data models.  
We find IG-ResNext teacher to provide a balanced improvement across all 
students. IG-ResNext models are also trained with $224\times 224$ inputs while, 
for example, the best teacher from EfficientNet-NS family, EfficientNet-L2-NS, 
performs best at larger resolutions that is significantly more expensive to 
train with.

One of the benefits of dataset reinforcement paradigm is that the teacher can 
be expensive to train and use as long as we can afford to run it \emph{once} on 
the target dataset for reinforcement. Also, the process of dataset 
reinforcement is highly parallelizable because performing the forward-pass on 
the teacher to generate predictions on multiple augmentations does not depend 
on any state or any optimization trajectory. For these reasons, we also 
considered significantly scaling knowledge distillation to super large 
ensembles with up to 128 members.  We discuss our findings in 
\cref{sec:super_ensembles}.
Full table of accuracies for this section are in 
\cref{sec:timm_distill_appendix}.

\subsection{\INp{}: What is the best combination of 
reinforcements?}\label{sec:aug_difficulty}

\begin{table*}[t!]
    \centering
    \resizebox{1.8\columnwidth}{!}{
        \begin{tabular}{lcccc}
            \toprule[1.5pt]
            &
            \multirow{2}{*}{\bfseries \shortstack[l]{Sparse teacher prob.}} &
            \multirow{2}{*}{\bfseries \shortstack[l]{Random Resize Crop \\
            + Horizontal Flip}} &
            \multirow{2}{*}{\bfseries \shortstack[l]{Random Augment \\
            + Random Erase}} &
            \multirow{2}{*}{\bfseries \shortstack[l]{MixUp + CutMix}} \\
            \\
            \midrule[1.25pt]
            \textbf{\INp{} variant} & All & All & +RA/RE, +M*+R* & +Mixing, M*+R*\\
            \textbf{Apply probability} &
            1 & 1, 0.5 & 1, 0.25 & 0.5, 0.5\\
            \textbf{Parameters} &
            $10\times$ (Index, Prob) &
            $4\times$ Coords + Flip bit &
            $2\times$ (Op Id, Magnitude) +
            $4\times$ Coords &
            (Img Id, $\lambda$) +
            (Img Id, $4\times$ Coords) \\
            \textbf{Storage space (in bytes)} &
            $10\times (2\times 4)$ &
            $4\times 4 + 1$ &
            $2\times 2\times 4 + 4\times 4$ &
            $2\times 4 + (1+4)\times 4$\\
            \textbf{Total storage space ($400$ samples per image)} &
            38 GB & 8  GB & 15  GB & 13  GB\\
            \bottomrule[1.5pt]
        \end{tabular}
    }
    \vspace{-2mm}
    \caption{\textbf{Additional storage in \INp{} variants}.  Total additional 
    storage for \INp{} (\RRCRARE{}) is 61 GBs.}
    \label{tab:imagenet_plus_specs}
\end{table*}

\begin{figure*}[t!]
    \centering
    \begin{subfigure}[b]{2\columnwidth}
        \centering
        \begin{tabular}{cccc}
            \hspace{8mm} \scalebox{0.75}{MobileNetV1}
            & \hspace{8mm} \scalebox{0.75}{MobileNetV2}
            & \hspace{8mm} \scalebox{0.75}{MobileNetV3} 
            & \hspace{3mm} \scalebox{0.85}{\textbf{\INp{} Variants}}  \\[-1mm]
             \raisebox{-0.5\height}{\includegraphics[width=0.24\columnwidth]{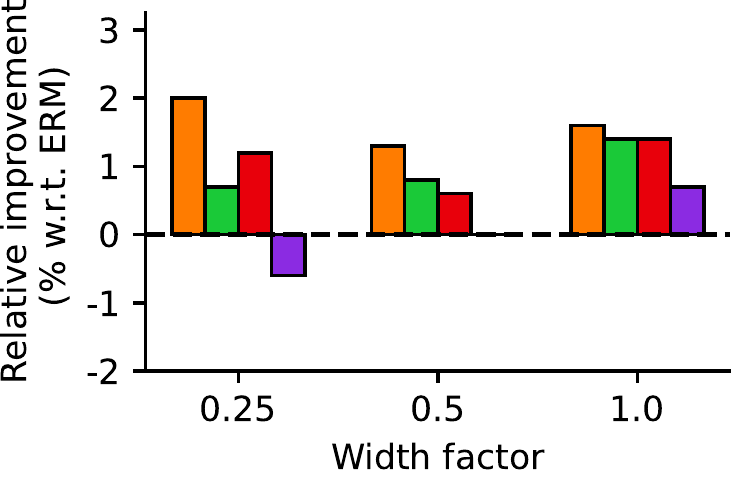}} 
             & \raisebox{-0.5\height}{\includegraphics[width=0.24\columnwidth]{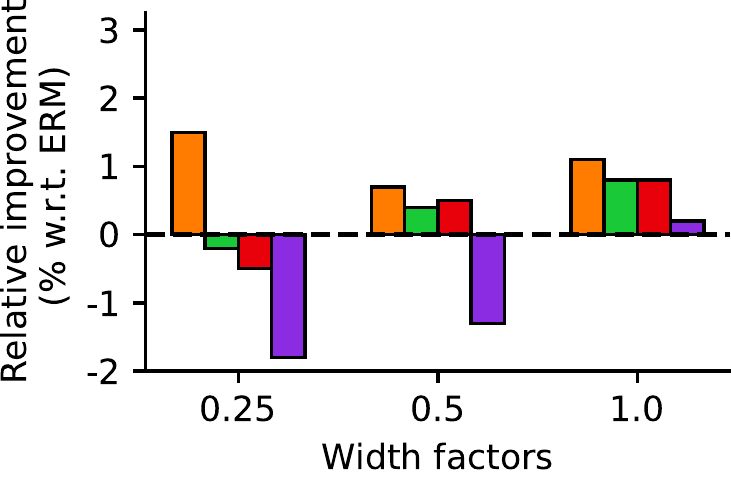}} 
             & \raisebox{-0.5\height}{\includegraphics[width=0.24\columnwidth]{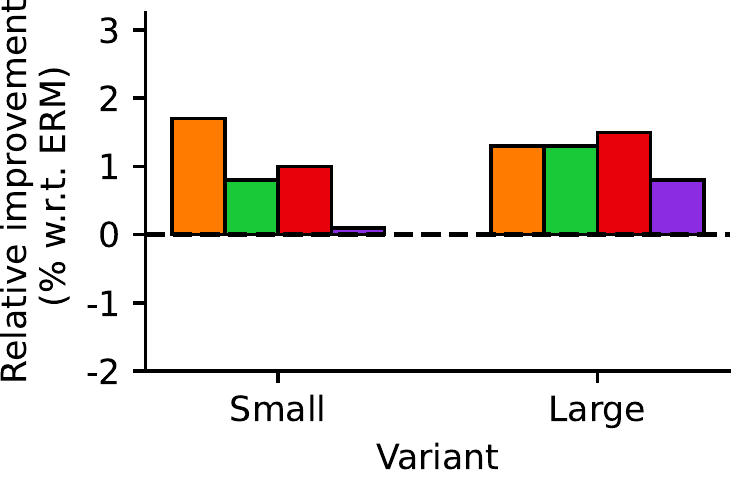}}
             & \hspace{4mm} \raisebox{-0.25\height}{\includegraphics[width=0.16\columnwidth]{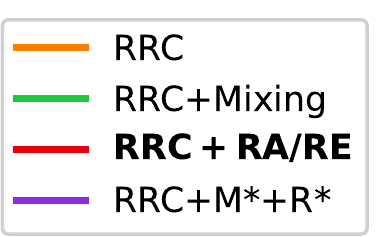}}
        \end{tabular}
        \caption{Light-weight CNNs}
        \label{fig:light_cnn_imagenet_delta}
    \end{subfigure}
    \vfill
    \begin{subfigure}[b]{\columnwidth}
        \centering
        \begin{tabular}{cc}
            \hspace{8mm} \scalebox{0.75}{ResNet} & \hspace{8mm} 
            \scalebox{0.75}{EfficientNet}\\[-1mm]
             \includegraphics[width=0.48\columnwidth]{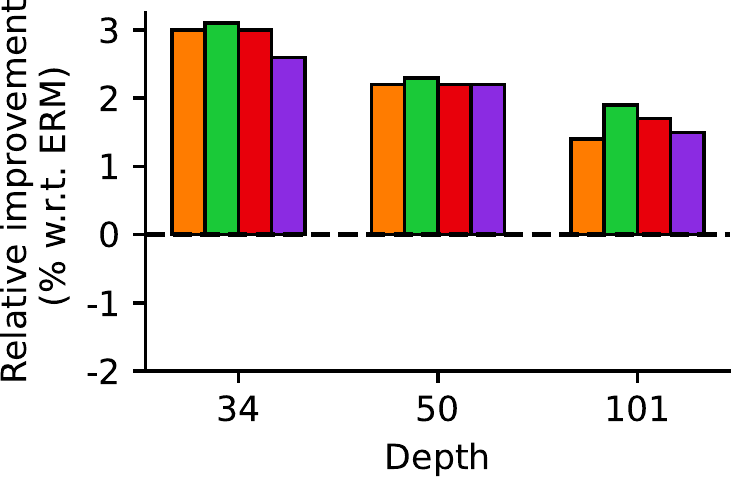} 
             & \includegraphics[width=0.48\columnwidth]{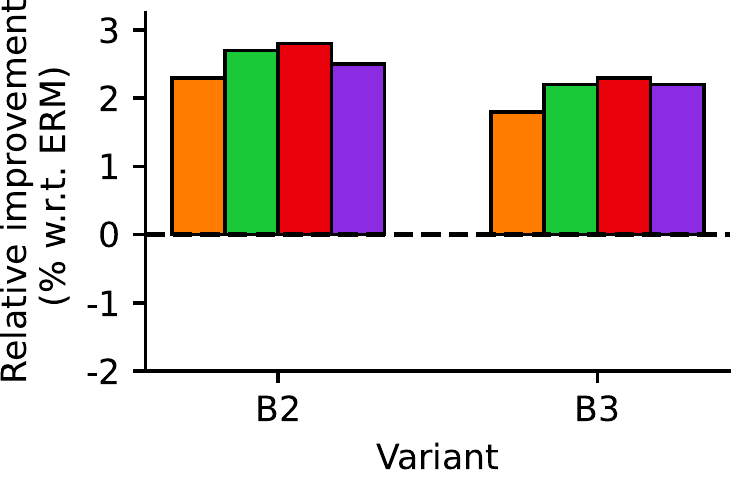}
        \end{tabular}
        \vspace{-2mm}
        \caption{Heavy-weight CNNs}
        \label{fig:heavy_cnn_imagenet_delta}
    \end{subfigure}
    \hfill
    \begin{subfigure}[b]{\columnwidth}
        \centering
        \begin{tabular}{cc}
            \hspace{8mm} \scalebox{0.75}{ViT} & \hspace{8mm} 
            \scalebox{0.75}{SwinTransformer}\\[-1mm]
             \includegraphics[width=0.48\columnwidth]{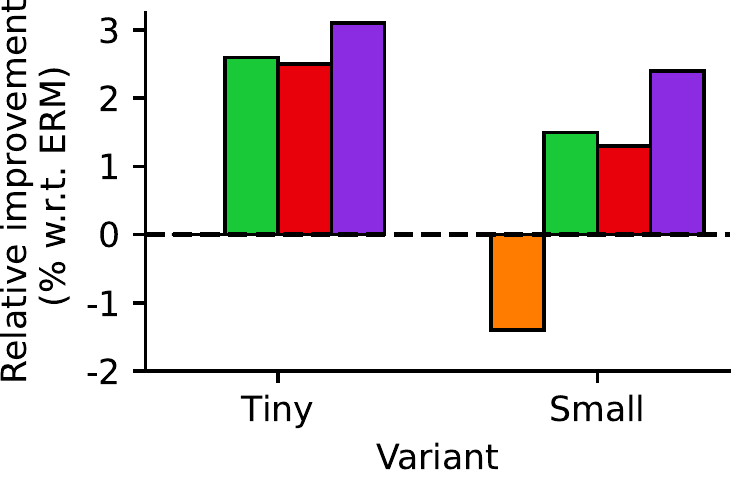} 
             & \includegraphics[width=0.48\columnwidth]{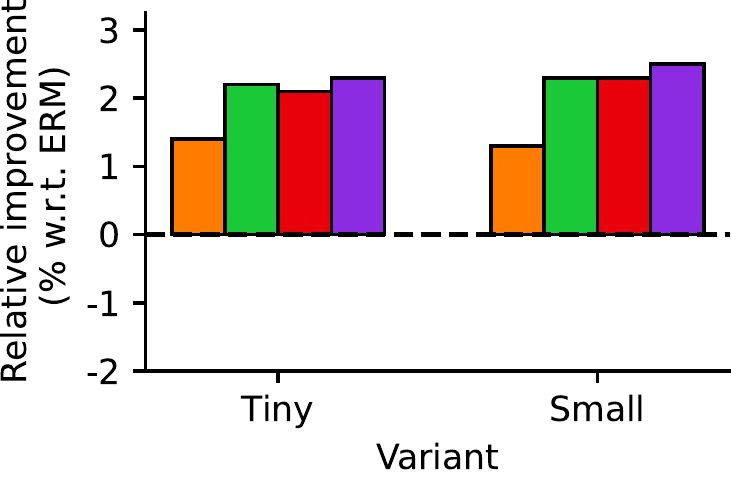}
        \end{tabular}
        \vspace{-2mm}
        \caption{Transformers}
        \label{fig:transformer_imagenet_delta}
    \end{subfigure}
    \vspace{-2mm}
    \caption{\textbf{Improvements across architectures with \INp{} variants 
    compared with \IN{}.}  Top-1 accuracy of different models on the \IN{} 
    validation set consistently improves when trained with the proposed 
    datasets as compared to the standard \IN{} training set (Epochs=150).  Our 
    proposed dataset variant with \RRCRARE{}, \textbf{\INp{}}, provides 
    balanced improvements of 1-4\% across architectures. Further improvements 
    with longer training (300-1000 epochs) in \cref{tab:long_train_effect}.
    }\label{fig:imagenet_delta_150}
    \vspace{-4mm}
\end{figure*}

In this section, we introduce \INp{}, a reinforcement of \IN{}.  We 
create \INp{} using the IG-ResNext ensemble (\cref{sec:good_teacher}).  
Following \citep{shen2021fast}, we store top $10$ sparse probabilities for 
$400$ augmentations per training sample in the \IN{} dataset 
\cite{deng2009imagenet}.
We consider the following augmentations: Random-Resize-Crop (\RRC{}), 
MixUp~\citep{zhang2017mixup} and CutMix~\citep{yun2019cutmix} (\Mixing{}), and 
RandomAugment~\citep{cubuk2020randaugment} and RandomErase (\RARE{}).
We also combine \Mixing{} with \RARE{} and refer to it as \MsRs.  We add all 
augmentations on top of \RRC{} and for clarity add + as shorthand for 
{\RRC{}$+$}. We provide a summary of the reinforcement data stored for each 
\INp{} variant in \cref{tab:imagenet_plus_specs}.

\vspace{-4mm}
\paragraph{Models} We study light-weight CNN-based (MobileNetV1 \cite{howard2017mobilenets}/ V2 
\cite{sandler2018mobilenetv2}/ V3\cite{howard2019searching}), heavy-weight 
CNN-based (ResNet \cite{he2016deep} and EfficientNet 
\cite{tan2019efficientnet}), and transformer-based (ViT 
\cite{dosovitskiy2020image} and SwinTransformer \cite{liu2021swin}) models. We 
follow \cite{mehta2022cvnets, wightman2021resnet} and use  state-of-the-art 
recipes, including optimizers, hyperparameters, and learning schedules, 
specific to each model on the \IN{}. We perform \textbf{no hyperparameter 
tuning specific to \INp{}} and achieve improvements with the same setup as 
\IN{} for all models.

\vspace{-4mm}
\paragraph{Better accuracy}
We evaluate the performance of each model in terms of top-1 accuracy on the 
\IN{} validation set.
\Cref{fig:imagenet_delta_150} compares the performance of different models 
trained using \IN{} and \INp{} datasets.  
\cref{fig:light_cnn_imagenet_delta} shows that light-weight CNN models do not 
benefit from difficult reinforcements. This is expected because of their 
limited capacity. On the other side, both heavy-weight CNN 
(\cref{fig:heavy_cnn_imagenet_delta}) and transformer-based 
(\cref{fig:transformer_imagenet_delta}) models benefit from difficult 
reinforcements (\RRCMixing{}, \RRCRARE{}, and \RRCMsRs{}). However, 
transformer-based models deliver best performance with the most difficult 
reinforcement (\RRCMsRs{}). This concurs with previous works that show 
transformer-based models, unlike CNNs, benefit from more data regularization as 
they do not have inductive 
biases~\citep{dosovitskiy2020image,touvron2021training}.  
 
Overall, \textbf{\RRCRARE{}} provides a balanced trade-off between performance 
and model size across different models.  Therefore, in the rest of this paper, 
we use \RRCRARE{} as our reinforced dataset and call it \textbf{\INp{}}. In 
the rest of the paper, we show results for three models that spans different 
model sizes and architecture designs (MobileNetV3-Large, ResNet-50, and 
SwinTransformer-Tiny).

We note that our observations are consistent across different architectures and 
recommend to see \cref{sec:imagenet_plus_full} for comprehensive results on 25 
architectures.
We provide expanded ablation studies in \cref{sec:reinforce_imagenet_sup}
using a cheaper teacher, ConvNext-Base-IN22FT1K. For example, we find 
1) The number of stored samples can be $3\times$ fewer than intended training 
epochs,
2) Additional augmentations on top of \INp{} are not useful.
3) Tradeoff in reinforcement difficulty can be further reduced with
curriculums.
4) Curriculums are better than various sample selection methods at the time of 
reinforcing the dataset.
We provide all hyperparameters and training recipes in \cref{sec:hparams}.

\subsection{\CIFARp{}, \Flowersp{}, \Foodp{}: How to reinforce other 
datasets?}
\label{sec:other_datasets}

We reinforced \IN{} due to its popularity and effectiveness as a pretraining 
dataset for other tasks (e.g., object detection).  Our findings on \IN{} are 
also useful for reinforcing other datasets and reduce the need for exhaustive 
studies.  Specifically, we suggest the following guidelines:
1) use ensemble of strong teachers trained on large diverse data
2) balance reinforcement difficulty and model complexity.

In this section, we extend dataset reinforcement to three other datasets, 
\CIFAR{}~\citep{krizhevsky2009learning}, 
\Flowers{}~\citep{nilsback2008automated}, and \Food{}~\citep{bossard14},
with 50K, 1K, and 75K training data respectively.
We build a teacher for each dataset by fine-tuning \INp{} pretrained 
ResNet-152 that reaches the accuracy of 90.6\%, 96.6\%, and 91.8\%, respectively.
By repeating fine-tuning 4 times, we get three teacher ensembles of 
4xResNet-152.
Next we generate reinforcements using similar augmentations to \INp{}, that 
is \RRCRARE{}. We store 800, 8000, and 800 augmentations per original sample.  
After that, we train various models on the reinforced data at similar training 
time to standard training. To achieve the best performance, we use pretrained 
models on \IN{}/\INp{} and fine-tune on each dataset for varying epochs 
up to 1000, 10000, and 1000 (for \CIFAR{}, \Flowers{}, and \Food{}, 
respectively) and report the best result.

\Cref{tab:transfer_cls_mobilenetv3} shows that MobileNetV3-Large pretrained and 
fine-tuned with reinforced datasets reaches up to 3\% better accuracy. We 
observe that pretraining and fine-tuning on reinforced datasets together give 
the largest improvements. We provide results for other models in
\cref{sec:transfer_full}.

\begin{table}[t!]
    \centering
    \resizebox{0.95\columnwidth}{!}{
        \begin{tabular}{lcc|cc|cc}
            \toprule[1.5pt]
            \multirow{2}{*}{\textbf{Pretraining Dataset}} 
            & \multicolumn{2}{c}{\textbf{\CIFAR{}}}
            & \multicolumn{2}{c}{\textbf{\Flowers{}}}
            & \multicolumn{2}{c}{\textbf{\Food{}}}  \\
            \cmidrule[1.25pt]{2-7}
             & \textbf{Orig.} & \textbf{+}
             & \textbf{Orig.} & \textbf{+}
             & \textbf{Orig.} & \textbf{+}\\
             \midrule[1.25pt]
             None & 80.2 & 83.6 & 68.8 & 87.5 & 85.1 & 88.2\\
             \IN{} &  84.4 & 87.2 & 92.5 & 94.1 & 86.1 & 89.2 \\
             \INp{} (Ours) &  86.0 & \textbf{87.5} & 93.7 & \textbf{95.3} & 86.6 &\textbf{89.5} \\
            \bottomrule[1.5pt]
        \end{tabular}
    }
    \vspace{-2mm}
    \caption{\textbf{Pretraining and fine-tuning on reinforced datasets is up 
    to  
    3.4\% better than using non-reinforced datasets.}
    Top-1 accuracy on the test set for MobileNetV3-Large is shown.
    On \Food{}, 86.1\% is improved to 89.5\%, demonstrating composition of 
    reinforced datasets.}
    \label{tab:transfer_cls_mobilenetv3}
    \vspace{-4mm}
\end{table}

\section{Experiments}\label{sec:experiments}
\paragraph{Baseline methods} We compare the performance of models trained using 
\INp{} with the following baseline methods: (1) 
\emph{KD}~\citep{hinton2015distilling,beyer2022knowledge} (Online 
distillation): A standard knowledge distillation method with strong teacher 
models and model-specific augmentations, (2)  
\emph{MEALV2}~\citep{shen2020meal} (Fine-tuning distillation): Distill 
knowledge to student with good initialization from multiple teachers, (3) 
\emph{FunMatch}~\citep{beyer2022knowledge} (Patient online distillation): 
Distill for significantly many epochs with strong augmentations, (4) 
\emph{ReLabel}~\citep{yun2021re} (Offline label-map distillation): Pre-computes 
global label maps from the pre-trained teacher, and (5) 
\emph{FKD}~\citep{shen2021fast} (Offline distillation): Pre-computes soft 
labels using multi-crop knowledge distillation. We consider FKD as the baseline 
approach for dataset reinforcement.

\vspace{-4mm}
\paragraph{Longer training}\label{sec:long_train}
Recent works have shown that models trained for few epochs (e.g.,\ 100 epochs) 
are sub-optimal and their performance improves with longer training 
\cite{wightman2021resnet,dosovitskiy2020image,touvron2021training}. Following these works, we train 
different models at three epoch budgets, i.e., 150, 300, and 1000 epochs, using 
both \IN{} and \INp{} datasets.  \Cref{tab:long_train_effect} shows  
models trained with \INp{} dataset consistently deliver better accuracy in 
comparison to the ones trained on \IN{}.
An epoch of \INp{} consists of exactly one random reinforcement per sample 
in \IN{}.

\begin{table}[t!]
    \centering
    \resizebox{0.95\columnwidth}{!}{
        \begin{tabular}{llccc}
            \toprule[1.5pt]
            \multirow{2}{*}{\textbf{Model}} & \multirow{2}{*}{\textbf{Dataset}} 
            & \multicolumn{3}{c}{\textbf{Training Epochs}}  \\
            \cmidrule[1.25pt]{3-5}
             & & \textbf{150} & \textbf{300} & \textbf{1000} \\
             \midrule[1.25pt]
             \multirow{2}{*}{MobileNetV3-Large} & \IN{} & 74.7 & 74.9 & 75.1 \\
             & \INp{} (Ours) & \textbf{76.2} & \textbf{77.0} &  \textbf{77.9} \\
             \midrule
             \multirow{2}{*}{ResNet-50} & \IN{} & 77.4 &  78.8 & 79.6 \\
             & \INp{} (Ours) & \textbf{79.6} & \textbf{80.6} & \textbf{81.7} \\
             \midrule
             \multirow{2}{*}{SwinTransformer-Tiny} & \IN{} & 79.9 & 80.9 &  80.9 \\
              & \INp{} (Ours) & \textbf{82.0} & \textbf{83.0} & \textbf{83.8} \\
            \bottomrule[1.5pt]
        \end{tabular}
    }
    \vspace{-2mm}
    \caption{\textbf{\INp{} models consistently outperform \IN{} models 
        when trained for longer}. Top-1 accuracy on the \IN{} validation set 
        is shown. An epoch has the same number of iterations for 
        \IN{}/\INp{}.}
    \label{tab:long_train_effect}
    \vspace{-4mm}
\end{table}

\vspace{-4mm}
\paragraph{Training and reinforcement time} \Cref{tab:long_train_effect} shows 
\INp{} improves the performance of various models. A natural question 
that arises is: \emph{Does \INp{} introduce computational overhead when 
training models?}
On average, training MobileNetV3-Large, ResNet-50, and SwinTransformer-Tiny is 
$1.12\times$, $1.01\times$, and $0.99\times$ the total training time on 
\IN{}. The extra time for MobileNetV3 is because there is no data 
augmentations in our baseline.
\INp{} took 2205 GPUh to generate using 64xA100 GPUs, which is highly 
parallelizable.
For comparison, training ResNet-50 for 
300 epochs on 8xA100 GPUs takes 206 GPUh.
The reinforcement generation is a one-time cost that is amortized over many 
uses.  The time to reinforce other datasets and the storage is discussed in 
\cref{sec:cost}.

\begin{table}[b!]
    \centering
    \resizebox{\columnwidth}{!}{
        \begin{tabular}{llccccc}
            \toprule[1.5pt]
            \multirow{2}{*}{\textbf{Model}} & \multirow{2}{*}{\textbf{Dataset}} & \textbf{Offline} & \textbf{Random } & \multirow{2}{*}{\textbf{Epochs}} & \multirow{2}{*}{\textbf{Accuracy}}   \\
             &  & \textbf{KD?} & \textbf{Init.?} &  &   \\
             \midrule[1.25pt]
              & \IN{}~\cite{howard2019searching} & NA & \cmark & 600 & 75.2 \\
             MobileNetV3& FunMatch~\citep{beyer2022knowledge}* & \xmark & \cmark & 1200 & 76.3\\
             -Large & MEALV2~\citep{shen2020meal} & \xmark & \xmark & 180 & 76.9 \\
             & \INp{} (Ours) & \cmark & \cmark & 300 & \textbf{77.0}\\
             \midrule[1pt]
             \multirow{6}{*}{ResNet-50} & \IN{} \cite{wightman2021resnet} & NA & \cmark & 600 & 80.4 \\
             & ReLabel~\citep{yun2021re} & \cmark & \cmark & 300 & 78.9\\
             & FKD~\citep{shen2021fast} & \cmark & \cmark & 300 & 80.1\\
             & MEALV2~\citep{shen2020meal} & \xmark & \xmark & 180 & 80.6\\
             & \INp{} (Ours) & \cmark & \cmark & 300 & 80.6\\
             & \INp{} (Ours) & \cmark & \cmark & 1000 & \textbf{81.7}\\
             & FunMatch~\citep{beyer2022knowledge}* & \xmark & \cmark & 1200 & \textbf{81.8}\\
             \midrule[0.5pt]
             ResNet-101
             & \IN{} \cite{wightman2021resnet} & NA & \cmark & 1000 & 81.5 \\
             \midrule[1pt]
             \multirow{4}{*}{ViT-Tiny} & \IN{} \cite{touvron2021training} & NA & \cmark & 300 & 72.2 \\
              & DeiT \cite{touvron2021training} & \xmark & \cmark & 300 & 74.5 \\
             & FKD~\citep{shen2021fast} & \cmark & \cmark & 300 & 75.2\\
             & \INp{} (Ours) & \cmark & \cmark & 300 & \textbf{75.8}\\
             \midrule[1pt]
             \multirow{3}{*}{ViT-Small}
             & \IN{}~\cite{touvron2021training} & NA & \cmark & 300 & 79.8 \\
             & DeiT~\cite{touvron2021training} & \xmark & \cmark & 300 & 81.2 \\
             & \INp{} (Ours) & \cmark & \cmark & 300 & \textbf{81.4}\\
             \midrule[1pt]
             \multirow{3}{*}{ViT-Base$\uparrow$384}
             & \IN{}~\citep{touvron2021training} & NA & \cmark & 300 & 83.1\\
             & DeiT~\cite{touvron2021training} & \xmark & \cmark & 300 & 83.4 \\
             & \INp{} (Ours) & \cmark & \cmark & 300 & \textbf{84.5}\\
            \bottomrule[1.5pt]
        \end{tabular}
    }
    \vspace{-2mm}
    \caption{\textbf{Comparison with state-of-the-art methods on the \IN{} 
    validation set.} Models trained with \INp{} dataset deliver similar or 
    better performance than existing methods. Importantly, unlike online KD 
    methods (e.g., FunMatch or DeiT), \INp{} does not add computational 
    overhead to standard \IN{} training (\cref{fig:wall_clock}). Here, NA 
    denotes standard supervised \IN{} training with no online/offline KD.
    $\uparrow$384 denotes training at 384 resolution. An epoch has the same 
    number of iterations for \IN{}/\INp{}.}
    \label{tab:comparison_to_literature}
    \vspace{-4mm}
\end{table}

\vspace{-4mm}
\paragraph{Comparison with state-of-the-art methods} 
\Cref{tab:comparison_to_literature} compares the performance of models trained 
with \INp{} and existing methods. We make following observations: (1) Compared 
to the closely related method, i.e., FKD, models trained using \INp{} deliver 
better accuracy. (2) We achieve comparable results to online distillation 
methods (e.g., FunMatch), but with fewer epochs and faster training 
(\cref{fig:wall_clock}). (3) Small variants of the same family trained with 
\INp{} achieve similar performance to larger models  trained with \IN{} 
dataset. For example, ResNet-50 (81.7\%) with \INp{} achieves similar 
performance  as ResNet-101 with \IN{} (81.5\%). We observe similar phenomenon 
across other models, including light-weight CNN models. This enables replacing 
large models with smaller variants in their family for faster inference across 
devices, including edge devices, without sacrificing accuracy.

\subsection{Transfer Learning}
\label{sec:transfer} 
To evaluate the transferability of models 
pre-trained using \INp{} dataset, we evaluate on following tasks: (1) 
semantic segmentation with DeepLabv3 \cite{chen2017rethinking} on the ADE20K 
dataset~\citep{zhou2019semantic}, (2) object detection with Mask-RCNN 
\cite{he2017mask} on the MS-COCO dataset~\citep{lin2014microsoft}, and (3) 
fine-grained classification on the \CIFAR{}~\citep{krizhevsky2009learning}, 
\Flowers{}~\citep{nilsback2008automated}, and \Food{}~\citep{bossard14} 
datasets.

\Cref{tab:transfer_det_seg,tab:transfer_cls} show models trained on the 
\INp{} dataset have better transferability properties as compared to the 
\IN{} dataset across different tasks (detection, segmentation, and 
fine-grained classification). To analyze the isolated impact of \INp{} in 
this section, the fine-tuning datasets are not reinforced. We present all 
combinations of training with reinforced/non-reinforced pretraining/fine-tuning 
datasets in \cref{sec:transfer_full}.

\begin{table}[t!]
    \centering
    \resizebox{0.8\columnwidth}{!}{
        \begin{tabular}{llcc}
            \toprule[1.5pt]
            \multirow{2}{*}{\textbf{Model}} 
            & \multicolumn{1}{c}{\multirow{2}{*}{\textbf{Pretraining dataset}}}
            & \multicolumn{2}{c}{\textbf{Task}}  \\
            \cmidrule[1.25pt]{3-4}
             & & ObjDet & SemSeg \\
             \midrule[1.25pt]
             \multirow{2}{*}{MobileNetV3-Large} & \IN{} & 35.5 & 37.2 \\
             & \INp{} (Ours) & \textbf{36.1} & \textbf{38.5} \\
             \midrule
             \multirow{2}{*}{ResNet-50} & \IN{} & 42.2 &  42.8 \\
             & \INp{} (Ours) & \textbf{42.5} & \textbf{44.2} \\
             \midrule
             \multirow{2}{*}{SwinTransformer-Tiny} & \IN{} & 45.8 & 41.2
             \\
              & \INp{} (Ours) & \textbf{46.5} & \textbf{42.5} \\
            \bottomrule[1.5pt]
        \end{tabular}
    }
    \vspace{-2mm}
    \caption{\textbf{Transfer learning for object detection and semantic 
    segmentation}. For object detection (ObjDet), we report standard mean 
    average precision on MS-COCO dataset while for sementic segmentation 
    (SemSeg), we report mean intersection accuracy on ADE20K dataset. Task 
    datasets are not reinforced.}
    \label{tab:transfer_det_seg}
    \vspace{-4mm}
\end{table}

\subsection{Robustness analysis} 
\label{sec:robustnes}
To evaluate the robustness of different models 
trained using the \INp{} dataset, we evaluate on three subsets of the 
\IN{}V2 dataset \citep{recht2019imagenet}, which is specifically designed to 
study the robustness of models trained on the \IN{} dataset. We also 
evaluate \IN{} models on other distribution shift datasets, 
\IN{}-A~\citep{hendrycks2021nae}, \IN{}-R~\citep{hendrycks2021many}, 
\IN{}-Sketch~\citep{wang2019learning}, ObjectNet~\citep{barbu2019objectnet}, 
and \IN{}-C~\citep{hendrycks2019robustness}.
We measure the top-1 accuracy except for \IN{}-C. On \IN{}-C, we measure 
the mean corruption error (mCE) and report 100 minus {mCE}.

\cref{tab:imagenetv2_accuracy} shows that models trained using \INp{} 
dataset are up to 20\% more robust.  Overall, these robustness results 
in conjunction with results in
\cref{tab:long_train_effect} highlight the effectiveness of the proposed 
dataset.

\begin{table*}[t!]
    \centering
    \resizebox{0.95\textwidth}{!}{
        \begin{tabular}{llcccccccc|c}
            \toprule[1.5pt]
            \multirow{2}{*}{\textbf{Model}} & \multirow{2}{*}{\textbf{Dataset}} 
            & \multicolumn{3}{c}{\textbf{\IN{}-V2}}
            & \multirow{2}{*}{\textbf{\IN{}-A}} 
            & \multirow{2}{*}{\textbf{\IN{}-R}} 
            & \multirow{2}{*}{\textbf{\IN{}-Sketch}} 
            & \multirow{2}{*}{\textbf{ObjectNet}} 
            & \multirow{2}{*}{\textbf{\IN{}-C}}
            & \multirow{2}{*}{\textbf{Avg.}}\\
            \cmidrule[1.25pt]{3-5}
             & & V2-A & V2-B & V2-C\\
             \midrule[1.25pt]
             \multirow{2}{*}{MobileNetV3-Large} & \IN{} & 71.5 & 62.9 & 76.8 
             & 4.5 & 32.4 & 20.6 & 32.8 & 21.8 & 30.4\\
             & \INp{} (Ours) & \textbf{75.1} & \textbf{66.3} 
             &  \textbf{80.5} & \textbf{7.6} & \textbf{42.0} & \textbf{29.0} 
             & \textbf{38.1} & \textbf{32.0} & \textbf{37.1}\\
             \midrule
             \multirow{2}{*}{ResNet-50} & \IN{} & 76.3 &  67.4 & 81.3
             & 11.9 & 38.1 & 27.4 & 41.6 & 33.2 & 37.9\\
             & \INp{} (Ours) & \textbf{79.3} & \textbf{71.3} 
             & \textbf{83.8}
             & \textbf{15.1} & \textbf{48.1} & \textbf{34.9} & \textbf{46.8} 
             & \textbf{39.0} & \textbf{43.7}\\
             \midrule
             \multirow{2}{*}{SwinTransformer-Tiny} & \IN{} & 77.0 & 69.3 
             &  81.6 & 21.0 & 37.7 & 25.4 & 40.5 & 36.9 & 39.6\\
              & \INp{} (Ours) & \textbf{81.5} & \textbf{74.1} 
              & \textbf{85.3} & \textbf{30.2} & \textbf{58.0}$^*$ 
              & \textbf{40.8} & \textbf{50.6} & \textbf{46.6} & \textbf{51.1}\\
            \bottomrule[1.5pt]
        \end{tabular}
    }
    \vspace{-2mm}
    \caption{\textbf{\INp{} models are up to 20\% more robust on \IN{} 
    distribution shifts}. All models are trained for 
    1000 epochs.  We report on \IN{}V2 variations Threshold-0.7 (V2-A), 
         Matched-Frequency (V2-B), and Top-Images (V2-C). We report accuracy on 
         all datasets except for \IN{}-C where we report 100 minus mCE
         metric. $^*$ Largest improvement.}
    \label{tab:imagenetv2_accuracy}
    \vspace{-2mm}
\end{table*}

\begin{table}[b!]
    \centering
    \resizebox{\columnwidth}{!}{
        \begin{tabular}{llccc}
            \toprule[1.5pt]
            \multirow{2}{*}{\textbf{Model}} 
            & \multicolumn{1}{c}{\multirow{2}{*}{\textbf{Pretraining dataset}}} 
            & \multicolumn{3}{c}{\textbf{Fine-tuning dataset}}  \\
            \cmidrule[1.25pt]{3-5}
            & & \CIFAR{} & \Flowers{} & \Food{}\\
             \midrule[1.25pt]
             \multirow{2}{*}{MobileNetV3-Large}
             & \IN{} &  84.4 & 92.5 & 86.1  \\
             & \INp{} (Ours) &  \textbf{86.0} & \textbf{93.7}  
             & \textbf{86.6}  \\
             \midrule
             \multirow{2}{*}{ResNet-50}
             & \IN{} & 88.4 &  93.6 & 90.0 \\
             & \INp{} (Ours) & \textbf{88.8} & \textbf{95.0} & \textbf{90.5} 
             \\
             \midrule
             \multirow{2}{*}{SwinTransformer-Tiny}
             & \IN{} & 90.6 & 96.3 &  92.3 \\
             & \INp{} (Ours) & \textbf{90.9} & \textbf{96.6} & \textbf{93.0} 
             \\
            \bottomrule[1.5pt]
        \end{tabular}
    }
    \vspace{-2mm}
    \caption{\textbf{Transfer learning for fine-grained object classification.} 
    Only pretraining dataset is reinforced and fine-tuning datasets are not 
    reinforced.  Reinforced pretraining/fine-tuning results in 
    \cref{tab:teaser_results}.}
    \label{tab:transfer_cls}
    \vspace{-4mm}
\end{table}

\subsection{Calibration: Why are \INp{} models robust and transferable?}
\label{sec:calibration}
\begin{figure}[b!]
    \centering
    \includegraphics[width=0.8\linewidth]{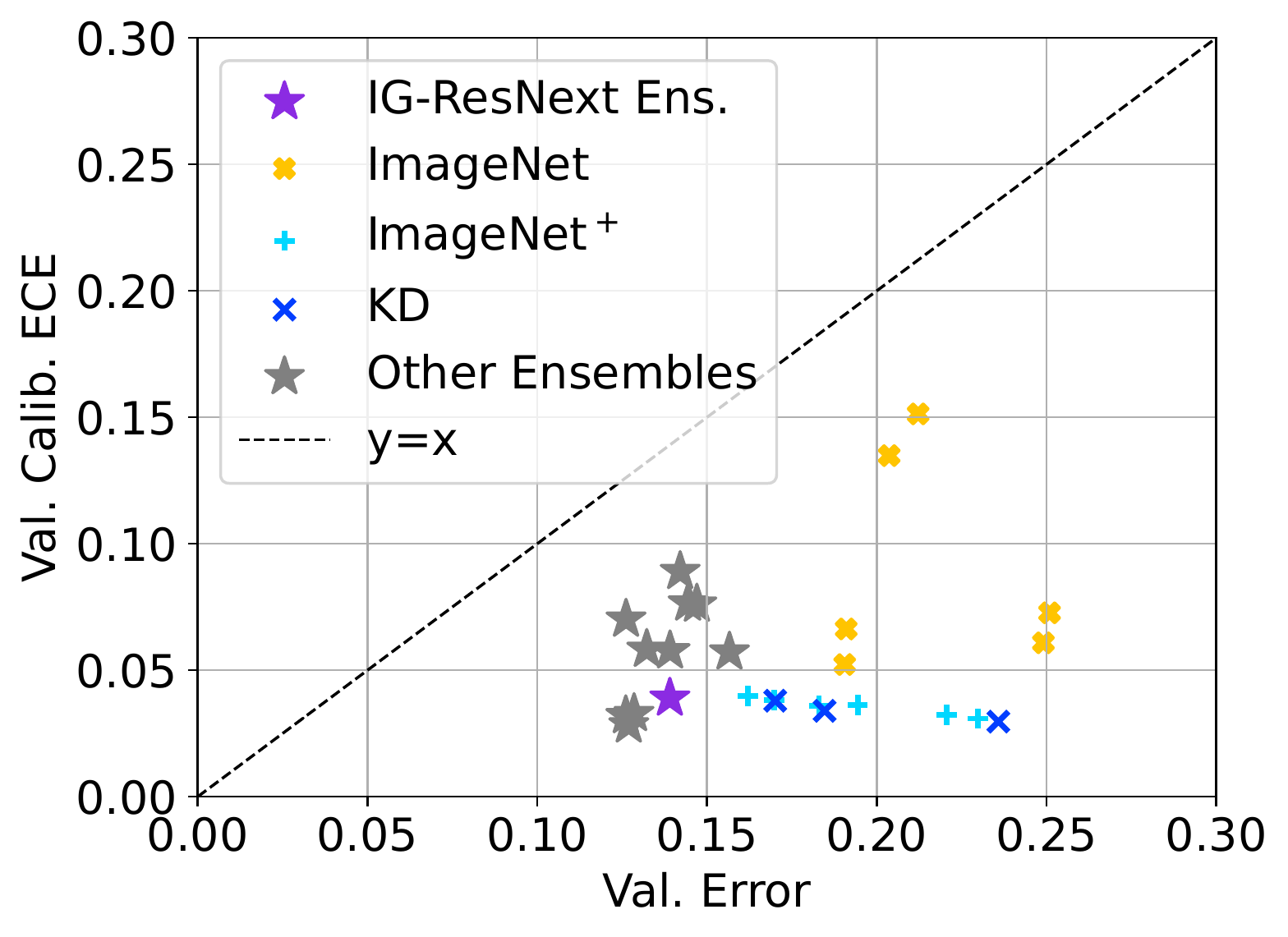}
       \vspace{-4mm}
    \caption{\textbf{\INp{} models are well-calibrated}.
    We plot the Expected Calibration Error (ECE) on the \IN{} validation set 
    over the validation error (normalized by 100 to range $[0, 1]$) for 
    MobileNetV3/ResNet-50/Swin-Tiny architectures trained for 300 and 1000 
    epochs on \IN{} and \INp{}.  \INp{} models are significantly more 
    calibrated, even matching or better than their teacher (IG-ResNext 
    Ensemble). We also observe that the IG-ResNext model is one of the best 
    calibrated models on the validation set from our pool of teachers.
    }
    \label{fig:val_calib_error}
\end{figure}

To understand why \INp{} models are significantly more robust than \IN{} 
models we evaluate their Expected Calibration Error 
(ECE)~\citep{kumar2019verified} on the validation set.  
\cref{fig:val_calib_error} shows that \INp{} models are well-calibrated and 
significantly better than \IN{} models. This matches recent observations 
about ensembles that out-of-distribution robustness is better for 
well-calibrated models~\citep{kumar2022calibrated}. Full calibration results are presented in 
\cref{sec:val_calib_error_full}.

\subsection{Comparison with FKD and ReLabel.}

We reproduce FKD and ReLabel with our training recipe as well as regenerate the 
dataset of FKD.
We compare the accuracy on \IN{} validation and its distribution shifts as 
well as the cost of dataset generation/storage.  We train models for 300 
epochs.

\vspace{-4mm}
\paragraph{Training recipe}
We report results of training with our code on the released datasets of 
ReLabel and {FKD}.
In addition to reproducing FKD results by training on their released dataset of 
500-sample per image, we also reproduce their dataset using our code and their 
teacher.  \cref{tab:fkd_comparison} verifies that our improvements are due to 
the superiority of \INp{}, not any other factors such as the training recipe.
Our \INp{}-RRC is also closely related to FKD as it uses the same set of 
augmentations (random-resized-crop and horizontal flip) but together with our 
optimal teacher (4xIG-ResNext).  We observe that \INp{}-RRC achieves better 
results than FKD but still lower than \INp{} 
(\cref{tab:imagenet_plus_table6_e1000,fig:imagenet_delta_150}).

\vspace{-4mm}
\paragraph{Generation/Storage Cost} We provide comparison of generation/storage 
costs in \cref{tab:fkd_comparison}. In our reproduction, generating FKD's data 
takes 2260 GPUh, slightly more than \INp{} because their teacher processes 
inputs at the larger resolution of $475\times 475$ compared to our resolution 
of $224\times 224$.

\vspace{-4mm}
\paragraph{\INp{}-Small} We subsampled \INp{} into a variant that is {10.6} 
GBs, comparable to prior work. We reduce the number of samples per image to 
{100} and store teacher probabilities with top-5 sparsity. If not subsampled 
from \INp{}, generating \INp{}-Small would take half the time of FKD (200 
samples) while still comparable in accuracy to \INp{}.  Note that \INp{} is 
more general-purpose and preferred, especially for long training.

\begin{table*}[tbh!]
    \vspace{-2mm}
    \centering
    \resizebox{0.95\textwidth}{!}{
        \begin{tabular}{ccccc|cc|cc|c|cc|cc}
            \toprule[1.5pt]
\textbf{Dataset} & \textbf{Our} & \textbf{Our} & \multicolumn{2}{c}{\textbf{Optimal}} & \textbf{Top-K} & \textbf{Num.} & \multicolumn{2}{c}{\textbf{Storage (GBs)}} & \textbf{Gen. Time} & \multicolumn{2}{c}{\textbf{ResNet-50}}& \multicolumn{2}{c}{\textbf{Swin-Tiny}} \\
& \textbf{Gen.} &\textbf{Train} & \textbf{Teacher} & \textbf{Aug.} &&\textbf{Samples}& Raw &GZIP& \textbf{(GPUh)}& IN & IN-OOD & IN & IN-OOD\\
ReLabel                    & \xmark & \cmark & \xmark & \xmark & 5 &   1 &          10.7 &  4.8&          10 & 79.5 & 45.7 & 81.2 & 48.2\\
FKD                        & \xmark & \cmark & \xmark & \xmark & 5 & 200 &          13.6 &  8.9&     904$^*$ & 79.8 & 45.0 & 82.0 & 48.7\\
FKD                        & \xmark & \cmark & \xmark & \xmark & 5 & 500 &          34.0 & 22.0&     2260$^*$& 80.1 & 45.0 & 82.2 & 48.9\\
FKD                        & \cmark & \cmark & \xmark & \xmark &10 & 400 &          46.3 & 33.4&        1808 & 79.8 & 45.0 & 82.1 & 49.0\\
\midrule[1.25pt]
\INp{}-RRC                 & \cmark & \cmark & \cmark & \xmark &10 & 400 &          46.3 & 33.4&         1993 & 80.3 & 46.5 & 82.4 & 51.0\\
\INp{}-Small  & \cmark & \cmark & \cmark & \cmark & 5 & 100 
            &  10.6& 5.6 & 551& \textbf{80.6} & \textbf{48.9} 
            & \textbf{82.9} & \textbf{54.6}\\
\INp{}                     & \cmark & \cmark & \cmark & \cmark &10 & 400 &          61.5 & 37.5&         2205 & \textbf{80.6} & \textbf{49.1} & \textbf{83.0} & \textbf{54.7}\\
\bottomrule[1.5pt]
        \end{tabular}
    }
    \vspace{-2mm}
    \caption{\textbf{Comparison with Relabel and FKD. Up to 5.6\% better than 
    FKD on ImageNet-OOD}, the average of \IN{}-V2/A/R/S/O/C accuracies.  
    Highlighted accuracies are within 0.2\% of the best. Compared with prior 
    work, we use an optimal teacher (4xIG-ResNext) and optimal combination of 
    augmentations (RRC+RA/RE). $^*$ Our estimates.}
    \label{tab:fkd_comparison}
\end{table*}

\subsection{CLIP-pretrained Teachers}

In this section, we evaluate the performance of CLIP-pretrained 
models~\citep{radford2021learning} fine-tuned on \IN{} as teachers. This 
study complements our large-scale study of teachers in \cref{sec:good_teacher} 
where we evaluated more than {100} SOTA large models and ensembles.  
\Cref{tab:imagenet_plus_clip_vit_mixed_short} compares an ensemble of {4} 
CLIP-pretrained models to our selected ensemble of 4 IG-ResNext models as well 
as a mixture of ResNext, ConvNext, CLIP-ViT, and ViT (abrv. RCCV) models (See 
\cref{sec:clip_vit_mix} for the model names). We generate new \INp{} 
variants and train various architectures for {1000} epochs on each dataset. We 
observe that \INp{} with our previously selected IG-ResNext ensemble is 
superior to CLIP-pretrained and mixed-architecture teachers across 
architectures.
The CLIP variant provides near the maximum gain on Swin-Tiny and mixing it with 
IG-ResNext reduces the gap on CNNs.

\begin{table}[tbh!]
    \centering
    \resizebox{0.98\columnwidth}{!}{
        \begin{tabular}{l c ccc}
\toprule
\multirow{2}{*}{\textbf{Model}} & \textbf{\IN{}} & \multicolumn{3}{c}{\textbf{\INp{}}}\\
\cmidrule[1.25pt]{3-5}
&  & \textbf{IG-ResNext}$^*$ & CLIP & Mixed\\
\midrule[1.25pt]
MobileNetV3-Large & ${75.1}$ & $\bm{77.9}_{\scriptscriptstyle +2.9}$ & ${77.2}_{\scriptscriptstyle +2.1}$  & ${77.4}_{\scriptscriptstyle +2.3}$\\
\midrule
ResNet-50 & ${79.6}$ & $\bm{81.7}_{\scriptscriptstyle +2.1}$ & ${81.1}_{\scriptscriptstyle +1.4}$  & $\bm{81.5}_{\scriptscriptstyle +1.8}$ \\
\midrule
Swin-Tiny & ${80.9}$ & $\bm{83.8}_{\scriptscriptstyle +2.8}$ & $\bm{83.7}_{\scriptscriptstyle +2.7}$  & $\bm{83.8}_{\scriptscriptstyle +2.8}$\\
\bottomrule[1.5pt]
\end{tabular}

    }
    \caption{\textbf{Our selected IG-ResNext ensemble is superior to 
    CLIP-pretrained ensembles.} We reinforce \IN{} dataset with an ensemble 
    of CLIP-pretrained models as well as a mixture of multiple architectures 
    and train various models for 1000 epochs.
    Subscripts show the improvement on top of the \IN{} accuracy.
    $^*$ Our chosen \INp{} variant.
    }
    \label{tab:imagenet_plus_clip_vit_mixed_short}
    \vspace{-5mm}
\end{table}

\section{Related work}
We build on top of the well-known Knowledge Distillation 
framework~\citep{bucilua2006model,ba2014deep,hinton2015distilling}, the 
effectiveness of which has been extensively 
studied~\citep{cho2019efficacy,stanton2021does}. Numerous variants of KD have 
been proposed, including feature distillation~\citep{ji2021show,zhang2020improve}, 
iterative distillation~\citep{mirzadeh2020improved,yang2019snapshot}, and
self-distillation~\citep{xie2020self,mobahi2020self,furlanello2018born,ji2021refine}.
Label smoothing, an effective regularizer and related to KD, is particularly 
related to our work when interpreted as
augmenting the output space~\citep{yuan2020revisiting,shen2021label}.

Closely related to our work, investigating and improving the accuracy on the \IN{} dataset has attracted 
much interest lately.
\citet{beyer2020we} eliminated erroneous labeled examples in the training with 
reference to a strong classifier. In \citet{shankar2020evaluating}, \IN{} 
dataset evaluation was revisited and alternative test sets were released.  
Relabel~\citet{yun2021re} proposed storing multiple labels on various regions 
of an image using a teacher. FKD~\citet{shen2021fast} further pushed this 
direction by caching the predictions of a strong teacher but with a limited 
augmentation.
Similarly, in \citet{ridnik2022solving}, the architecture-independent 
generalization of KD was exploited to propose a unified scheme for training with \IN{} 
seamlessly without any hyperparameter tuning or per-model training recipes.
\citet{liu2022meta} identified the temperature hyperparameter in KD as an 
important factor limiting benefits of stronger augmentations and teachers, and 
proposed an adaptive scheme to dynamically set the temperature during training. 
Distilling feature maps and probability distributions between the random pair 
of original images and their MixUp images was proposed to guide the network to 
learn cross-image knowledge \citet{pouransari2021extracurricular,yang2022mixskd}.
For self-supervised learning, \citet{kim2022generalized} adapted modern 
image-based regularizations with KD to improve the contrastive loss with some 
supervision. Our work has also been inspired by~\citep{beyer2022knowledge} 
where they proposed imitating the teacher on severe augmentations and train for 
thousands of epochs.
With our proposed DR strategy, we significantly reduce the 
cost of function matching by storing a few samples and reusing them for longer 
training.

\section{Conclusion}\label{sec:conclusion}

We go beyond the conventional online knowledge distillation and introduce 
Dataset Reinforcement (DR) as a general offline strategy.
Our investigation unwraps tradeoffs in finding generalizable reinforcements 
controlled by the difficulty of augmentations and we propose ways to balance.

We study the choice of the teacher (more than {100} SOTA large models and 
ensembles), augmentation (4 more than prior work), and their impact on 
a diverse collection of models (25 architectures), especially for long training 
(up to 1000 epochs). We demonstrate significant improvements (up to 20\%) in 
robustness, calibration and transfer (in/out of distribution classification, 
segmentation, and detection).  Our novel method of training and fine-tuning on 
doubly reinforced datasets (e.g., \INp{} to \CIFARp{}) demonstrates new 
possibilities of DR as a generic strategy.  We also study ideas that were not 
used in \INp{}, including curriculums, mixing augmentations and more in the 
appendix.

The proposed DR strategy is only an example of the large category of ideas 
possible within the scope of dataset reinforcement.  Our desiderata would also 
be satisfied by methods that expand the training data, especially in limited 
data domains, using strong generative foundation models.

\vspace{-4mm}
\paragraph{Limitations}
Limitations of the teacher can potentially transfer through dataset 
reinforcement.
For example, over-confident biased teachers should not be used and diverse 
ensembles are preferred.
Human verification of the reinforcements is also a solution.
Note that original labels are unmodified in reinforced datasets and can be used
in curriculums.
Our robustness and transfer learning evaluations consistently show better 
transfer and generalization for \INp{} models likely because of lower bias 
of the teacher ensemble trained on diverse data.

\subsubsection*{Acknowledgments}
We would like to thank Arsalan Farooq, Farzad Abdolhosseini, Keivan 
Alizadeh-Vahid, Pavan Kumar Anasosalu Vasu, and Raviteja Vemulapalli for the 
enriching discussions. We also thank the reviewers for their valuable feedback.
{\small
\bibliographystyle{ieee_fullname}
\bibliography{ref}
}

\newpage
\onecolumn
\appendix
\clearpage

\section{Full results of training on \IN{} and \INp{}, compared with 
Knowledge Distillation}\label{sec:imagenet_plus_full}

\Cref{tab:imagenet_plus_full} provides the full results of training with 
\INp{} compared with \IN{} and Knowledge Distillation (KD). We choose 
\RRCRARE{} that provides a balanced trade-off across architectures and training 
durations, and call it \INp{}.
Results in \cref{tab:imagenet_plus_full} are without some state-of-the-art 
training features that are further improved in \cref{tab:imagenet_plus_cvnets}.  

\Cref{tab:imagenet_plus_cvnets} provides improved results using 
state-of-the-art training recipes from the CVNets 
library~\citep{mehta2022cvnets}. We use the exact same \INp{} variant and 
only write a new dataset class in CVNets, further confirming our minimal code 
change claim. We note the training changes that help each model:
\begin{itemize}[leftmargin=*]
        \vspace{-2mm}
    \itemsep0em
    \item Higher resolution training: EfficientNets, ViT-Base, Swin-Base. We 
        observe that \INp{} reinforcements are resolution independent and 
        provide improvements even if the resolution is different from the one 
        used to generate them.
    \item Variable resolution with variable batch size training (VBS): ViTs, 
        EfficientNets, Swin.
    \item Mixed-precision: ViTs, Swin.
    \item Multi-node training: EfficientNets (resolution larger than 224).
    \item Exponential Model Averaging (EMA): MobileViTs.
    \item New results for MobileViT.
\end{itemize}

\begin{table*}[thb!]
    \centering
    \begin{subtable}[b]{.79\columnwidth}
        \centering
        \resizebox{0.63\columnwidth}{!}{
            \begin{tabular}{l|cc|cccc}%
\toprule[1.5pt]
\textbf{Model} & \textbf{\IN{}} & \textbf{KD} & \textbf{RRC} 
    & \textbf{RRC+Mixing} & \textbf{RRC+RA/RE} & \textbf{RRC+M*+R*}\\ 
    \midrule[1.25pt]
\verb|MobileNetv1-0.25| & $54.5$ & ${56.5}_{\scriptscriptstyle +2.0}$ & ${\bm{56.5}_{\scriptscriptstyle +2.0}}$ & ${55.2}_{\scriptscriptstyle +0.7}$ & ${55.7}_{\scriptscriptstyle +1.2}$ & ${53.9}_{\scriptscriptstyle -0.6}$ \\
\verb|MobileNetv1-0.5| & $66.3$ & ${66.3}_{\scriptscriptstyle -0.0}$ & ${\bm{67.6}_{\scriptscriptstyle +1.3}}$ & ${67.1}_{\scriptscriptstyle +0.8}$ & ${66.9}_{\scriptscriptstyle +0.6}$ & ${66.3}_{\scriptscriptstyle -0.0}$ \\
\verb|MobileNetv1-1.0| & $73.6$ & ${74.6}_{\scriptscriptstyle +1.0}$ & ${\bm{75.2}_{\scriptscriptstyle +1.6}}$ & ${\bm{75.0}_{\scriptscriptstyle +1.4}}$ & ${\bm{75.0}_{\scriptscriptstyle +1.4}}$ & ${74.3}_{\scriptscriptstyle +0.7}$ \\
\midrule
\verb|MobileNetv2-0.25| & $54.3$ & ${56.9}_{\scriptscriptstyle +2.6}$ & ${\bm{55.8}_{\scriptscriptstyle +1.5}}$ & ${54.1}_{\scriptscriptstyle -0.2}$ & ${53.8}_{\scriptscriptstyle -0.5}$ & ${52.5}_{\scriptscriptstyle -1.8}$ \\
\verb|MobileNetv2-0.5| & $65.3$ & ${66.2}_{\scriptscriptstyle +0.9}$ & ${\bm{66.0}_{\scriptscriptstyle +0.7}}$ & ${65.7}_{\scriptscriptstyle +0.4}$ & ${\bm{65.8}_{\scriptscriptstyle +0.4}}$ & ${64.0}_{\scriptscriptstyle -1.3}$ \\
\verb|MobileNetv2-1.0| & $72.7$ & ${72.8}_{\scriptscriptstyle +0.1}$ & ${\bm{73.8}_{\scriptscriptstyle +1.1}}$ & ${73.5}_{\scriptscriptstyle +0.8}$ & ${73.5}_{\scriptscriptstyle +0.8}$ & ${72.9}_{\scriptscriptstyle +0.2}$ \\
\midrule
\verb|MobileNetv3-Small| & $66.3$ & ${65.8}_{\scriptscriptstyle -0.5}$ & ${\bm{68.0}_{\scriptscriptstyle +1.7}}$ & ${67.1}_{\scriptscriptstyle +0.8}$ & ${67.3}_{\scriptscriptstyle +1.0}$ & ${66.4}_{\scriptscriptstyle +0.2}$ \\
\verb|MobileNetv3-Large| & $74.7$ & ${75.5}_{\scriptscriptstyle +0.9}$ & ${\bm{76.0}_{\scriptscriptstyle +1.4}}$ & ${\bm{76.0}_{\scriptscriptstyle +1.4}}$ & ${\bm{76.2}_{\scriptscriptstyle +1.6}}$ & ${75.5}_{\scriptscriptstyle +0.8}$ \\
\midrule
\verb|ResNet-18| & $67.8$ & ${72.1}_{\scriptscriptstyle +4.3}$ & ${\bm{72.3}_{\scriptscriptstyle +4.5}}$ & ${71.7}_{\scriptscriptstyle +4.0}$ & ${71.9}_{\scriptscriptstyle +4.1}$ & ${71.0}_{\scriptscriptstyle +3.2}$ \\
\verb|ResNet-34| & $73.2$ & ${76.4}_{\scriptscriptstyle +3.2}$ & ${\bm{76.2}_{\scriptscriptstyle +3.0}}$ & ${\bm{76.3}_{\scriptscriptstyle +3.1}}$ & ${\bm{76.2}_{\scriptscriptstyle +3.0}}$ & ${75.8}_{\scriptscriptstyle +2.6}$ \\
\verb|ResNet-50| & $77.4$ & ${80.3}_{\scriptscriptstyle +2.9}$ & ${\bm{79.6}_{\scriptscriptstyle +2.3}}$ & ${\bm{79.7}_{\scriptscriptstyle +2.3}}$ & ${\bm{79.6}_{\scriptscriptstyle +2.3}}$ & ${\bm{79.6}_{\scriptscriptstyle +2.2}}$ \\
\verb|ResNet-101| & $79.8$ & ${81.7}_{\scriptscriptstyle +1.9}$ & ${81.2}_{\scriptscriptstyle +1.4}$ & ${\bm{81.7}_{\scriptscriptstyle +1.8}}$ & ${\bm{81.5}_{\scriptscriptstyle +1.7}}$ & ${81.3}_{\scriptscriptstyle +1.5}$ \\
\verb|ResNet-152| & $80.8$ & ${82.3}_{\scriptscriptstyle +1.4}$ & ${81.6}_{\scriptscriptstyle +0.8}$ & ${\bm{82.0}_{\scriptscriptstyle +1.2}}$ & ${\bm{82.0}_{\scriptscriptstyle +1.1}}$ & ${\bm{81.9}_{\scriptscriptstyle +1.0}}$ \\
\midrule
\verb|EfficientNet-B2| & $77.9$ & ${80.0}_{\scriptscriptstyle +2.1}$ & ${80.2}_{\scriptscriptstyle +2.3}$ & ${\bm{80.6}_{\scriptscriptstyle +2.7}}$ & ${\bm{80.7}_{\scriptscriptstyle +2.7}}$ & ${80.4}_{\scriptscriptstyle +2.4}$ \\
\verb|EfficientNet-B3| & $79.3$ & ${80.9}_{\scriptscriptstyle +1.6}$ & ${81.1}_{\scriptscriptstyle +1.8}$ & ${\bm{81.5}_{\scriptscriptstyle +2.2}}$ & ${\bm{81.6}_{\scriptscriptstyle +2.3}}$ & ${\bm{81.5}_{\scriptscriptstyle +2.2}}$ \\
\verb|EfficientNet-B4|  & $79.4$ & ${81.8}_{\scriptscriptstyle +2.4}$ & ${81.0}_{\scriptscriptstyle +1.6}$ & ${\bm{81.3}_{\scriptscriptstyle +1.9}}$ & ${\bm{81.5}_{\scriptscriptstyle +2.1}}$ & ${\bm{81.3}_{\scriptscriptstyle +1.9}}$ \\
\midrule
\verb|ViT-Tiny| & $71.5$ & ${72.0}_{\scriptscriptstyle +0.5}$ & ${71.5}_{\scriptscriptstyle +0.0}$ & ${74.1}_{\scriptscriptstyle +2.6}$ & ${74.0}_{\scriptscriptstyle +2.6}$ & ${\bm{74.6}_{\scriptscriptstyle +3.1}}$ \\
\verb|ViT-Small| & $78.4$ & ${80.2}_{\scriptscriptstyle +1.7}$ & ${77.0}_{\scriptscriptstyle -1.5}$ & ${79.9}_{\scriptscriptstyle +1.4}$ & ${79.7}_{\scriptscriptstyle +1.2}$ & ${\bm{80.8}_{\scriptscriptstyle +2.4}}$ \\
\midrule
\verb|Swin-Tiny| & $79.9$ & ${81.7}_{\scriptscriptstyle +1.7}$ & ${81.3}_{\scriptscriptstyle +1.4}$ & ${\bm{82.1}_{\scriptscriptstyle +2.2}}$ & ${\bm{82.0}_{\scriptscriptstyle +2.1}}$ & ${\bm{82.2}_{\scriptscriptstyle +2.2}}$ \\
\verb|Swin-Small| & $80.6$ & ${83.4}_{\scriptscriptstyle +2.9}$ & ${81.9}_{\scriptscriptstyle +1.3}$ & ${\bm{82.9}_{\scriptscriptstyle +2.3}}$ & ${\bm{82.9}_{\scriptscriptstyle +2.4}}$ & ${\bm{83.1}_{\scriptscriptstyle +2.6}}$ \\
\bottomrule[1.5pt]
\end{tabular}

        }
        \caption{150 epochs}
        \label{tab:imagenet_plus_table6_e150}
    \end{subtable}
    \vfill
    \begin{subtable}[b]{.79\columnwidth}
        \centering
        \resizebox{0.63\columnwidth}{!}{
            \begin{tabular}{l|cc|cccc}%
\toprule[1.5pt]
\textbf{Model} & \textbf{\IN{}} & \textbf{KD} & \textbf{RRC} 
    & \textbf{RRC+Mixing} & \textbf{RRC+RA/RE} & \textbf{RRC+M*+R*}\\ 
    \midrule[1.25pt]
\verb|MobileNetv1-0.25| & $55.7$ & ${57.4}_{\scriptscriptstyle +1.8}$ & ${\bm{56.8}_{\scriptscriptstyle +1.2}}$ & ${56.2}_{\scriptscriptstyle +0.5}$ & ${55.9}_{\scriptscriptstyle +0.3}$ & ${55.1}_{\scriptscriptstyle -0.6}$ \\
\verb|MobileNetv1-0.5| & $67.1$ & ${67.5}_{\scriptscriptstyle +0.4}$ & ${\bm{68.0}_{\scriptscriptstyle +1.0}}$ & ${67.7}_{\scriptscriptstyle +0.7}$ & ${\bm{68.0}_{\scriptscriptstyle +0.9}}$ & ${66.9}_{\scriptscriptstyle -0.1}$ \\
\verb|MobileNetv1-1.0| & $74.0$ & ${75.9}_{\scriptscriptstyle +1.9}$ & ${75.6}_{\scriptscriptstyle +1.6}$ & ${\bm{75.8}_{\scriptscriptstyle +1.8}}$ & ${\bm{75.9}_{\scriptscriptstyle +1.8}}$ & ${75.4}_{\scriptscriptstyle +1.3}$ \\
\midrule
\verb|MobileNetv2-0.25| & $54.8$ & ${57.7}_{\scriptscriptstyle +2.9}$ & ${\bm{56.2}_{\scriptscriptstyle +1.4}}$ & ${55.2}_{\scriptscriptstyle +0.4}$ & ${54.9}_{\scriptscriptstyle +0.1}$ & ${53.1}_{\scriptscriptstyle -1.7}$ \\
\verb|MobileNetv2-0.5| & $66.0$ & ${66.9}_{\scriptscriptstyle +0.9}$ & ${\bm{67.2}_{\scriptscriptstyle +1.3}}$ & ${66.5}_{\scriptscriptstyle +0.5}$ & ${66.4}_{\scriptscriptstyle +0.5}$ & ${65.3}_{\scriptscriptstyle -0.7}$ \\
\verb|MobileNetv2-1.0| & $73.3$ & ${74.3}_{\scriptscriptstyle +1.0}$ & ${\bm{74.8}_{\scriptscriptstyle +1.5}}$ & ${74.2}_{\scriptscriptstyle +0.9}$ & ${74.5}_{\scriptscriptstyle +1.2}$ & ${73.9}_{\scriptscriptstyle +0.6}$ \\
\midrule
\verb|MobileNetv3-Small|  & $67.2$ & ${67.0}_{\scriptscriptstyle -0.2}$ & ${\bm{69.0}_{\scriptscriptstyle +1.8}}$ & ${68.1}_{\scriptscriptstyle +0.8}$ & ${68.2}_{\scriptscriptstyle +1.0}$ & ${67.1}_{\scriptscriptstyle -0.1}$ \\
\verb|MobileNetv3-Large| & $74.9$ & ${76.4}_{\scriptscriptstyle +1.5}$ & ${76.6}_{\scriptscriptstyle +1.7}$ & ${\bm{76.9}_{\scriptscriptstyle +2.0}}$ & ${\bm{77.0}_{\scriptscriptstyle +2.1}}$ & ${76.5}_{\scriptscriptstyle +1.6}$ \\
\midrule
\verb|ResNet-18| & $68.7$ & ${73.5}_{\scriptscriptstyle +4.8}$ & ${\bm{72.7}_{\scriptscriptstyle +4.0}}$ & ${\bm{72.7}_{\scriptscriptstyle +4.0}}$ & ${\bm{72.7}_{\scriptscriptstyle +4.0}}$ & ${72.1}_{\scriptscriptstyle +3.4}$ \\
\verb|ResNet-34| & $74.3$ & ${77.9}_{\scriptscriptstyle +3.6}$ & ${76.9}_{\scriptscriptstyle +2.6}$ & ${\bm{77.3}_{\scriptscriptstyle +3.0}}$ & ${\bm{77.2}_{\scriptscriptstyle +2.9}}$ & ${76.9}_{\scriptscriptstyle +2.6}$ \\
\verb|ResNet-50| & $78.8$ & ${81.5}_{\scriptscriptstyle +2.8}$ & ${80.3}_{\scriptscriptstyle +1.5}$ & ${\bm{80.8}_{\scriptscriptstyle +2.0}}$ & ${\bm{80.6}_{\scriptscriptstyle +1.8}}$ & ${80.5}_{\scriptscriptstyle +1.7}$ \\
\verb|ResNet-101| & $80.9$ & ${83.0}_{\scriptscriptstyle +2.1}$ & ${81.8}_{\scriptscriptstyle +0.9}$ & ${\bm{82.3}_{\scriptscriptstyle +1.4}}$ & ${\bm{82.3}_{\scriptscriptstyle +1.4}}$ & ${\bm{82.4}_{\scriptscriptstyle +1.5}}$ \\
\verb|ResNet-152| & $81.5$ & ${83.4}_{\scriptscriptstyle +1.9}$ & ${82.5}_{\scriptscriptstyle +1.0}$ & ${\bm{83.0}_{\scriptscriptstyle +1.5}}$ & ${\bm{82.8}_{\scriptscriptstyle +1.3}}$ & ${\bm{82.9}_{\scriptscriptstyle +1.4}}$ \\
\midrule
\verb|EfficientNet-B2| & $78.9$ & ${81.3}_{\scriptscriptstyle +2.3}$ & ${80.9}_{\scriptscriptstyle +1.9}$ & ${\bm{81.2}_{\scriptscriptstyle +2.3}}$ & ${\bm{81.2}_{\scriptscriptstyle +2.2}}$ & ${\bm{81.1}_{\scriptscriptstyle +2.2}}$ \\
\verb|EfficientNet-B3| & $80.1$ & ${82.1}_{\scriptscriptstyle +2.0}$ & ${81.7}_{\scriptscriptstyle +1.6}$ & ${\bm{82.2}_{\scriptscriptstyle +2.1}}$ & ${\bm{82.1}_{\scriptscriptstyle +2.0}}$ & ${\bm{82.1}_{\scriptscriptstyle +2.0}}$ \\
\verb|EfficientNet-B4| & $80.6$ & ${83.0}_{\scriptscriptstyle +2.3}$ & ${82.1}_{\scriptscriptstyle +1.5}$ & ${\bm{82.4}_{\scriptscriptstyle +1.8}}$ & ${\bm{82.3}_{\scriptscriptstyle +1.7}}$ & ${\bm{82.4}_{\scriptscriptstyle +1.7}}$ \\
\midrule
\verb|ViT-Tiny| & $74.1$ & ${75.5}_{\scriptscriptstyle +1.4}$ & ${73.5}_{\scriptscriptstyle -0.6}$ & ${76.2}_{\scriptscriptstyle +2.0}$ & ${75.8}_{\scriptscriptstyle +1.7}$ & ${\bm{76.9}_{\scriptscriptstyle +2.7}}$ \\
\verb|ViT-Small| & $78.9$ & ${82.3}_{\scriptscriptstyle +3.4}$ & ${79.2}_{\scriptscriptstyle +0.2}$ & ${81.6}_{\scriptscriptstyle +2.7}$ & ${81.4}_{\scriptscriptstyle +2.4}$ & ${\bm{82.3}_{\scriptscriptstyle +3.3}}$ \\
\midrule
\verb|Swin-Tiny| & $80.9$ & ${83.0}_{\scriptscriptstyle +2.1}$ & ${82.4}_{\scriptscriptstyle +1.5}$ & ${\bm{83.1}_{\scriptscriptstyle +2.2}}$ & ${\bm{83.0}_{\scriptscriptstyle +2.1}}$ & ${\bm{83.2}_{\scriptscriptstyle +2.3}}$ \\
\verb|Swin-Small| & $81.4$ & ${84.4}_{\scriptscriptstyle +3.0}$ & ${82.5}_{\scriptscriptstyle +1.1}$ & ${\bm{83.7}_{\scriptscriptstyle +2.3}}$ & ${\bm{83.9}_{\scriptscriptstyle +2.5}}$ & ${\bm{83.9}_{\scriptscriptstyle +2.5}}$ \\
\bottomrule[1.5pt]
\end{tabular}

        }
        \caption{300 epochs}
        \label{tab:imagenet_plus_table6_e300}
    \end{subtable}
    \vfill
    \begin{subtable}[b]{.79\columnwidth}
        \centering
        \resizebox{0.6\columnwidth}{!}{
            \begin{tabular}{l|c|cccc}%
\toprule[1.5pt]
\textbf{Model} & \textbf{\IN{}} & \textbf{RRC} & \textbf{RRC+Mixing} 
    & \textbf{RRC+RA/RE} & \textbf{RRC+M*+R*}\\ \midrule[1.25pt]
\verb|MobileNetv1-0.25| & $56.7$ & ${\bm{57.6}_{\scriptscriptstyle +0.9}}$ & ${57.1}_{\scriptscriptstyle +0.4}$ & ${57.1}_{\scriptscriptstyle +0.4}$ & ${56.3}_{\scriptscriptstyle -0.4}$ \\
\verb|MobileNetv1-0.5| & $67.8$ & ${\bm{69.2}_{\scriptscriptstyle +1.4}}$ & ${68.8}_{\scriptscriptstyle +1.0}$ & ${68.5}_{\scriptscriptstyle +0.7}$ & ${68.0}_{\scriptscriptstyle +0.2}$ \\
\verb|MobileNetv1-1.0| & $74.1$ & ${76.4}_{\scriptscriptstyle +2.2}$ & ${\bm{76.7}_{\scriptscriptstyle +2.6}}$ & ${\bm{76.7}_{\scriptscriptstyle +2.5}}$ & ${76.4}_{\scriptscriptstyle +2.3}$ \\
\midrule
\verb|MobileNetv2-0.25| & $55.7$ & ${\bm{56.7}_{\scriptscriptstyle +1.0}}$ & ${55.8}_{\scriptscriptstyle +0.1}$ & ${55.2}_{\scriptscriptstyle -0.5}$ & ${55.0}_{\scriptscriptstyle -0.7}$ \\
\verb|MobileNetv2-0.5| & $66.8$ & ${\bm{68.2}_{\scriptscriptstyle +1.4}}$ & ${67.3}_{\scriptscriptstyle +0.5}$ & ${67.1}_{\scriptscriptstyle +0.3}$ & ${65.9}_{\scriptscriptstyle -0.9}$ \\
\verb|MobileNetv2-1.0| & $73.9$ & ${\bm{75.4}_{\scriptscriptstyle +1.5}}$ & ${\bm{75.3}_{\scriptscriptstyle +1.4}}$ & ${\bm{75.5}_{\scriptscriptstyle +1.6}}$ & ${74.7}_{\scriptscriptstyle +0.8}$ \\
\midrule
\verb|MobileNetv3-Small| & $67.9$ & ${69.0}_{\scriptscriptstyle +1.1}$ & ${68.4}_{\scriptscriptstyle +0.5}$ & ${\bm{69.4}_{\scriptscriptstyle +1.4}}$ & ${68.4}_{\scriptscriptstyle +0.4}$ \\
\verb|MobileNetv3-Large| & $75.1$ & ${77.2}_{\scriptscriptstyle +2.1}$ & ${77.4}_{\scriptscriptstyle +2.3}$ & ${\bm{77.9}_{\scriptscriptstyle +2.9}}$ & ${77.5}_{\scriptscriptstyle +2.4}$ \\
\midrule
\verb|ResNet-18| & $69.9$ & ${73.6}_{\scriptscriptstyle +3.6}$ & ${\bm{73.9}_{\scriptscriptstyle +4.0}}$ & ${\bm{73.8}_{\scriptscriptstyle +3.8}}$ & ${73.3}_{\scriptscriptstyle +3.4}$ \\
\verb|ResNet-34| & $75.6$ & ${77.8}_{\scriptscriptstyle +2.2}$ & ${\bm{78.4}_{\scriptscriptstyle +2.8}}$ & ${\bm{78.4}_{\scriptscriptstyle +2.8}}$ & ${78.1}_{\scriptscriptstyle +2.6}$ \\
\verb|ResNet-50| & $79.6$ & ${81.1}_{\scriptscriptstyle +1.4}$ & ${\bm{81.8}_{\scriptscriptstyle +2.2}}$ & ${\bm{81.7}_{\scriptscriptstyle +2.1}}$ & ${\bm{81.8}_{\scriptscriptstyle +2.2}}$ \\
\verb|ResNet-101| & $81.4$ & ${82.7}_{\scriptscriptstyle +1.3}$ & ${\bm{83.6}_{\scriptscriptstyle +2.2}}$ & ${83.2}_{\scriptscriptstyle +1.8}$ & ${\bm{83.4}_{\scriptscriptstyle +2.0}}$ \\
\verb|ResNet-152| & $81.7$ & ${83.4}_{\scriptscriptstyle +1.7}$ & ${\bm{84.0}_{\scriptscriptstyle +2.3}}$ & ${\bm{83.8}_{\scriptscriptstyle +2.2}}$ & ${\bm{83.9}_{\scriptscriptstyle +2.3}}$ \\
\midrule
\verb|EfficientNet-B2| & $79.3$ & ${81.5}_{\scriptscriptstyle +2.2}$ & ${\bm{81.9}_{\scriptscriptstyle +2.7}}$ & ${\bm{81.9}_{\scriptscriptstyle +2.7}}$ & ${\bm{81.7}_{\scriptscriptstyle +2.5}}$ \\
\verb|EfficientNet-B3| & $79.6$ & ${82.3}_{\scriptscriptstyle +2.7}$ & ${\bm{82.9}_{\scriptscriptstyle +3.3}}$ & ${\bm{82.8}_{\scriptscriptstyle +3.3}}$ & ${\bm{82.7}_{\scriptscriptstyle +3.2}}$ \\
\verb|EfficientNet-B4| & $81.2$ & ${82.9}_{\scriptscriptstyle +1.8}$ & ${\bm{83.2}_{\scriptscriptstyle +2.1}}$ & ${\bm{83.1}_{\scriptscriptstyle +1.9}}$ & ${\bm{83.2}_{\scriptscriptstyle +2.0}}$ \\
\midrule
\verb|ViT-Tiny| & $75.9$ & ${76.6}_{\scriptscriptstyle +0.7}$ & ${\bm{78.5}_{\scriptscriptstyle +2.6}}$ & ${78.1}_{\scriptscriptstyle +2.1}$ & ${\bm{78.7}_{\scriptscriptstyle +2.8}}$ \\
\verb|ViT-Small| & $78.4$ & ${80.8}_{\scriptscriptstyle +2.4}$ & ${83.2}_{\scriptscriptstyle +4.8}$ & ${82.6}_{\scriptscriptstyle +4.2}$ & ${\bm{83.7}_{\scriptscriptstyle +5.3}}$ \\
\midrule
\verb|Swin-Tiny| & $80.9$ & ${83.4}_{\scriptscriptstyle +2.5}$ & ${\bm{84.1}_{\scriptscriptstyle +3.1}}$ & ${83.8}_{\scriptscriptstyle +2.8}$ & ${\bm{84.0}_{\scriptscriptstyle +3.1}}$ \\
\verb|Swin-Small| & $81.9$ & ${83.9}_{\scriptscriptstyle +1.9}$ & ${84.5}_{\scriptscriptstyle +2.6}$ & ${84.4}_{\scriptscriptstyle +2.5}$ & ${\bm{84.8}_{\scriptscriptstyle +2.9}}$ \\
\bottomrule[1.5pt]
\end{tabular}

        }
        \caption{1000 epochs}
        \label{tab:imagenet_plus_table6_e1000}
    \end{subtable}
    \caption{Comparison of training different models using knowledge 
    distillation and different \IN{}/\INp{} datasets. Subscripts show the 
    improvement on top of the \IN{} accuracy. We highlight the best accuracy 
    on each row from our proposed datasets and any number that is within $0.2$ 
    of the best.
    Knowledge distillation results are not reported for E=1000 
    (\cref{tab:imagenet_plus_table6_e1000}) as it is computationally very 
    expensive.}
    \label{tab:imagenet_plus_full}
\end{table*}

\begin{table*}[thb!]
    \centering
    \begin{subtable}[b]{.77\columnwidth}
        \centering
        \resizebox{0.59\columnwidth}{!}{
            \begin{tabular}{l|cc|cc|cc}
\toprule
\multirow{2}{*}{\textbf{Model}} & \multicolumn{2}{c}{\textbf{Base Recipes (\cref{tab:imagenet_plus_full})}} & \multicolumn{2}{c}{\textbf{CVNets}} & \multicolumn{2}{c}{\textbf{CVNets-EMA}}\\
\cmidrule[1.25pt]{2-7}
& \textbf{\IN{}} & \textbf{\INp{}} & \textbf{\IN{}} & \textbf{\INp{}} & \textbf{\IN{}} & \textbf{\INp{}}\\
\midrule[1.25pt]
\verb|MobileNetV1-0.25| & ${54.5}$ & $\bm{55.7}_{\scriptscriptstyle +1.2}$ & ${55.2}$ & ${55.4}_{\scriptscriptstyle +0.2}$ & ${55.4}$ & ${55.4}_{\scriptscriptstyle +0.0}$ \\
\verb|MobileNetV1-0.5| & ${66.3}$ & $\bm{66.9}_{\scriptscriptstyle +0.6}$ & ${66.2}$ & $\bm{67.1}_{\scriptscriptstyle +0.9}$ & ${66.4}$ & $\bm{67.1}_{\scriptscriptstyle +0.7}$ \\
\verb|MobileNetV1-1.0| & ${73.6}$ & $\bm{75.0}_{\scriptscriptstyle +1.4}$ & ${73.5}$ & $\bm{75.1}_{\scriptscriptstyle +1.5}$ & ${73.6}$ & $\bm{75.1}_{\scriptscriptstyle +1.5}$ \\
\midrule
\verb|MobileNetV2-0.25| & ${54.3}$ & ${53.8}_{\scriptscriptstyle -0.5}$ & $\bm{54.7}$ & ${54.2}_{\scriptscriptstyle -0.5}$ & $\bm{54.7}$ & ${54.2}_{\scriptscriptstyle -0.5}$ \\
\verb|MobileNetV2-0.5| & ${65.3}$ & $\bm{65.8}_{\scriptscriptstyle +0.4}$ & $\bm{65.7}$ & $\bm{65.7}_{\scriptscriptstyle -0.0}$ & $\bm{65.7}$ & $\bm{65.7}_{\scriptscriptstyle +0.0}$ \\
\verb|MobileNetV2-1.0| & ${72.7}$ & ${73.5}_{\scriptscriptstyle +0.8}$ & ${72.8}$ & $\bm{73.8}_{\scriptscriptstyle +1.0}$ & ${72.8}$ & $\bm{73.9}_{\scriptscriptstyle +1.1}$ \\
\midrule
\verb|MobileNetV3-Small| & ${66.3}$ & ${67.3}_{\scriptscriptstyle +1.0}$ & ${66.6}$ & $\bm{67.7}_{\scriptscriptstyle +1.1}$ & ${66.7}$ & $\bm{67.7}_{\scriptscriptstyle +1.1}$ \\
\verb|MobileNetV3-Large| & ${74.7}$ & ${76.2}_{\scriptscriptstyle +1.6}$ & ${74.7}$ & $\bm{76.5}_{\scriptscriptstyle +1.8}$ & ${74.8}$ & $\bm{76.5}_{\scriptscriptstyle +1.7}$ \\
\midrule
\verb|MobileViT-XXSmall| & - & - & ${66.0}$ & ${67.4}_{\scriptscriptstyle +1.5}$ & ${66.7}$ & $\bm{67.9}_{\scriptscriptstyle +1.2}$ \\
\verb|MobileViT-XSmall| & - & - & ${72.6}$ & ${74.0}_{\scriptscriptstyle +1.4}$ & ${73.3}$ & $\bm{74.7}_{\scriptscriptstyle +1.3}$ \\
\verb|MobileViT-Small| & - & - & ${76.3}$ & ${78.3}_{\scriptscriptstyle +2.0}$ & ${76.7}$ & $\bm{78.6}_{\scriptscriptstyle +1.9}$ \\
\midrule
\verb|ResNet-18| & ${67.8}$ & ${71.9}_{\scriptscriptstyle +4.1}$ & ${69.9}$ & $\bm{73.2}_{\scriptscriptstyle +3.3}$ & ${69.8}$ & $\bm{73.2}_{\scriptscriptstyle +3.4}$ \\
\verb|ResNet-34| & ${73.2}$ & ${76.2}_{\scriptscriptstyle +3.0}$ & ${74.6}$ & $\bm{76.9}_{\scriptscriptstyle +2.3}$ & ${74.7}$ & $\bm{76.9}_{\scriptscriptstyle +2.3}$ \\
\verb|ResNet-50| & ${77.4}$ & ${79.6}_{\scriptscriptstyle +2.3}$ & ${79.0}$ & $\bm{80.3}_{\scriptscriptstyle +1.3}$ & ${79.1}$ & $\bm{80.3}_{\scriptscriptstyle +1.2}$ \\
\verb|ResNet-101| & ${79.8}$ & ${81.5}_{\scriptscriptstyle +1.7}$ & ${80.5}$ & $\bm{81.8}_{\scriptscriptstyle +1.3}$ & ${80.5}$ & $\bm{81.9}_{\scriptscriptstyle +1.3}$ \\
\verb|ResNet-152| & ${80.8}$ & ${82.0}_{\scriptscriptstyle +1.1}$ & ${81.3}$ & $\bm{82.2}_{\scriptscriptstyle +1.0}$ & ${81.3}$ & $\bm{82.3}_{\scriptscriptstyle +0.9}$ \\
\midrule
\verb|EfficientNet-B2| & ${77.9}$ & ${80.7}_{\scriptscriptstyle +2.7}$ & ${79.5}$ & $\bm{81.5}_{\scriptscriptstyle +2.1}$ & ${79.5}$ & $\bm{81.6}_{\scriptscriptstyle +2.1}$ \\
\verb|EfficientNet-B3| & ${79.3}$ & ${81.6}_{\scriptscriptstyle +2.3}$ & ${80.9}$ & $\bm{82.4}_{\scriptscriptstyle +1.6}$ & ${80.8}$ & $\bm{82.5}_{\scriptscriptstyle +1.6}$ \\
\verb|EfficientNet-B4| & ${79.4}$ & ${81.5}_{\scriptscriptstyle +2.1}$ & ${82.7}$ & $\bm{83.6}_{\scriptscriptstyle +1.0}$ & ${82.7}$ & $\bm{83.7}_{\scriptscriptstyle +1.0}$ \\
\midrule
\verb|ViT-Tiny| & ${71.5}$ & ${74.0}_{\scriptscriptstyle +2.6}$ & ${72.1}$ & $\bm{74.3}_{\scriptscriptstyle +2.2}$ & ${72.1}$ & $\bm{74.4}_{\scriptscriptstyle +2.3}$ \\
\verb|ViT-Small| & ${78.4}$ & $\bm{79.7}_{\scriptscriptstyle +1.2}$ & ${78.4}$ & $\bm{79.8}_{\scriptscriptstyle +1.4}$ & ${78.7}$ & $\bm{79.9}_{\scriptscriptstyle +1.2}$ \\
\verb|ViT-Base| & - & - & ${79.5}$ & $\bm{81.7}_{\scriptscriptstyle +2.3}$ & ${80.6}$ & $\bm{81.7}_{\scriptscriptstyle +1.1}$ \\
\midrule
\verb|ViT-384-Base| & - & - & ${80.5}$ & $\bm{83.0}_{\scriptscriptstyle +2.5}$ & ${81.9}$ & $\bm{83.1}_{\scriptscriptstyle +1.1}$ \\
\midrule
\verb|Swin-Tiny| & ${79.9}$ & $\bm{82.0}_{\scriptscriptstyle +2.1}$ & ${80.5}$ & $\bm{82.1}_{\scriptscriptstyle +1.6}$ & ${80.3}$ & $\bm{81.9}_{\scriptscriptstyle +1.6}$ \\
\verb|Swin-Small| & ${80.6}$ & ${82.9}_{\scriptscriptstyle +2.4}$ & ${82.2}$ & $\bm{83.6}_{\scriptscriptstyle +1.4}$ & ${81.9}$ & ${83.3}_{\scriptscriptstyle +1.4}$ \\
\verb|Swin-Base| & - & - & ${82.7}$ & $\bm{83.9}_{\scriptscriptstyle +1.2}$ & ${82.2}$ & $\bm{83.7}_{\scriptscriptstyle +1.4}$ \\
\midrule
\verb|Swin-384-Base| & - & - & ${82.6}$ & $\bm{83.2}_{\scriptscriptstyle +0.6}$ & ${82.4}$ & $\bm{83.0}_{\scriptscriptstyle +0.6}$ \\
\bottomrule[1.5pt]
\end{tabular}

        }
        \caption{150 epochs}
        \label{tab:imagenet_plus_cvnets_e150}
    \end{subtable}
    \vfill
    \begin{subtable}[b]{.79\columnwidth}
        \centering
        \resizebox{0.59\columnwidth}{!}{
            \begin{tabular}{l|cc|cc|cc}
\toprule
\multirow{2}{*}{\textbf{Model}} & \multicolumn{2}{c}{\textbf{Base Recipes (\cref{tab:imagenet_plus_full})}} & \multicolumn{2}{c}{\textbf{CVNets}} & \multicolumn{2}{c}{\textbf{CVNets-EMA}}\\
\cmidrule[1.25pt]{2-7}
& \textbf{\IN{}} & \textbf{\INp{}} & \textbf{\IN{}} & \textbf{\INp{}} & \textbf{\IN{}} & \textbf{\INp{}}\\
\midrule[1.25pt]
\verb|MobileNetV1-0.25| & ${55.7}$ & ${55.9}_{\scriptscriptstyle +0.3}$ & ${56.0}$ & $\bm{56.5}_{\scriptscriptstyle +0.6}$ & ${56.1}$ & $\bm{56.6}_{\scriptscriptstyle +0.5}$ \\
\verb|MobileNetV1-0.5| & ${67.1}$ & $\bm{68.0}_{\scriptscriptstyle +0.9}$ & ${67.0}$ & $\bm{68.0}_{\scriptscriptstyle +1.0}$ & ${67.0}$ & $\bm{68.1}_{\scriptscriptstyle +1.1}$ \\
\verb|MobileNetV1-1.0| & ${74.0}$ & $\bm{75.9}_{\scriptscriptstyle +1.8}$ & ${74.0}$ & $\bm{76.0}_{\scriptscriptstyle +2.0}$ & ${74.1}$ & $\bm{76.0}_{\scriptscriptstyle +1.9}$ \\
\midrule
\verb|MobileNetV2-0.25| & ${54.8}$ & ${54.9}_{\scriptscriptstyle +0.1}$ & $\bm{55.4}$ & ${55.1}_{\scriptscriptstyle -0.3}$ & $\bm{55.5}$ & ${55.1}_{\scriptscriptstyle -0.4}$ \\
\verb|MobileNetV2-0.5| & ${66.0}$ & $\bm{66.4}_{\scriptscriptstyle +0.5}$ & ${65.8}$ & $\bm{66.5}_{\scriptscriptstyle +0.7}$ & ${65.9}$ & $\bm{66.6}_{\scriptscriptstyle +0.7}$ \\
\verb|MobileNetV2-1.0| & ${73.3}$ & $\bm{74.5}_{\scriptscriptstyle +1.2}$ & ${73.7}$ & $\bm{74.5}_{\scriptscriptstyle +0.8}$ & ${73.7}$ & $\bm{74.5}_{\scriptscriptstyle +0.8}$ \\
\midrule
\verb|MobileNetV3-Small| & ${67.2}$ & ${68.2}_{\scriptscriptstyle +1.0}$ & ${67.4}$ & $\bm{68.6}_{\scriptscriptstyle +1.2}$ & ${67.4}$ & $\bm{68.5}_{\scriptscriptstyle +1.1}$ \\
\verb|MobileNetV3-Large| & ${74.9}$ & $\bm{77.0}_{\scriptscriptstyle +2.1}$ & ${74.9}$ & $\bm{77.2}_{\scriptscriptstyle +2.3}$ & ${75.1}$ & $\bm{77.2}_{\scriptscriptstyle +2.1}$ \\
\midrule
\verb|MobileViT-XXSmall| & - & - & ${67.5}$ & ${68.8}_{\scriptscriptstyle +1.3}$ & ${68.6}$ & $\bm{69.7}_{\scriptscriptstyle +1.1}$ \\
\verb|MobileViT-XSmall| & - & - & ${74.0}$ & ${75.6}_{\scriptscriptstyle +1.6}$ & ${74.9}$ & $\bm{76.3}_{\scriptscriptstyle +1.4}$ \\
\verb|MobileViT-Small| & - & - & ${77.3}$ & ${79.6}_{\scriptscriptstyle +2.3}$ & ${77.9}$ & $\bm{80.1}_{\scriptscriptstyle +2.2}$ \\
\midrule
\verb|ResNet-18| & ${68.7}$ & ${72.7}_{\scriptscriptstyle +4.0}$ & ${71.2}$ & $\bm{74.2}_{\scriptscriptstyle +3.0}$ & ${71.1}$ & $\bm{74.2}_{\scriptscriptstyle +3.1}$ \\
\verb|ResNet-34| & ${74.3}$ & ${77.2}_{\scriptscriptstyle +2.9}$ & ${75.6}$ & $\bm{77.8}_{\scriptscriptstyle +2.1}$ & ${75.6}$ & $\bm{77.8}_{\scriptscriptstyle +2.2}$ \\
\verb|ResNet-50| & ${78.8}$ & ${80.6}_{\scriptscriptstyle +1.8}$ & ${79.6}$ & $\bm{81.2}_{\scriptscriptstyle +1.6}$ & ${79.7}$ & $\bm{81.2}_{\scriptscriptstyle +1.6}$ \\
\verb|ResNet-101| & ${80.9}$ & ${82.3}_{\scriptscriptstyle +1.4}$ & ${81.3}$ & $\bm{82.6}_{\scriptscriptstyle +1.3}$ & ${81.3}$ & $\bm{82.7}_{\scriptscriptstyle +1.4}$ \\
\verb|ResNet-152| & ${81.5}$ & ${82.8}_{\scriptscriptstyle +1.3}$ & ${81.8}$ & $\bm{83.1}_{\scriptscriptstyle +1.3}$ & ${81.8}$ & $\bm{83.1}_{\scriptscriptstyle +1.3}$ \\
\midrule
\verb|EfficientNet-B2| & ${78.9}$ & ${81.2}_{\scriptscriptstyle +2.2}$ & ${80.7}$ & $\bm{82.1}_{\scriptscriptstyle +1.4}$ & ${80.8}$ & $\bm{82.1}_{\scriptscriptstyle +1.3}$ \\
\verb|EfficientNet-B3| & ${80.1}$ & ${82.1}_{\scriptscriptstyle +2.0}$ & ${81.8}$ & $\bm{83.3}_{\scriptscriptstyle +1.5}$ & ${81.8}$ & $\bm{83.3}_{\scriptscriptstyle +1.5}$ \\
\verb|EfficientNet-B4| & ${80.6}$ & ${82.3}_{\scriptscriptstyle +1.7}$ & ${82.8}$ & $\bm{84.4}_{\scriptscriptstyle +1.6}$ & ${82.7}$ & $\bm{84.4}_{\scriptscriptstyle +1.7}$ \\
\midrule
\verb|ViT-Tiny| & ${74.1}$ & $\bm{75.8}_{\scriptscriptstyle +1.7}$ & ${74.8}$ & $\bm{76.0}_{\scriptscriptstyle +1.2}$ & ${74.9}$ & $\bm{76.0}_{\scriptscriptstyle +1.1}$ \\
\verb|ViT-Small| & ${78.9}$ & $\bm{81.4}_{\scriptscriptstyle +2.4}$ & ${79.1}$ & $\bm{81.4}_{\scriptscriptstyle +2.3}$ & ${79.9}$ & $\bm{81.5}_{\scriptscriptstyle +1.5}$ \\
\verb|ViT-Base| & - & - & ${78.6}$ & $\bm{84.1}_{\scriptscriptstyle +5.5}$ & ${81.0}$ & $\bm{84.1}_{\scriptscriptstyle +3.1}$ \\
\midrule
\verb|ViT-384-Base| & - & - & ${80.0}$ & $\bm{84.5}_{\scriptscriptstyle +4.6}$ & ${82.6}$ & $\bm{84.5}_{\scriptscriptstyle +1.9}$ \\
\midrule
\verb|Swin-Tiny| & ${80.9}$ & $\bm{83.0}_{\scriptscriptstyle +2.1}$ & ${81.2}$ & $\bm{83.2}_{\scriptscriptstyle +2.1}$ & ${80.8}$ & ${82.8}_{\scriptscriptstyle +2.0}$ \\
\verb|Swin-Small| & ${81.4}$ & ${83.9}_{\scriptscriptstyle +2.5}$ & ${82.4}$ & $\bm{84.4}_{\scriptscriptstyle +2.0}$ & ${82.2}$ & ${84.1}_{\scriptscriptstyle +1.9}$ \\
\verb|Swin-Base| & - & - & ${82.7}$ & $\bm{84.7}_{\scriptscriptstyle +2.0}$ & ${82.5}$ & ${84.4}_{\scriptscriptstyle +1.9}$ \\
\midrule
\verb|Swin-384-Base| & - & - & ${83.9}$ & $\bm{84.4}_{\scriptscriptstyle +0.5}$ & ${83.7}$ & $\bm{84.2}_{\scriptscriptstyle +0.5}$ \\
\bottomrule[1.5pt]
\end{tabular}

        }
        \caption{300 epochs}
        \label{tab:imagenet_plus_cvnets_e300}
    \end{subtable}
    \vfill
    \begin{subtable}[b]{.79\columnwidth}
        \centering
        \resizebox{0.59\columnwidth}{!}{
            \begin{tabular}{l|cc|cc|cc}
\toprule
\multirow{2}{*}{\textbf{Model}} & \multicolumn{2}{c}{\textbf{Base Recipes (\cref{tab:imagenet_plus_full})}} & \multicolumn{2}{c}{\textbf{CVNets}} & \multicolumn{2}{c}{\textbf{CVNets-EMA}}\\
\cmidrule[1.25pt]{2-7}
& \textbf{\IN{}} & \textbf{\INp{}} & \textbf{\IN{}} & \textbf{\INp{}} & \textbf{\IN{}} & \textbf{\INp{}}\\
\midrule[1.25pt]
\verb|MobileNetV1-0.25| & ${56.7}$ & $\bm{57.1}_{\scriptscriptstyle +0.4}$ & $\bm{56.9}$ & $\bm{57.1}_{\scriptscriptstyle +0.2}$ & $\bm{56.9}$ & $\bm{57.1}_{\scriptscriptstyle +0.2}$ \\
\verb|MobileNetV1-0.5| & ${67.8}$ & ${68.5}_{\scriptscriptstyle +0.7}$ & ${68.1}$ & $\bm{68.7}_{\scriptscriptstyle +0.6}$ & ${68.1}$ & $\bm{68.8}_{\scriptscriptstyle +0.7}$ \\
\verb|MobileNetV1-1.0| & ${74.1}$ & $\bm{76.7}_{\scriptscriptstyle +2.5}$ & ${74.1}$ & $\bm{76.8}_{\scriptscriptstyle +2.7}$ & ${74.4}$ & $\bm{76.8}_{\scriptscriptstyle +2.4}$ \\
\midrule
\verb|MobileNetV2-0.25| & $\bm{55.7}$ & ${55.2}_{\scriptscriptstyle -0.5}$ & $\bm{55.7}$ & $\bm{55.7}_{\scriptscriptstyle -0.0}$ & $\bm{55.8}$ & $\bm{55.7}_{\scriptscriptstyle -0.1}$ \\
\verb|MobileNetV2-0.5| & ${66.8}$ & $\bm{67.1}_{\scriptscriptstyle +0.3}$ & ${66.8}$ & $\bm{67.1}_{\scriptscriptstyle +0.3}$ & ${66.8}$ & $\bm{67.2}_{\scriptscriptstyle +0.4}$ \\
\verb|MobileNetV2-1.0| & ${73.9}$ & $\bm{75.5}_{\scriptscriptstyle +1.6}$ & ${74.0}$ & $\bm{75.5}_{\scriptscriptstyle +1.5}$ & ${74.1}$ & $\bm{75.6}_{\scriptscriptstyle +1.5}$ \\
\midrule
\verb|MobileNetV3-Small| & ${67.9}$ & $\bm{69.4}_{\scriptscriptstyle +1.4}$ & ${68.1}$ & $\bm{69.6}_{\scriptscriptstyle +1.5}$ & ${68.1}$ & $\bm{69.6}_{\scriptscriptstyle +1.5}$ \\
\verb|MobileNetV3-Large| & ${75.1}$ & $\bm{77.9}_{\scriptscriptstyle +2.9}$ & ${74.8}$ & $\bm{77.9}_{\scriptscriptstyle +3.1}$ & ${75.8}$ & $\bm{77.9}_{\scriptscriptstyle +2.1}$ \\
\midrule
\verb|MobileViT-XXSmall| & - & - & ${68.6}$ & ${69.5}_{\scriptscriptstyle +0.9}$ & ${70.3}$ & $\bm{71.5}_{\scriptscriptstyle +1.2}$ \\
\verb|MobileViT-XSmall| & - & - & ${74.8}$ & ${76.2}_{\scriptscriptstyle +1.4}$ & ${76.1}$ & $\bm{77.5}_{\scriptscriptstyle +1.4}$ \\
\verb|MobileViT-Small| & - & - & ${77.7}$ & ${80.5}_{\scriptscriptstyle +2.8}$ & ${79.2}$ & $\bm{81.4}_{\scriptscriptstyle +2.2}$ \\
\midrule
\verb|ResNet-18| & ${69.9}$ & ${73.8}_{\scriptscriptstyle +3.8}$ & ${72.3}$ & $\bm{75.0}_{\scriptscriptstyle +2.7}$ & ${72.2}$ & $\bm{75.1}_{\scriptscriptstyle +2.9}$ \\
\verb|ResNet-34| & ${75.6}$ & ${78.4}_{\scriptscriptstyle +2.8}$ & ${76.6}$ & $\bm{78.8}_{\scriptscriptstyle +2.2}$ & ${76.6}$ & $\bm{78.9}_{\scriptscriptstyle +2.3}$ \\
\verb|ResNet-50| & ${79.6}$ & ${81.7}_{\scriptscriptstyle +2.1}$ & ${80.0}$ & $\bm{82.0}_{\scriptscriptstyle +1.9}$ & ${80.1}$ & $\bm{82.0}_{\scriptscriptstyle +1.9}$ \\
\verb|ResNet-101| & ${81.4}$ & ${83.2}_{\scriptscriptstyle +1.8}$ & ${80.9}$ & $\bm{83.5}_{\scriptscriptstyle +2.6}$ & ${81.4}$ & $\bm{83.5}_{\scriptscriptstyle +2.1}$ \\
\verb|ResNet-152| & ${81.7}$ & $\bm{83.8}_{\scriptscriptstyle +2.2}$ & ${81.3}$ & $\bm{83.9}_{\scriptscriptstyle +2.6}$ & ${82.0}$ & $\bm{83.9}_{\scriptscriptstyle +2.0}$ \\
\midrule
\verb|EfficientNet-B2| & ${79.3}$ & ${81.9}_{\scriptscriptstyle +2.7}$ & ${81.1}$ & $\bm{83.1}_{\scriptscriptstyle +2.0}$ & ${81.3}$ & $\bm{83.2}_{\scriptscriptstyle +1.9}$ \\
\verb|EfficientNet-B3| & ${79.6}$ & ${82.8}_{\scriptscriptstyle +3.3}$ & ${81.7}$ & $\bm{83.9}_{\scriptscriptstyle +2.2}$ & ${82.1}$ & $\bm{83.9}_{\scriptscriptstyle +1.8}$ \\
\verb|EfficientNet-B4| & ${81.2}$ & ${83.1}_{\scriptscriptstyle +1.9}$ & ${82.2}$ & $\bm{85.0}_{\scriptscriptstyle +2.9}$ & ${83.4}$ & $\bm{85.0}_{\scriptscriptstyle +1.6}$ \\
\midrule
\verb|ViT-Tiny| & ${75.9}$ & $\bm{78.1}_{\scriptscriptstyle +2.1}$ & ${76.7}$ & $\bm{77.9}_{\scriptscriptstyle +1.2}$ & ${76.9}$ & $\bm{78.0}_{\scriptscriptstyle +1.1}$ \\
\verb|ViT-Small| & ${78.4}$ & ${82.6}_{\scriptscriptstyle +4.2}$ & ${78.5}$ & $\bm{82.8}_{\scriptscriptstyle +4.2}$ & ${80.6}$ & $\bm{82.9}_{\scriptscriptstyle +2.3}$ \\
\verb|ViT-Base| & - & - & ${76.8}$ & $\bm{85.1}_{\scriptscriptstyle +8.3}$ & ${80.8}$ & $\bm{85.1}_{\scriptscriptstyle +4.3}$ \\
\midrule
\verb|ViT-384-Base| & - & - & ${79.4}$ & $\bm{85.4}_{\scriptscriptstyle +6.0}$ & ${83.1}$ & $\bm{85.5}_{\scriptscriptstyle +2.5}$ \\
\midrule
\verb|Swin-Tiny| & ${80.9}$ & $\bm{83.8}_{\scriptscriptstyle +2.8}$ & ${81.3}$ & $\bm{84.0}_{\scriptscriptstyle +2.7}$ & ${80.5}$ & ${83.5}_{\scriptscriptstyle +3.0}$ \\
\verb|Swin-Small| & ${81.9}$ & ${84.4}_{\scriptscriptstyle +2.5}$ & ${81.3}$ & $\bm{85.0}_{\scriptscriptstyle +3.7}$ & ${81.9}$ & ${84.5}_{\scriptscriptstyle +2.6}$ \\
\verb|Swin-Base| & - & - & ${81.5}$ & $\bm{85.4}_{\scriptscriptstyle +3.9}$ & ${81.8}$ & $\bm{85.2}_{\scriptscriptstyle +3.5}$ \\
\midrule
\verb|Swin-384-Base| & - & - & ${83.6}$ & $\bm{85.8}_{\scriptscriptstyle +2.2}$ & ${83.8}$ & ${85.5}_{\scriptscriptstyle +1.7}$ \\
\bottomrule[1.5pt]
\end{tabular}

        }
        \caption{1000 epochs}
        \label{tab:imagenet_plus_cvnets_e1000}
    \end{subtable}
    \caption{\textbf{Improved results with state-of-the-art training recepies 
    in CVNets library.} Subscripts show the improvement on top of the \IN{} 
    accuracy.  We highlight the best accuracy on each row from our proposed 
    datasets and any number that is within $0.2$ of the best.
    }
    \label{tab:imagenet_plus_cvnets}
\end{table*}

\subsection{Aggregated improvements of \INp{} across models}
\begin{figure*}[thb!]
    \centering
    \begin{subfigure}[t]{0.48\textwidth}
        \centering
        \includegraphics[width=0.8\textwidth]{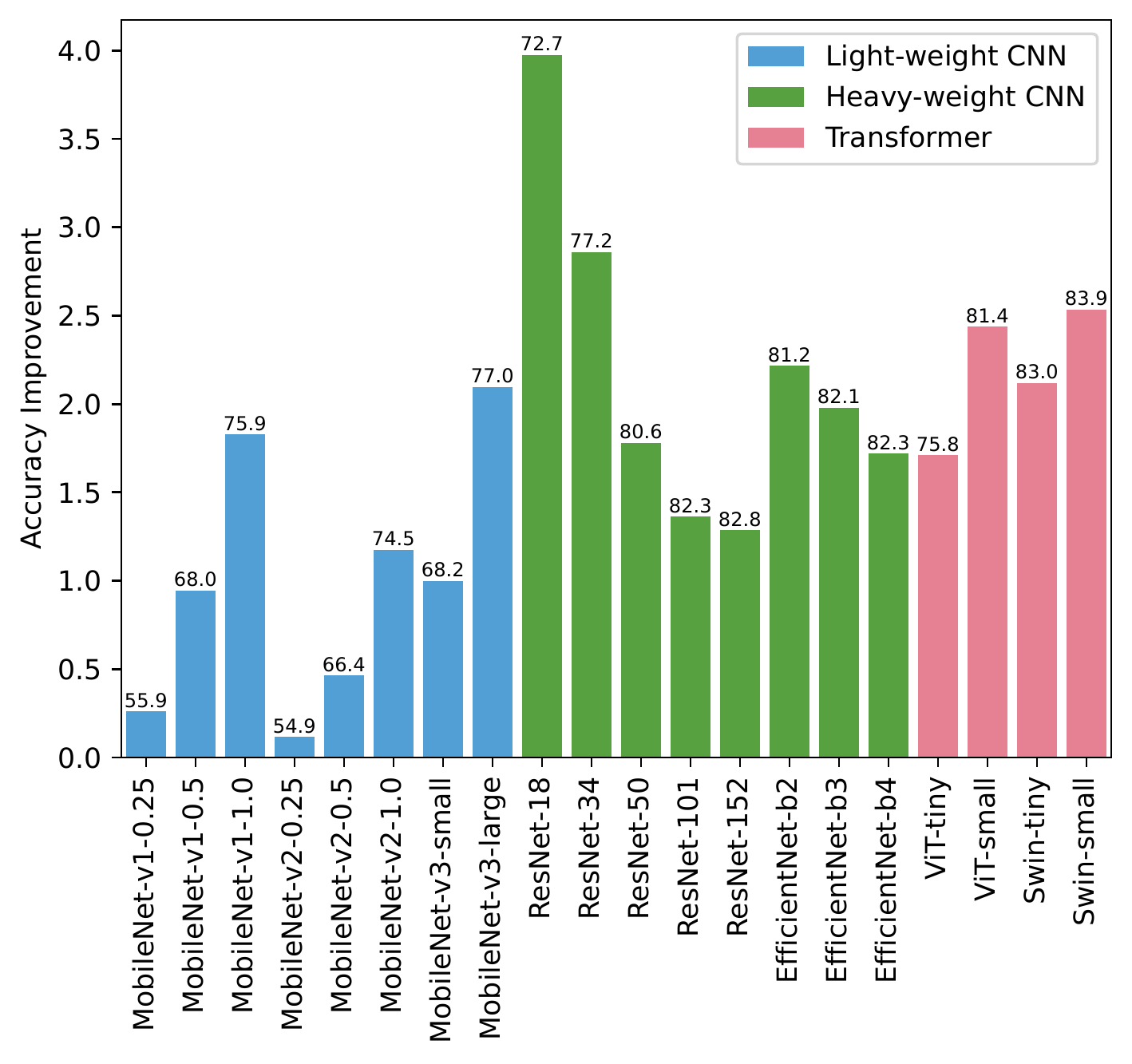}
        \caption{\INp{} (RRC+RA/RE)}
        \label{fig:figs_delta_ra_re_300}
        \vspace{-6mm}
    \end{subfigure}
    \begin{subfigure}[t]{0.48\textwidth}
        \centering
        \includegraphics[width=0.8\textwidth]{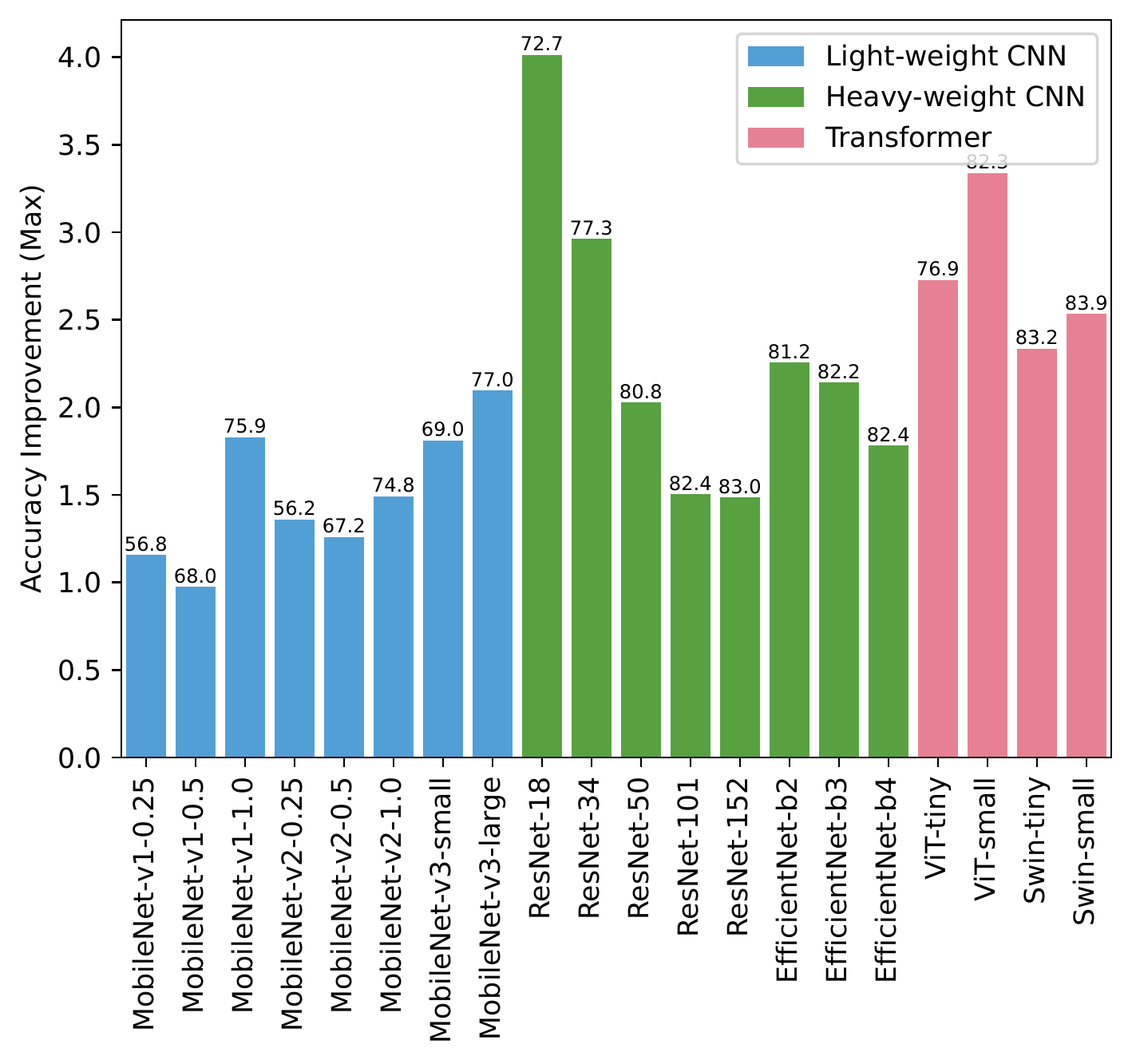}
        \caption{Maximum of 4 \INp{} variants}
        \label{fig:figs_delta_max}
    \end{subfigure}
    \vspace{-4mm}
    \centering
    \caption{\textbf{\INp{} training improves validation accuracy} compared 
        with \IN{} training (Epochs=$300$).
    To train models using the \INp{} dataset, we use the same publicly-available \IN{} training recipes with \emph{no hyperparameter tuning} on \INp{}.
    We use the same hyperparameters 
    tuned for \IN{} with \emph{no hyperparameter tuning} on \INp{}.  
    \INp{} provides a balanced tradeoff with more improvements for 
    Heavy-weight CNNs and Transformers.
    \Cref{fig:figs_delta_max}:
    further improvements are achieved using the best out of our 4 proposed 
    datasets (See \cref{tab:imagenet_plus_table6_e300} for details).}
\end{figure*}

To better demonstrate the scale of accuracy improvements, we plot the results 
of training on \INp{} (\RRCRARE{}) from \cref{tab:imagenet_plus_full} in 
\cref{fig:figs_delta_ra_re_300} (300 epochs).
\RRCRARE{} balances the tradeoff between various architectures.
Given prior knowledge of architecture characteristics or enough training 
resources, we can select the dataset optimal for any architecture.
\Cref{fig:figs_delta_max} shows the best accuracy achieved for each model when 
we train on all 4 of our reinforced datasets for 300 epochs (maximum of the 
four numbers).  We observe that alternative reinforced datasets can provide 
1-2\% additional improvement, especially for light-weight CNNs and 
Transformers. In practice, given the knowledge of the complexity of the model 
architecture, one can decide to use alternative datasets (\RRC{} for 
light-weight and \RRCMsRs for heavy-weight models or Transformers). Otherwise, 
given additional compute resources, one can train on all 4 datasets and choose 
the best model according to the validation accuracy.

\newpage
\clearpage
\section{Expanded study on what is a good teacher?}
In this section we provide additional results and studies such as super 
ensembles as teachers.

\subsection{Additional results of knowledge distillation with pretrained Timm 
models}\label{sec:timm_distill_appendix}
\begin{figure*}[thb!]
\begin{center}
    \begin{subfigure}[t]{0.27\textwidth}
        \includegraphics[width=\textwidth]{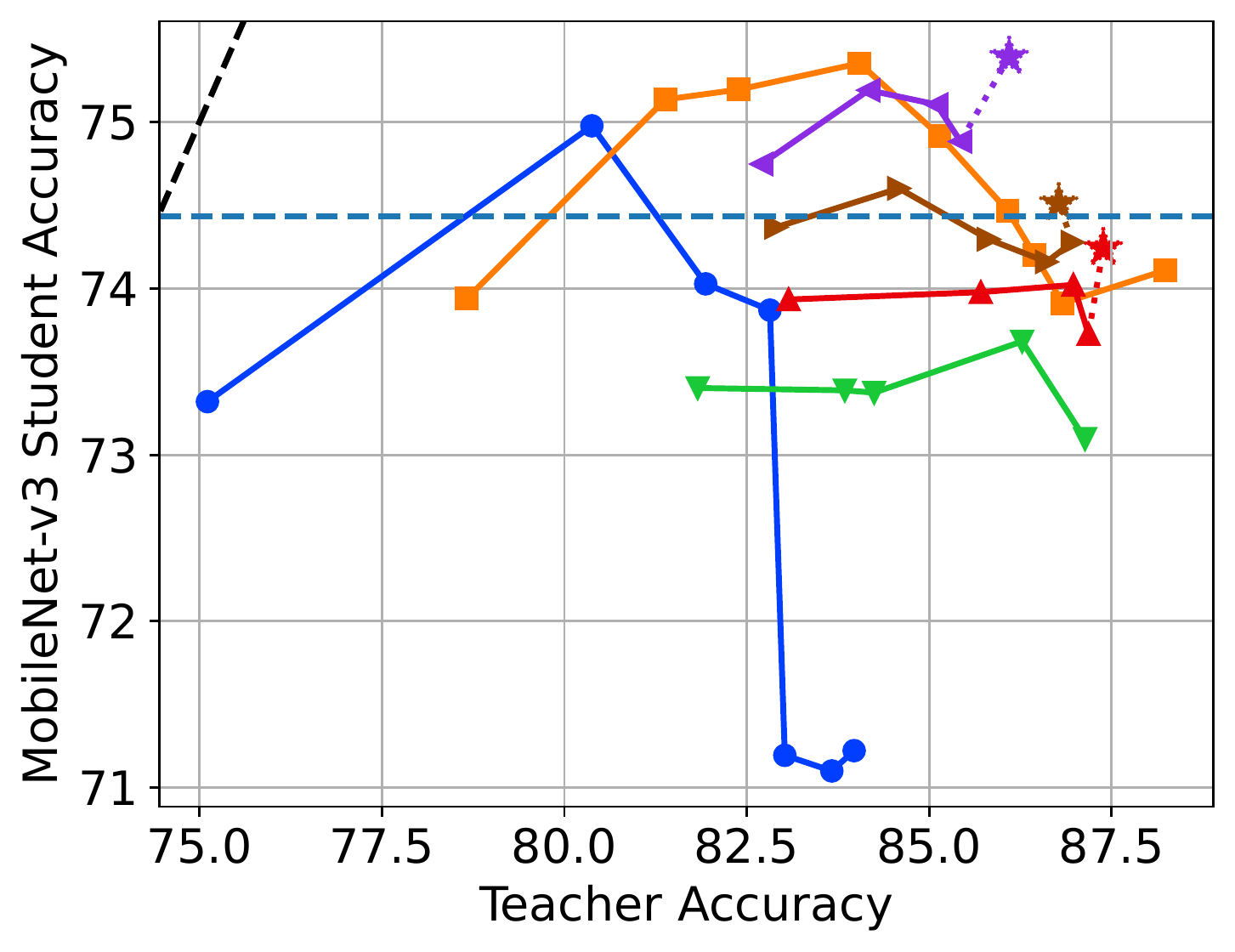}
        \caption{MobileNetV3}\label{fig:imagenet_MobileNetv3_distill_e150}
    \end{subfigure}
    \begin{subfigure}[t]{0.27\textwidth}
        \includegraphics[width=\textwidth]{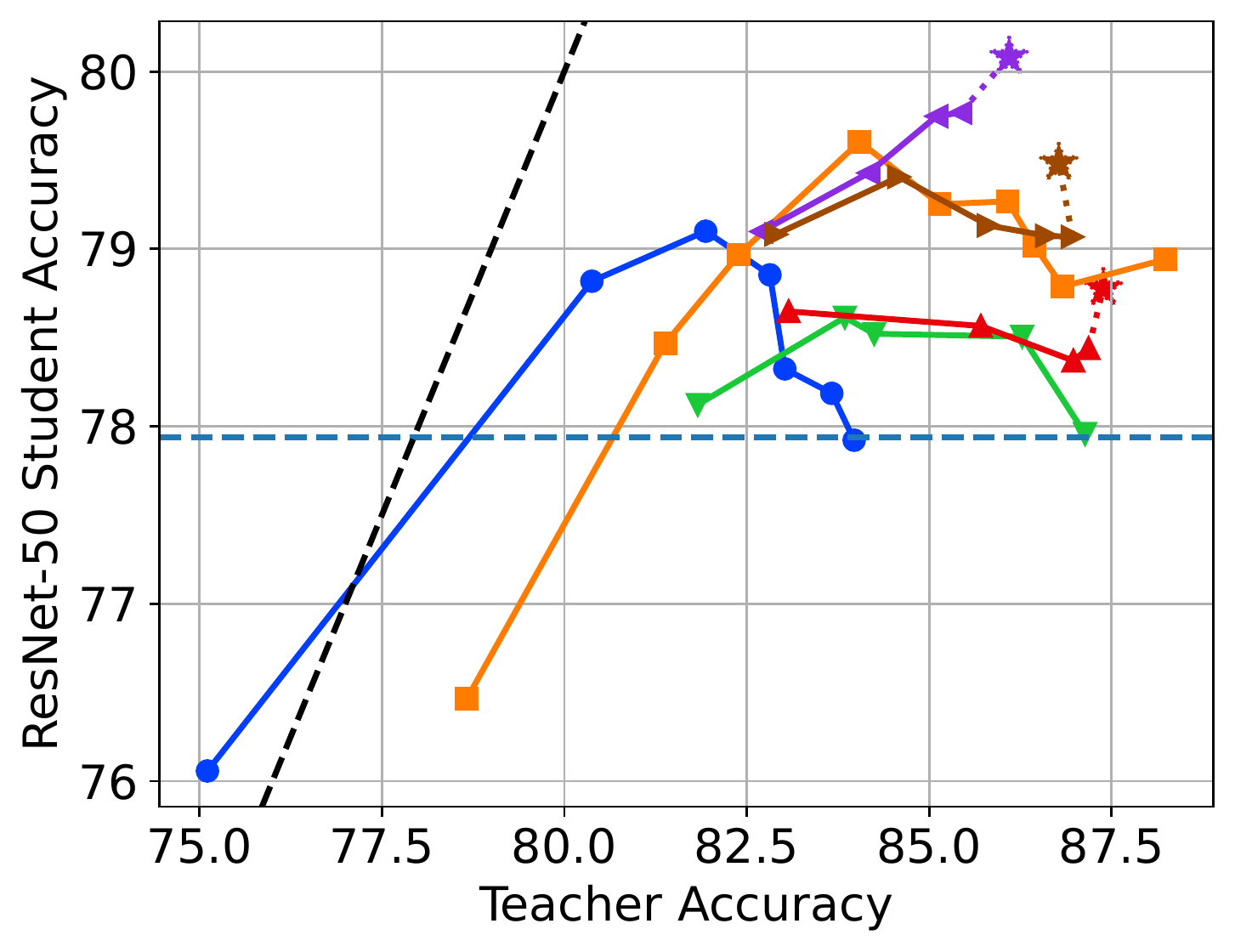}
        \caption{ResNet50}\label{fig:imagenet_R50_distill_e150}
    \end{subfigure}
    \hfill
    \begin{subfigure}[t]{0.27\textwidth}
        \includegraphics[width=\textwidth]{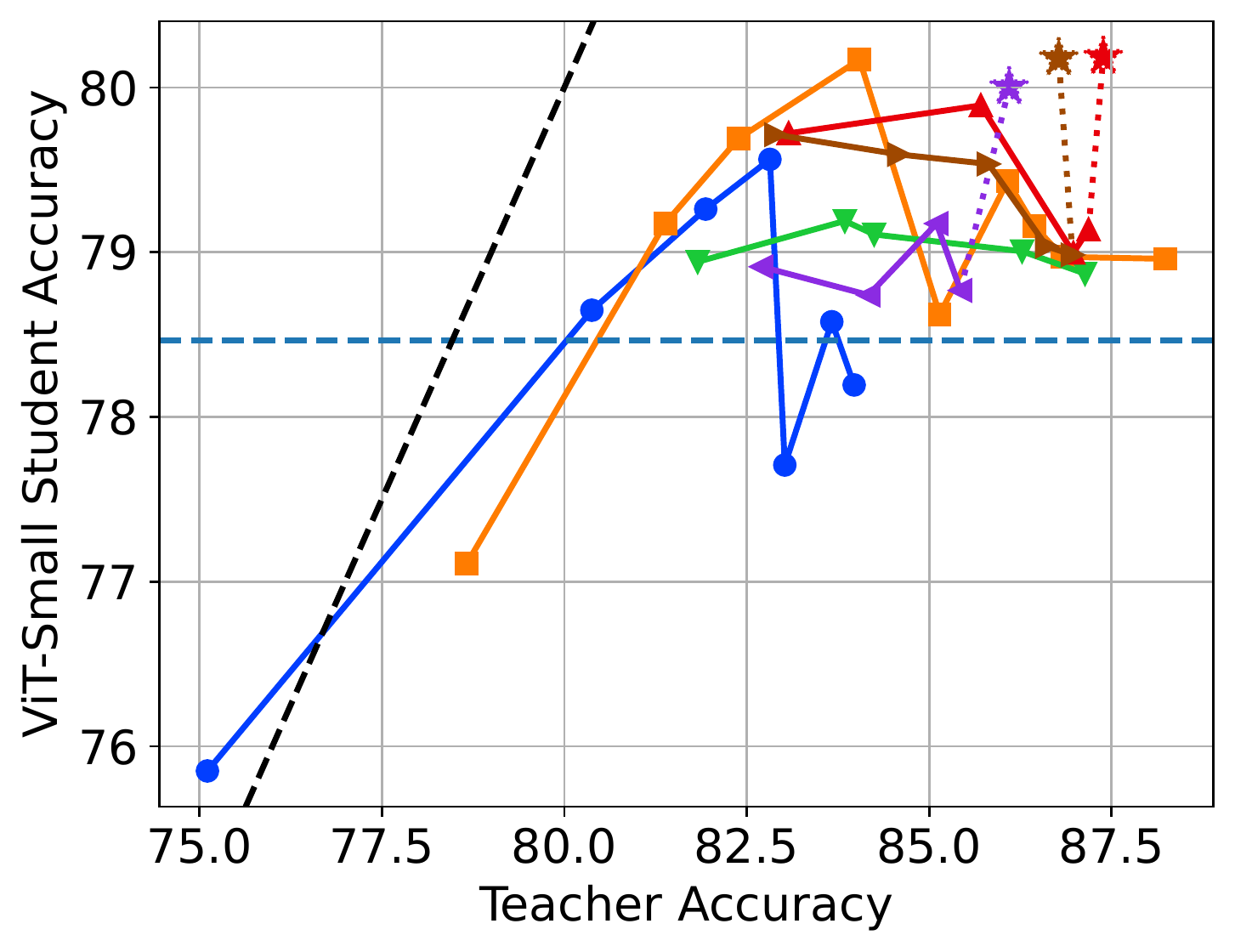}
        \caption{ViT-Small}\label{fig:imagenet_ViTSmall_distill_e150}
    \end{subfigure}
    \hfill
    \hfill
    \begin{minipage}[t]{0.17\linewidth}
        \vspace{-3.5cm}
        \begin{subfigure}[t]{\textwidth}
            \includegraphics[width=\textwidth]{figures/figs_distill/legend.pdf}
        \end{subfigure}
    \end{minipage}
\end{center}
    \vspace{-0.5cm}
    \caption{Knowledge distillation accuracy of representative student 
    architectures (ResNet-50, ViT-Small, MobileNetV3) for pretrained teachers 
    from Timm library. We train for 150 epochs and \cref{fig:timm_distill_e300} 
    shows results for 300 epoch training.}\label{fig:timm_distill_e150}
\end{figure*}

\Cref{fig:timm_distill_e150} (E=150) complements \cref{fig:timm_distill_e300} 
(E=300) demonstrating the validation accuracy using knowledge distillation for 
a variety of teachers from the Timm library~\citep{rw2019timm}.
\Cref{tab:timm_distill_all} shows the results in detail. For both 150 and 300 
epoch training durations, we observe that ensembles of the state-of-the-art 
models in the Timm library perform best as the teachers across different 
student architectures. We choose the IG-ResNext ensemble for dataset 
reinforcement.

\begin{table*}[htb!]
\centering
\resizebox{0.7\textwidth}{!}{
\begin{tabular}{p{9cm}l ll ll ll}
\toprule[1.5pt]
Teacher Name & Teacher Accuracy & \multicolumn{6}{c}{\bfseries Student Top-1 
    Accuracy} \\
\cmidrule[1.25pt]{3-8}
    &  & \multicolumn{2}{c}{\bfseries ResNet-50} & \multicolumn{2}{c}{\bfseries 
    ViT-Small} & \multicolumn{2}{c}{\bfseries MobileNetv3-Large}\\ 
    \cmidrule[1.1pt]{3-8}
    &  & 300 & 150 & 300 & 150 & 300 & 150 \\
    \midrule
\verb|beit_large_patch16_512| &  $88.58$ & $41.76$ & $41.45$ & $37.03$ & $35.87$ & --- & --- \\
\verb|convnext_tiny_in22ft1k| &  $82.90$ & $80.13$ & $79.08$ & $80.82$ & $79.71$ & $75.41$ & $74.37$ \\
\verb|convnext_large| \scalebox{1.25}{+}  \verb|convnext_base| \scalebox{1.25}{+}  \verb|convnext_small| \scalebox{1.25}{+}  \verb|convnext_tiny| \scalebox{1.25}{+}  \verb|convnext_nano| &  $84.34$ & $80.34$ & $78.94$ & $80.73$ & $79.82$ & --- & --- \\
\verb|convnext_small_in22ft1k| &  $84.59$ & $80.63$ & $79.41$ & $81.16$ & $79.59$ & $75.70$ & $74.60$ \\
\verb|convnext_base_in22ft1k| &  $85.81$ & $80.56$ & $79.13$ & $81.36$ & $79.54$ & $75.44$ & $74.30$ \\
\verb|convnext_large_in22ft1k| &  $86.61$ & $80.17$ & $79.07$ & $80.77$ & $79.04$ & $75.35$ & $74.16$ \\
\verb|convnext_xlarge_in22ft1k| \scalebox{1.25}{+}  \verb|convnext_large_in22ft1k| \scalebox{1.25}{+}  \verb|convnext_base_in22ft1k| \scalebox{1.25}{+}  \verb|convnext_small_in22ft1k| \scalebox{1.25}{+}  \verb|convnext_tiny_in22ft1k| &  $86.78$ & $80.78$ & $79.48$ & $81.67$ & $80.18$ & $75.82$ & $74.51$ \\
\verb|convnext_xlarge_in22ft1k| &  $86.96$ & $80.07$ & $79.07$ & $80.69$ & $78.98$ & $75.26$ & $74.28$ \\
\verb|convnext_xlarge_384_in22ft1k| &  $87.53$ & $79.67$ & $78.50$ & $79.92$ & $78.38$ & --- & --- \\
\verb|deit3_small_patch16_224_in21ft1k| &  $83.07$ & $79.60$ & $78.65$ & $81.23$ & $79.72$ & $75.04$ & $73.93$ \\
\verb|deit3_huge_patch14_224| \scalebox{1.25}{+}  \verb|deit3_large_patch16_224| \scalebox{1.25}{+}  \verb|deit3_base_patch16_224| \scalebox{1.25}{+}  \verb|deit3_small_patch16_224| &  $85.30$ & $79.69$ & $78.82$ & $80.25$ & $79.55$ & --- & --- \\
\verb|deit3_base_patch16_224_in21ft1k| &  $85.71$ & $79.98$ & $78.57$ & $81.24$ & $79.89$ & $75.16$ & $73.98$ \\
\verb|deit3_large_patch16_224_in21ft1k| &  $86.98$ & $79.61$ & $78.37$ & $80.81$ & $79.00$ & $75.08$ & $74.02$ \\
\verb|deit3_huge_patch14_224_in21ft1k| &  $87.18$ & $79.52$ & $78.44$ & $80.59$ & $79.14$ & $74.79$ & $73.73$ \\
\verb|deit3_huge_patch14_224_in21ft1k| \scalebox{1.25}{+}  \verb|deit3_large_patch16_224_in21ft1k| \scalebox{1.25}{+}  \verb|deit3_base_patch16_224_in21ft1k| \scalebox{1.25}{+}  \verb|deit3_small_patch16_224_in21ft1k| &  $87.39$ & $80.25$ & $78.78$ & $81.26$ & $80.19$ & $75.28$ & $74.24$ \\
\verb|deit3_large_patch16_384_in21ft1k| &  $87.73$ & $79.06$ & $78.00$ & $79.71$ & $78.80$ & --- & --- \\
\verb|ig_resnext101_32x8d| &  $82.70$ & $80.32$ & $79.10$ & $80.30$ & $78.91$ & $76.03$ & $74.75$ \\
\verb|ig_resnext101_32x16d| &  $84.17$ & $80.78$ & $79.43$ & $80.93$ & $78.74$ & $76.37$ & $75.19$ \\
\verb|ig_resnext101_32x32d| &  $85.10$ & $81.04$ & $79.75$ & $81.03$ & $79.17$ & $76.42$ & $75.11$ \\
\verb|ig_resnext101_32x48d| &  $85.43$ & $80.93$ & $79.77$ & $80.67$ & $78.77$ & $76.01$ & $74.88$ \\
\verb|ig_resnext101_32x48d| \scalebox{1.25}{+}  \verb|ig_resnext101_32x32d| \scalebox{1.25}{+}  \verb|ig_resnext101_32x16d| \scalebox{1.25}{+}  \verb|ig_resnext101_32x8d| &  $86.10$ & $81.30$ & $80.08$ & $82.01$ & $80.00$ & $76.70$ & $75.39$ \\
\verb|ig_resnext101_32x48d| \scalebox{1.25}{+}  \verb|convnext_xlarge_in22ft1k| \scalebox{1.25}{+}  \verb|volo_d5_224| \scalebox{1.25}{+}  \verb|deit3_huge_patch14_224| &  $87.39$ & $80.71$ & $79.63$ & $81.65$ & $80.00$ & --- & --- \\
\verb|resnet18| &  $69.74$ & $71.29$ & $71.29$ & $71.29$ & $71.18$ & --- & --- \\
\verb|resnet34| &  $75.11$ & $76.46$ & $76.06$ & $76.36$ & $75.85$ & $73.90$ & $73.32$ \\
\verb|resnet50| &  $80.38$ & $79.83$ & $78.82$ & $79.81$ & $78.65$ & $75.63$ & $74.98$ \\
\verb|resnet101| &  $81.94$ & $80.07$ & $79.10$ & $80.82$ & $79.26$ & $74.92$ & $74.03$ \\
\verb|resnet152| &  $82.82$ & $79.88$ & $78.85$ & $79.72$ & $79.56$ & $74.82$ & $73.87$ \\
\verb|resnet101d| &  $83.02$ & $79.75$ & $78.32$ & $78.92$ & $77.71$ & $72.78$ & $71.19$ \\
\verb|resnet152d| &  $83.67$ & $79.62$ & $78.19$ & $78.77$ & $78.58$ & $73.05$ & $71.10$ \\
\verb|resnet200d| &  $83.97$ & $79.40$ & $77.92$ & $80.18$ & $78.19$ & $73.10$ & $71.22$ \\
\verb|resnetv2_152x2_bitm| &  $84.46$ & $79.57$ & $78.63$ & $80.04$ & $78.52$ & $75.12$ & $73.87$ \\
\verb|resnetv2_152x4_bitm| &  $84.94$ & --- & $78.09$ & --- & $78.97$ & --- & $73.69$ \\
\verb|swinv2_cr_tiny_ns_224| &  $81.79$ & $79.41$ & $78.47$ & $80.56$ & $79.29$ & $74.23$ & $73.43$ \\
\verb|swinv2_tiny_window8_256| &  $81.83$ & $79.30$ & $78.12$ & $80.32$ & $78.94$ & $74.46$ & $73.40$ \\
\verb|swinv2_tiny_window16_256| &  $82.82$ & $79.43$ & $78.31$ & $80.59$ & $79.30$ & $74.31$ & $73.57$ \\
\verb|swinv2_cr_small_224| &  $83.12$ & $79.15$ & $78.33$ & $79.78$ & $78.64$ & $74.59$ & $73.42$ \\
\verb|swinv2_cr_small_ns_224| &  $83.48$ & $79.43$ & $78.62$ & $80.38$ & $78.94$ & $74.56$ & $73.62$ \\
\verb|swinv2_small_window8_256| &  $83.84$ & $79.66$ & $78.61$ & $80.35$ & $79.19$ & $74.44$ & $73.39$ \\
\verb|swinv2_small_window16_256| &  $84.22$ & $79.21$ & $78.41$ & $80.00$ & $78.30$ & $74.84$ & $73.45$ \\
\verb|swinv2_base_window8_256| &  $84.25$ & $79.57$ & $78.52$ & $80.32$ & $79.11$ & $74.73$ & $73.37$ \\
\verb|swinv2_base_window16_256| &  $84.59$ & $78.94$ & $78.30$ & $79.50$ & $78.32$ & $74.50$ & $73.63$ \\
\verb|swinv2_base_window12to16_192to256_22kft1k| &  $86.27$ & $79.76$ & $78.51$ & $80.80$ & $79.01$ & $74.35$ & $73.68$ \\
\verb|swinv2_large_window12to16_192to256_22kft1k| &  $86.94$ & $79.31$ & $78.38$ & $79.86$ & $78.43$ & $74.39$ & $73.51$ \\
\verb|swinv2_base_window12to24_192to384_22kft1k| &  $87.14$ & $79.18$ & $77.96$ & $80.01$ & $78.87$ & --- & $73.10$ \\
\verb|swinv2_large_window12to24_192to384_22kft1k| &  $87.47$ & $78.48$ & $77.65$ & $78.33$ & $78.38$ & --- & $73.07$ \\
\verb|tf_efficientnet_b0| &  $76.85$ & --- & $75.10$ & --- & $75.97$ & $73.68$ & $72.93$ \\
\verb|tf_efficientnet_b0_ns| &  $78.67$ & $77.27$ & $76.47$ & $77.96$ & $77.11$ & $74.94$ & $73.94$ \\
\verb|tf_efficientnet_b1| &  $78.83$ & --- & $77.45$ & --- & $77.97$ & $75.48$ & $74.69$ \\
\verb|tf_efficientnet_b2| &  $80.08$ & --- & $78.17$ & --- & $78.88$ & $75.97$ & $75.01$ \\
\verb|tf_efficientnet_b1_ns| &  $81.38$ & $79.52$ & $78.47$ & $80.38$ & $79.18$ & $76.14$ & $75.14$ \\
\verb|tf_efficientnet_b3| &  $81.65$ & --- & $78.88$ & --- & $79.56$ & $76.39$ & $75.38$ \\
\verb|tf_efficientnet_b2_ns| &  $82.39$ & $80.17$ & $78.97$ & $80.94$ & $79.69$ & $76.43$ & $75.20$ \\
\verb|tf_efficientnet_b4| &  $83.03$ & --- & $78.91$ & --- & $78.56$ & $75.64$ & $74.65$ \\
\verb|tf_efficientnet_b5| &  $83.81$ & --- & $79.20$ & --- & $79.18$ & $76.01$ & $75.06$ \\
\verb|tf_efficientnet_b3_ns| &  $84.05$ & $80.71$ & $79.60$ & $81.72$ & $80.17$ & $76.44$ & $75.35$ \\
\verb|tf_efficientnet_b6| &  $84.11$ & --- & $78.92$ & --- & $79.21$ & $75.58$ & $74.41$ \\
\verb|tf_efficientnet_b7| &  $84.93$ & --- & $79.16$ & --- & $79.24$ & $75.42$ & $74.43$ \\
\verb|tf_efficientnet_b4_ns| &  $85.14$ & $80.83$ & $79.25$ & $81.51$ & $78.62$ & $75.81$ & $74.92$ \\
\verb|tf_efficientnet_b8| &  $85.35$ & --- & $78.84$ & --- & $78.17$ & $75.15$ & $73.86$ \\
\verb|tf_efficientnet_b5_ns| &  $86.08$ & $80.71$ & $79.27$ & $81.05$ & $79.43$ & $75.57$ & $74.47$ \\
\verb|tf_efficientnetv2_xl_in21ft1k| &  $86.41$ & $11.58$ & $12.46$ & $8.40$ & $7.69$ & --- & --- \\
\verb|tf_efficientnet_b6_ns| &  $86.44$ & $80.21$ & $79.02$ & $80.91$ & $79.16$ & $75.36$ & $74.20$ \\
\verb|tf_efficientnet_b7_ns| &  $86.83$ & $80.34$ & $78.79$ & $80.89$ & $78.97$ & $74.93$ & $73.91$ \\
\verb|tf_efficientnet_l2_ns_475| &  $88.24$ & --- & $78.94$ & --- & $78.96$ & --- & $74.11$ \\
\verb|vit_large_patch16_384| &  $87.09$ & $79.63$ & $78.45$ & $80.48$ & $78.88$ & --- & --- \\
\verb|volo_d5_224| \scalebox{1.25}{+}  \verb|volo_d4_224| \scalebox{1.25}{+}  \verb|volo_d3_224| \scalebox{1.25}{+}  \verb|volo_d2_224| \scalebox{1.25}{+}  \verb|volo_d1_224| &  $86.09$ & $80.45$ & $79.31$ & $80.81$ & $79.16$ & --- & --- \\
\verb|volo_d5_512| &  $87.04$ & --- & $78.04$ & --- & $76.61$ & --- & --- \\
\bottomrule
\end{tabular}

    }
    \caption{Effect of distillation from pretrained teachers (Timm library) on 
    the performance of MobileNetV3-large, ResNet-50, ViT-Small trained for 
    $150$ and $300$ epochs. This table includes the details of 
    \cref{fig:timm_distill_e150,fig:timm_distill_e300}.}
\label{tab:timm_distill_all}
\end{table*}

\newpage
\clearpage
\subsection{Super ensembles on \IN{}}\label{sec:super_ensembles}

\begin{figure*}[thb!]
\begin{center}
    \includegraphics[width=\textwidth]{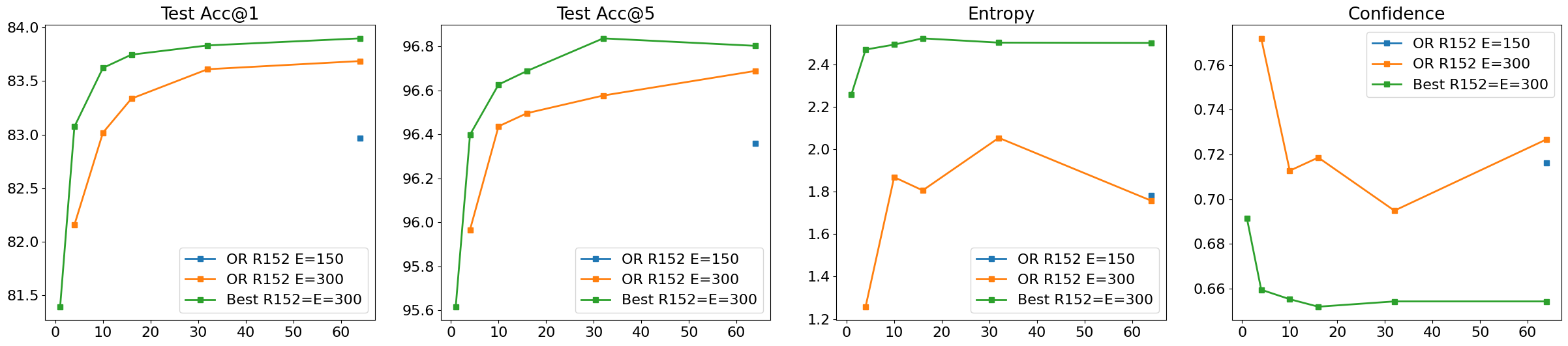}
\end{center}
    \vspace{-0.5cm}
    \caption{\IN{} accuracy of super ensembles as the size of the ensemble 
    is increased. OR means the ensemble is created from diverse augmentation 
    choices while Best means only the random seed is different between models.
    }\label{fig:imagenet_superensemble}
\end{figure*}

It is common to limit the number of models in an ensemble to less than 10 
members and typically only $4$. The reason is partly that larger ensembles are 
more expensive to evaluate at test time as well as training with knowledge 
distillation. Dataset reinforcement allows us to consider expensive teachers 
such as super ensembles with significantly more than 10 members. In 
\cref{tab:cifar100_super_ensembles_diverse_distill} we present results for 
super ensembles on \CIFAR{} and in \cref{fig:imagenet_superensemble} we 
present results on \IN{}. On \CIFAR{} we create super ensembles by training 
{128} models in parallel for ResNet-18, ResNet-50, and ResNet-152 
architectures.  To increase diversity, we train models with 16 choices of 
enable/disable {4} augmentations (CutMix, MixUp, RandAugment, and Label 
Smoothing) and train with $8$ different random seeds for each choice. In total 
we train $8\times 16=128$ models.  
\cref{tab:cifar100_super_ensembles_diverse_acc} shows the accuracy of the super 
ensembles while \cref{tab:cifar100_super_ensembles_diverse_distill} shows the 
accuracy of distillation with the super ensembles. We observe that the best 
student accuracy is achieved with the largest ensemble 128xR152.
Interestingly, super ensembles of small models (128xR50) are better than 
standard ensembles of large models (10xR152).
With super ensembles we achieve strong accuracies for ResNet-50 at 86.30 and 
ResNet-152 at 87.03.

\begin{figure}[thb!]
\begin{center}
    \includegraphics[width=0.3\linewidth]{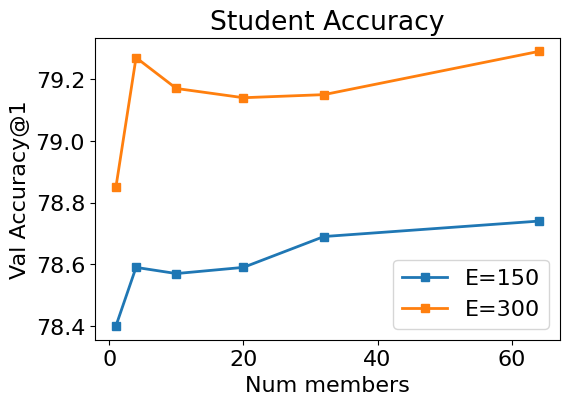}
\end{center}
    \vspace{-0.5cm}
    \caption{\IN{} super ensemble distillation accuracy for ResNet-50 
    facilitated by dataset reinforcement.}\label{fig:imagenet_superensemble_distill}
\end{figure}

We also consider super ensembles for \IN{} using dataset reinforcement.  
Knowledge distillation with super ensembles of larger than 10 members on 
\IN{} becomes challenging and resource demanding.
\cref{fig:imagenet_superensemble} shows the validation accuracy of the ensemble 
and \cref{fig:imagenet_superensemble_distill} shows the accuracy of student 
training with the reinforced \IN{} dataset using the super ensemble.
We observe that the entropy and confidence of the teacher on the validation set 
are not more correlated with the distillation accuracy than the accuracy of the 
validation teacher.  In particular, large ensembles are more accurate but not 
necessarily better teachers. In summary, we observe that the optimal ensemble 
size for KD is around 4.
\begin{table*}[thb!]
\centering
\parbox{0.6\linewidth}{
\resizebox{0.6\columnwidth}{!}{
\begin{tabular}{l cccccc}
    \toprule[1.5pt]
    & \textbf{10xR18} & \textbf{128xR18} & \textbf{10xR50} & \textbf{128xR50} & \textbf{10x152} & \textbf{128xR152}
    \\ 
    \midrule[1.25pt]
    ResNet-18         & 83.57 & 84.24 & 83.43 & 84.07 & 83.65 & {\bf 84.25}\\
    ResNet-50         & 84.40 & 85.16 & 84.33 & 86.38 & 86.03 & {\bf 86.30}\\
    ResNet-152        & 85.01 & 85.74 & 85.00 & 86.80 & 86.85 & {\bf 87.03}\\
    \bottomrule[1.5pt]
\end{tabular}
    }
\caption{Distillation on \CIFAR{} with super diverse ensembles.}
    \label{tab:cifar100_super_ensembles_diverse_distill}
}
\hfill
\parbox{0.35\linewidth}{
\resizebox{0.35\columnwidth}{!}{
\begin{tabular}{l cc}
    \toprule[1.5pt]
    & \textbf{Single best} & \textbf{Ensemble (128x)}
    \\ \midrule[1.25pt]
    ResNet-18         & 81.57 & 85.88 \\
    ResNet-50         & 83.43 & 87.29 \\
    ResNet-152        & 84.44 & 87.92 \\
    \bottomrule[1.5pt]
\end{tabular}
    }
\caption{Accuracy of super diverse ensembles on \CIFAR{}.}
\label{tab:cifar100_super_ensembles_diverse_acc}
}
\end{table*}

\newpage
\clearpage
\section{Expanded study on reinforcing 
\IN{}}\label{sec:reinforce_imagenet_sup}
In this section, we provide ablations on the number and type of augmentations 
using a single relatively cheap teacher (ConvNext-Base-IN22FT1K) that still 
provides comparatively good improvements across all students.

\subsection{What is the best combination of augmentations for
reinforcement?}\label{sec:aug_difficulty_sup}

To recap, using our selected teachers from \cref{sec:good_teacher}, we 
investigate the choice of augmentations for dataset reinforcement.
Utilizing Fast Knowledge Distillation~\citep{shen2021fast}, we store the sparse 
outputs of a teacher on multiple augmentations.
For efficiency, we store top 10 probabilities predicted by the teacher, along 
with the augmentation parameters and reapply augmented images in the data 
loader of the student.
We observe that light-weight CNNs perform best on easier reinforcements while 
transformers perform best with difficult reinforcements. We balance this 
tradeoff using a mid-difficulty reinforcement.

We refer to the combination of baseline augmentations fixed resize, random crop 
and horizontal flip by CropFlip. In addition, we consider the following 
augmentations for dataset reinforcement: Random-Resize-Crop (\RRC{}), 
MixUp~\citep{zhang2017mixup} and CutMix~\citep{yun2019cutmix} (\Mixing{}), and 
RandomAugment~\citep{cubuk2020randaugment} and RandomErase (\RARE{}).  We also 
combine \Mixing{} with \RARE{} and refer to it as \MsRs.  We add all 
augmentations on top of \RRC{} and for clarity add + as shorthand for 
{\RRC{}$+$}.  Except for mixing augmentations, reapplying all augmentations has 
zero overhead compared to standard training with the same augmentations.  For 
mixing augmentations, our current implementation has approximately $30\%$ 
wall-clock time overhead because of the extra load time of mixing pairs stored 
with each reinforced sample. We discuss efficient alternatives in 
\cref{sec:library}.  Our balanced solution, \RRCRARE{}, does not use mixing and 
has zero overhead.

\begin{figure*}[t!]
\centering
    \begin{subfigure}[t]{0.28\textwidth}
        \centering
        \includegraphics[width=\textwidth]{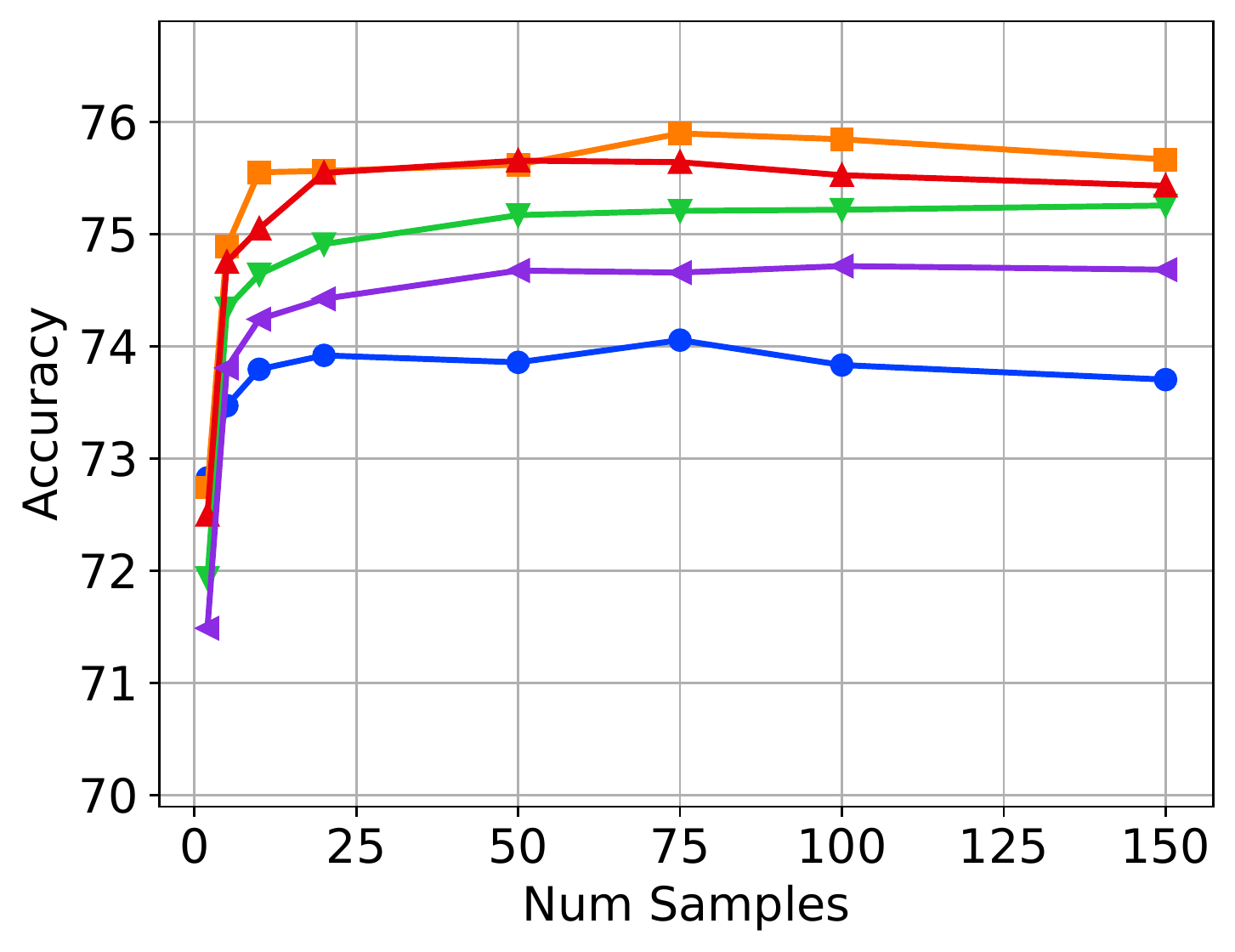}
        \caption{Light-weight CNN 
        (MobileNetV3)}\label{fig:imagenet_MobileNetv3_augs_e150}
    \end{subfigure}
     \hfill
    \begin{subfigure}[t]{0.28\textwidth}
        \centering
        \includegraphics[width=\textwidth]{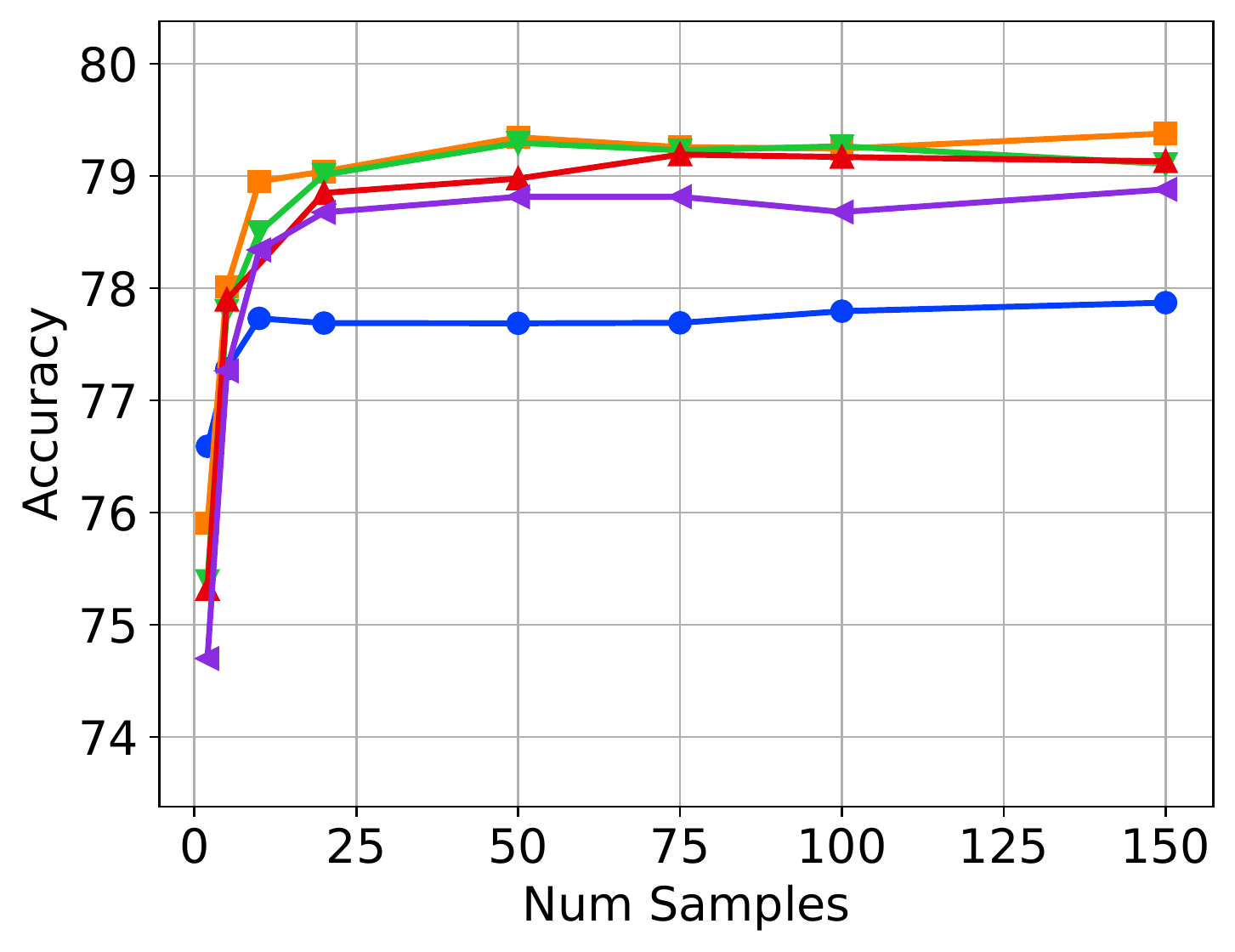}
        \caption{Heavy-weight CNN 
        (ResNet-50)}\label{fig:imagenet_R50_augs_e150}
    \end{subfigure}
    \hfill
    \begin{subfigure}[t]{0.28\textwidth}
        \centering
        \includegraphics[width=\textwidth]{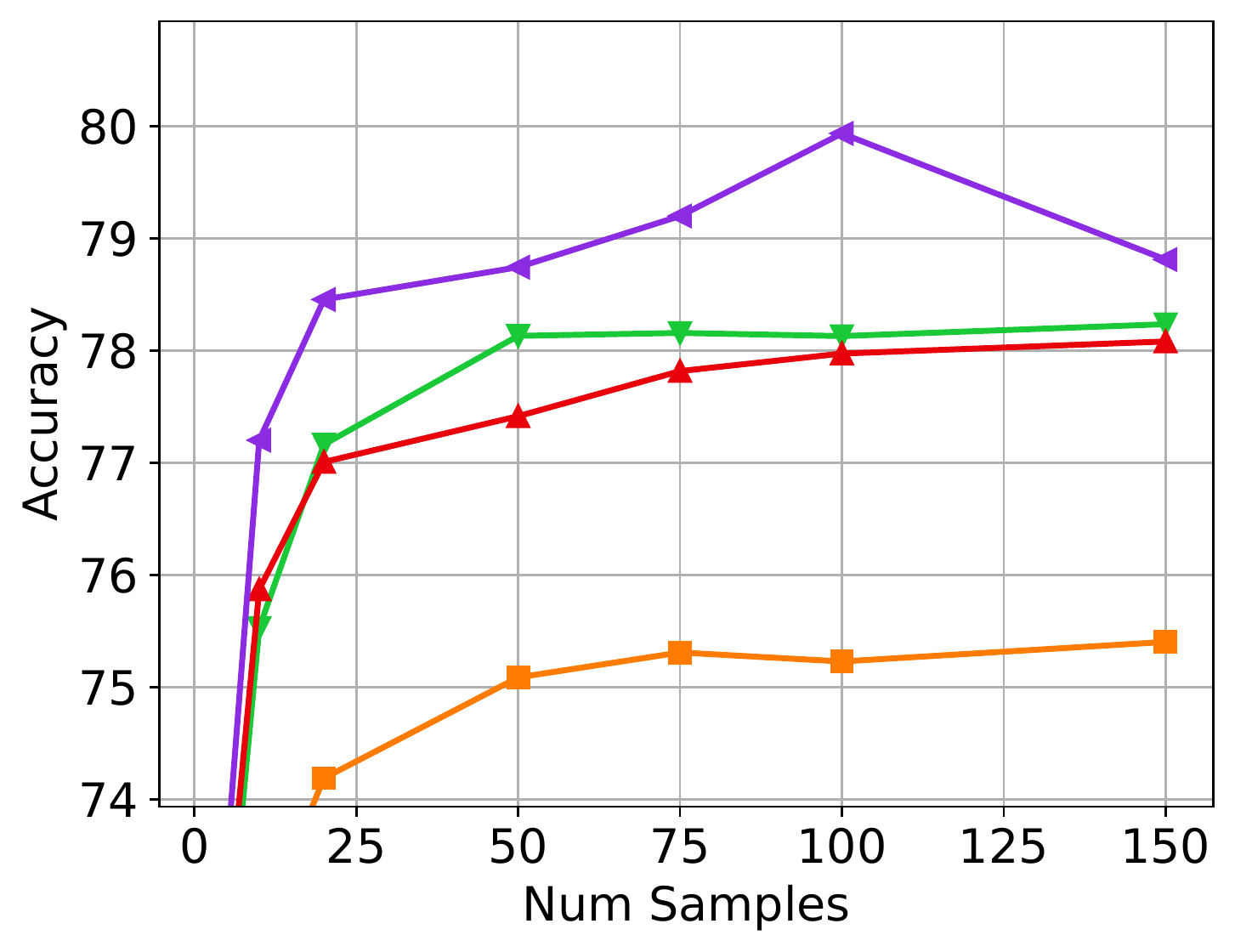}
        \caption{Transformer 
        (ViT-Small)}\label{fig:imagenet_ViTSmall_augs_e150}
    \end{subfigure}
    \hfill
    \begin{minipage}[t]{0.14\linewidth}
        \vspace{-3.5cm}
        \begin{subfigure}[t]{\textwidth}
            \centering
            \includegraphics[width=\textwidth]{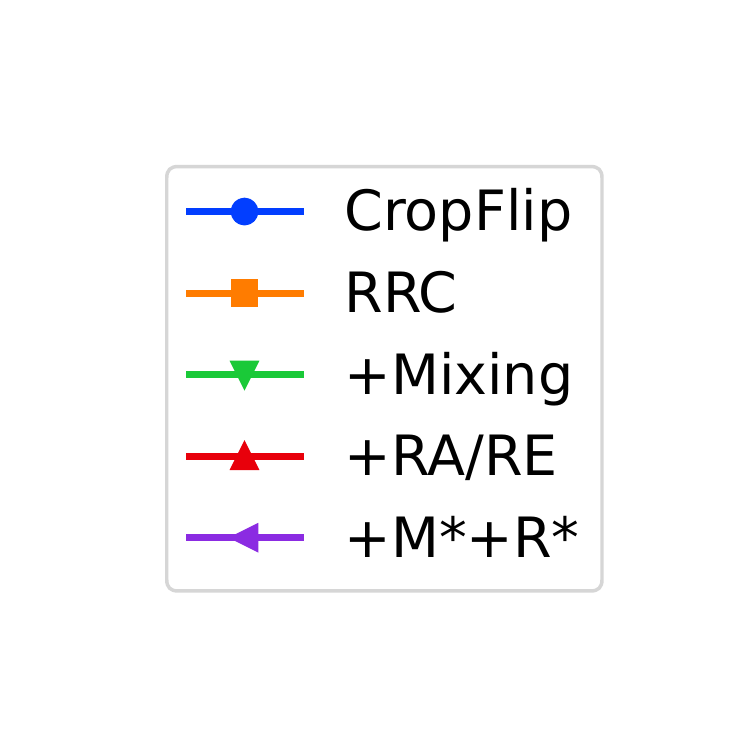}
        \end{subfigure}
    \end{minipage}
    \vspace{-3mm}
    \caption{\textbf{Light-weight CNNs prefer easy while Transformers prefer 
    difficult reinforcements and we balance the tradeoff.}
    \IN{} validation accuracy of three representative architectures trained 
    on reinforcements of \IN{}. We use ConvNext-Base-IN22FT1K as the teacher 
    and train for 150 epochs.  The x-axis is the number of augmentations stored 
    per original sample in the \IN{} training set.
    In favor of dataset reinforcement, we observe that training with $25-50$ 
    samples provides similar gains to training with more samples.
    The baseline augmentation is Fixed Resize-RandomCrop and horizontal flip 
    (CropFlip). In addition we consider the following augmentations for 
    reinforcement:
    Random-Resize-Crop and horizontal flip (\RRC{}), MixUp and CutMix 
    (\Mixing{}), RandomAugment/RandomErase (\RARE{}) and Mixing+RA/RE 
    (\MsRs{}).  We add these augmentations on top of \RRC{} and for clarity add 
    + as shorthand for {\RRC{}$+$}.
    }\label{fig:aug_difficulty}
    \vspace{-4mm}
\end{figure*}

\Cref{fig:aug_difficulty} shows the accuracy of various models trained on 
reinforced datasets. We observe that the light-weight CNN performs best with 
\RRC{} as the most simple augmentation after CropFlip while the transformer 
performs best with the most difficult set of reinforcements in \RRCMsRs{}. This 
observation matches the standard state-of-the-art recipes for training these 
models. At the same time, we observe that \RRCRARE{} provides nearly the best 
performance for all models without the extra overhead of mixing methods in our 
implementation.

Consistent across three models and reinforcements, we observe that even though 
we train for 150 epochs, at most $25-50$
different augmentations of each training sample
is enough to achieve the 
best accuracy for almost all methods. This gives at least $\times 3$ reduction 
in the number of samples we can take advantage of given a fixed training 
budget. Based on this observation and following \citet{beyer2022knowledge}, in 
\cref{sec:long_train} we train models for up to 1000 epochs while reinforce 
datasets with 400 augmentation samples.

\subsection{Augmentation: invariance vs imitation}\label{sec:invariance}
Data augmentation is crucial to train generalizable models in various domains. 
The key objective is to make the model invariant to content-preserving 
transformations. In knowledge distillation, however, it is not clear whether the student benefits more from being invariant to data augmentations as in \cref{eqn:invariance} or from imitating teacher's variations on augmented data as in \cref{eqn:imitation}. The training objective for each case is as follows:
\begin{equation}\label{eqn:invariance}
\text{(Invariance)}\hspace{5mm}
    \min_{\theta}\, \mathbb{E}_{\vx \sim \mathcal{D}, \hat{\vx} \sim \mathcal{A}(\vx)} \mathcal{L}(f_{\theta}(\hat{\vx}), g({\color{blue} \vx})) 
\end{equation}
\begin{equation}\label{eqn:imitation}
\text{(Imitation)}\hspace{5mm}
    \min_{\theta}\, \mathbb{E}_{\vx \sim \mathcal{D}, \hat{\vx} \sim \mathcal{A}(\vx)} \mathcal{L}(f_{\theta}(\hat{\vx}), g({\color{blue} \hat{\vx}}))
\end{equation}
where, $\mathcal{D}$ is the training dataset, $\mathcal{A}$ is augmentation function, $f_{\theta}$ is the student model parameterized with $\theta$, $g$ is the teacher model, and $\mathcal{L}$ is the loss function between student and 
teacher outputs.

\begin{figure}[thb!]
    \centering
    \includegraphics[width=.5\linewidth]{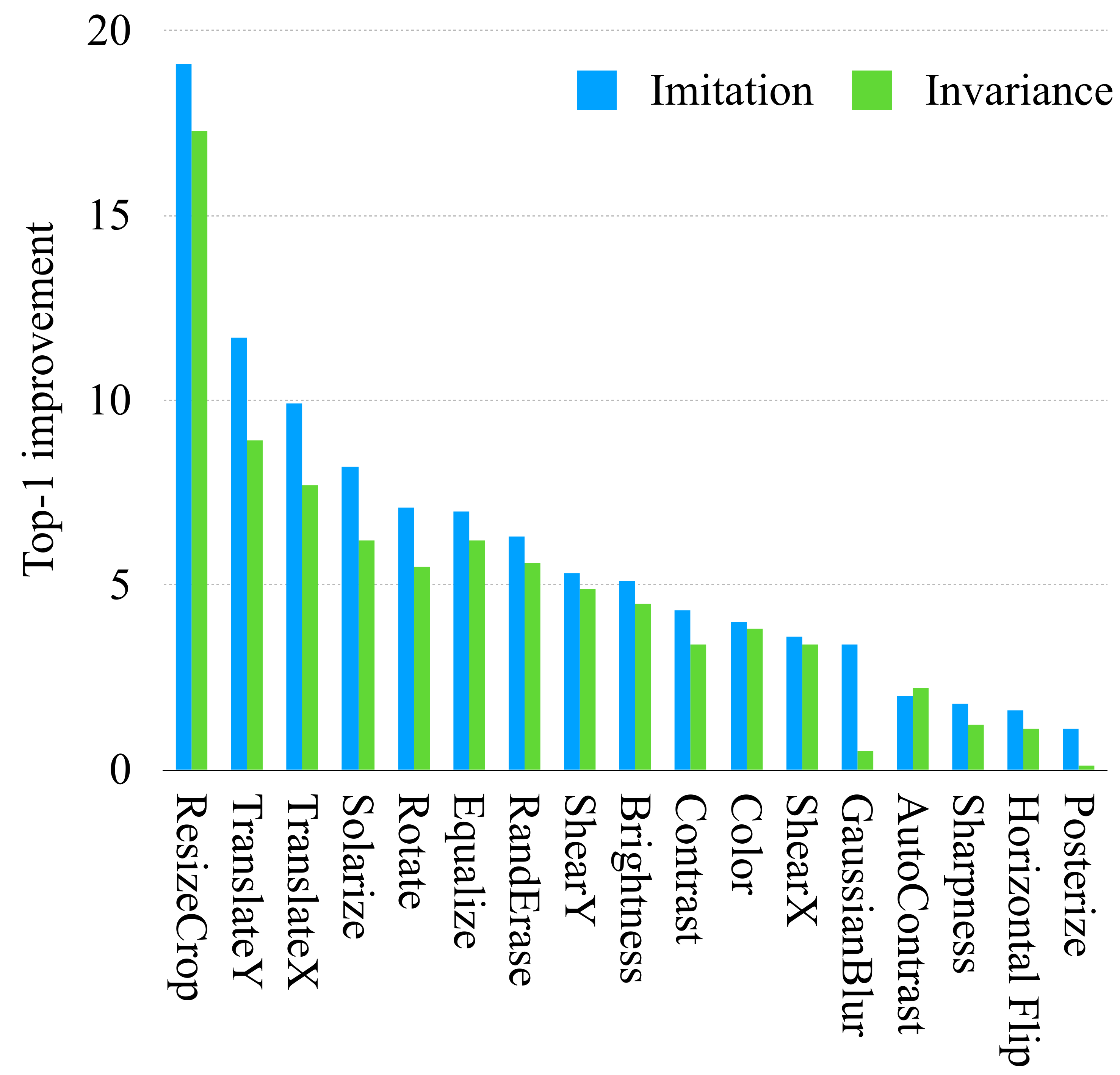}
    \vspace{-2mm}
    \caption{\IN{} top-1\% accuracy improvement when distilling knowledge 
    from a ConvNext (Base-IN22FT1K) teacher to a ViT-tiny student using 
    a single augmentation with training objectives in \cref{eqn:invariance} and 
    \cref{eqn:imitation}. No augmentation top-1\% accuracy is 
    $54\%$.\label{fig:invariance}}
    \vspace{-4mm}
\end{figure}

In \cref{fig:invariance}, we compare the above training objectives for a wide 
range of augmentations in computer vision. For most augmentations, we observe 
imitation is more effective than invariance. This is consistent with 
observation in \citet{beyer2022knowledge}.  Therefore, in our setup we use 
augmentations only for imitation (and not invariance).

\subsection{Library size: Can we limit the mixing pairs?}\label{sec:library}
Mixing augmentations have the extra overhead of the load time for the 
corresponding pair in each mini-batch. Standard training does not have such an 
overhead because the mixing is performed on random pairs within a mini-batch.  
In dataset reinforcement, the pairs that have been matched in the reinforcement 
phase are limited to the number of samples stored and do not always appear in 
the same mini-batch during the student training time. This means, we have to 
load the matching pair for every sample in the mini-batch that doubles the data 
load time and becomes an overhead for CPU-bound models. This overhead in the 
smallest models we consider is at most 30\%. Even though much lower than the 
cost of knowledge distillation, it is still more than our desiderata would 
allow.

\begin{figure*}[htb!]
\begin{center}
    \begin{subfigure}[t]{0.27\textwidth}
        \includegraphics[width=\textwidth]{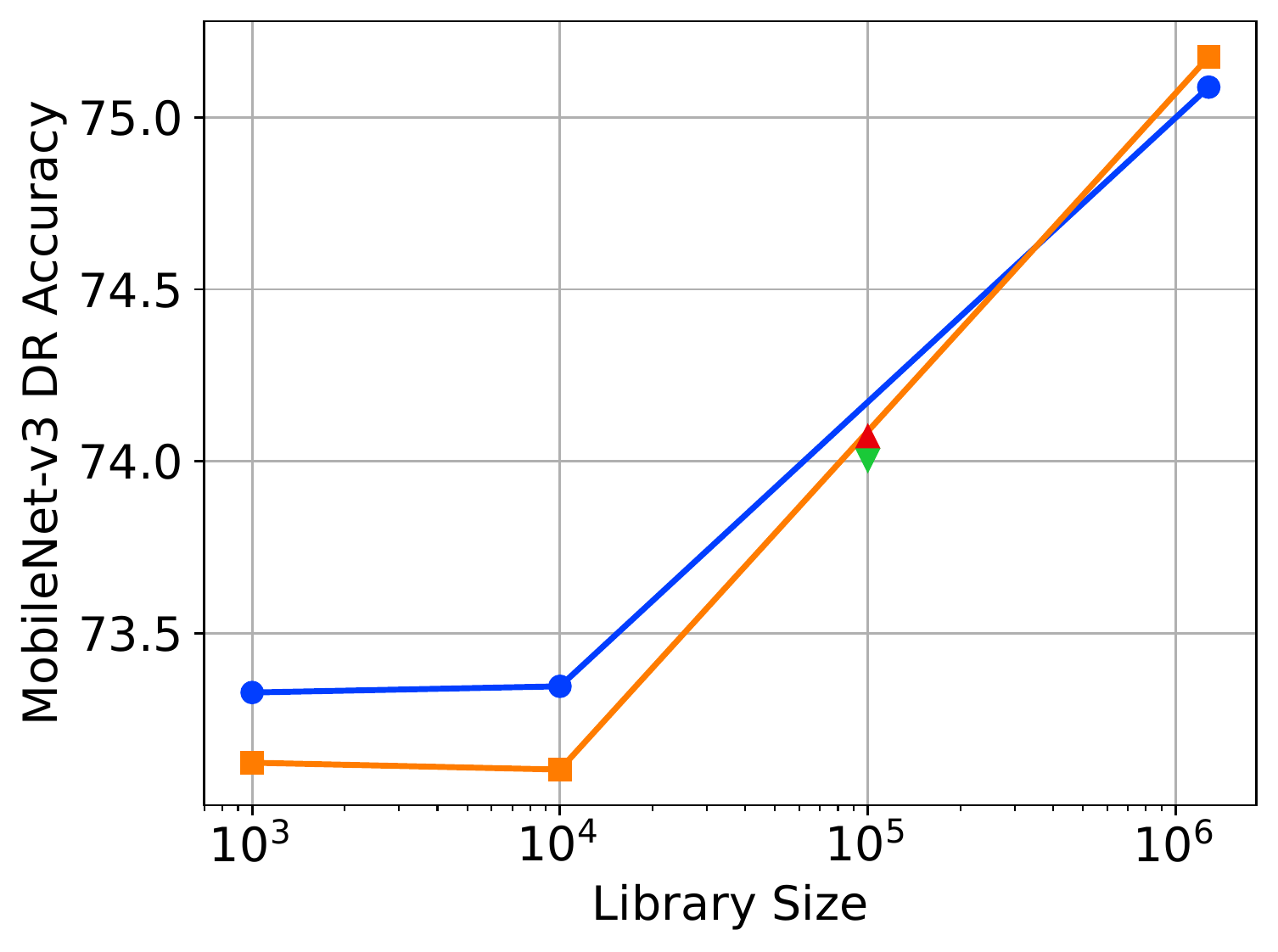}
        \caption{MobileNetV3}\label{fig:imagenet_MobileNetv3_library_e150}
    \end{subfigure}
    \hfill
    \begin{subfigure}[t]{0.27\textwidth}
        \includegraphics[width=\textwidth]{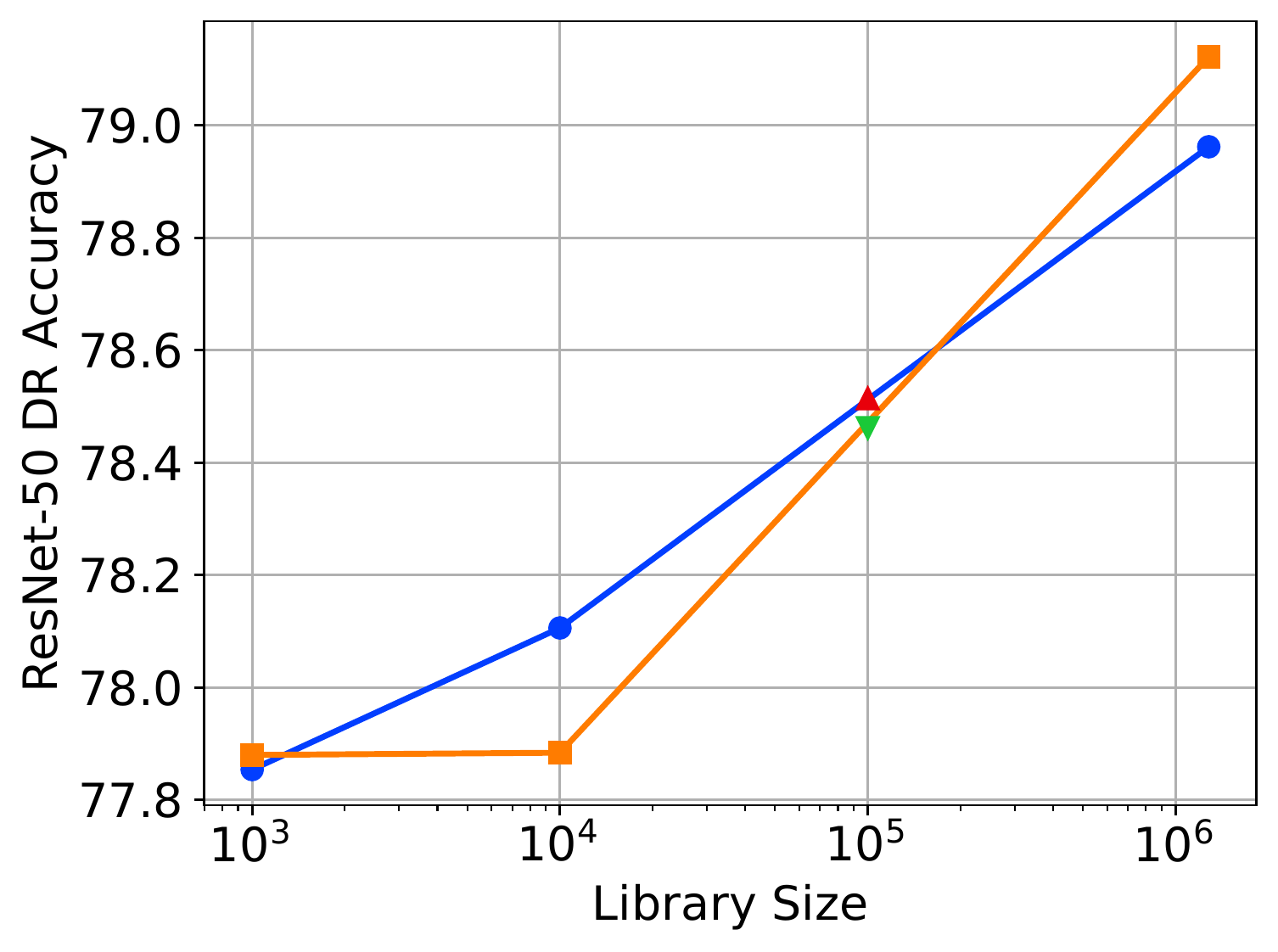}
        \caption{ResNet50}\label{fig:imagenet_R50_library_e150}
    \end{subfigure}
    \hfill
    \begin{subfigure}[t]{0.27\textwidth}
        \includegraphics[width=\textwidth]{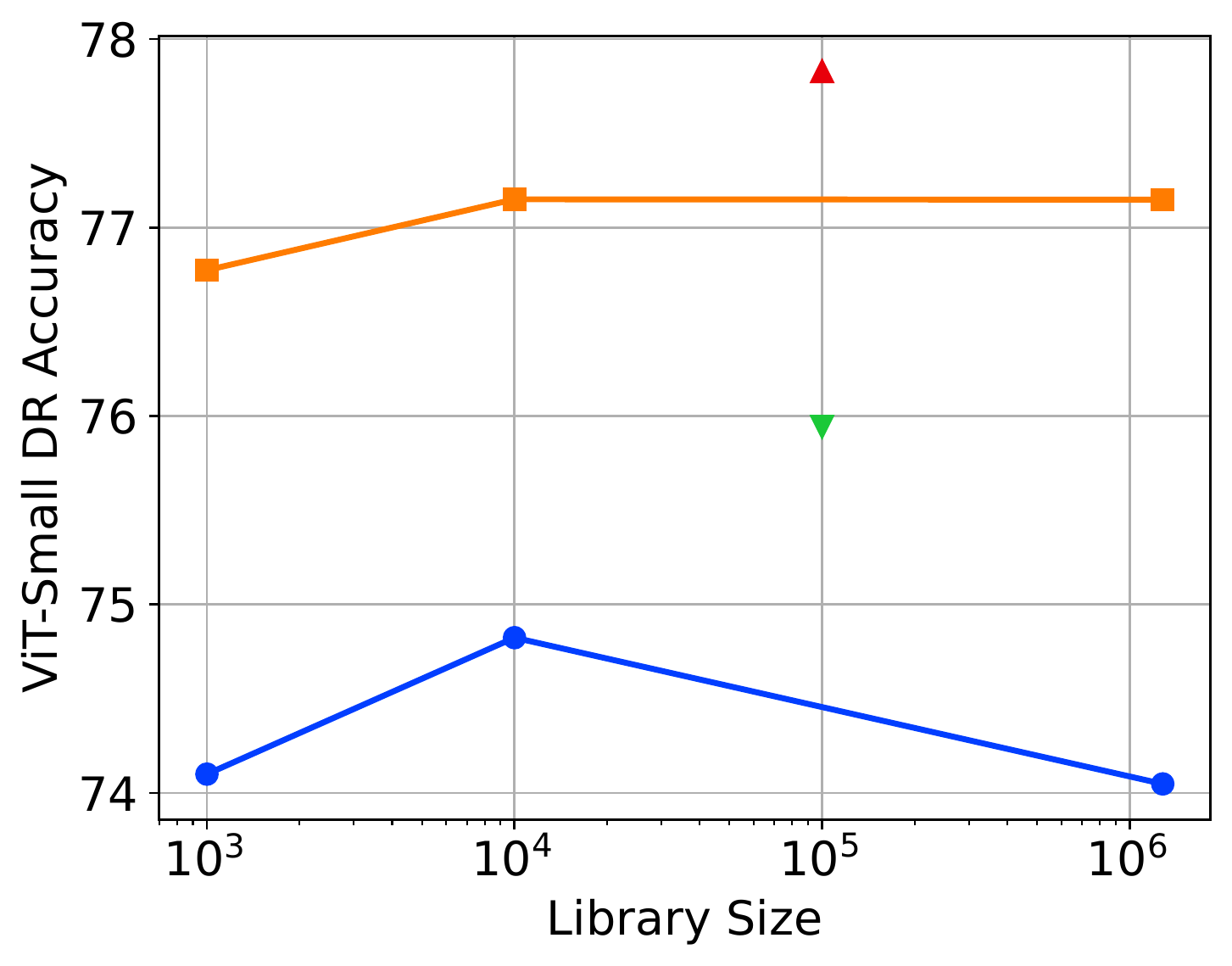}
        \caption{ViT-Small}\label{fig:imagenet_ViTSmall_library_e150}
    \end{subfigure}
    \hfill
    \begin{minipage}[t]{0.17\linewidth}
        \vspace{-4.2cm}
        \begin{subfigure}[t]{\textwidth}
            \includegraphics[width=\textwidth]{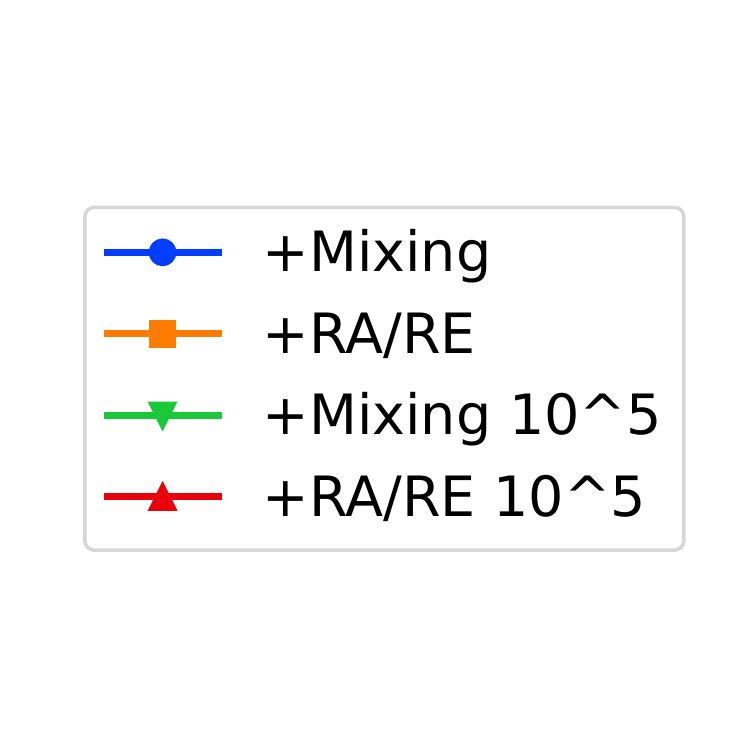}
        \end{subfigure}
    \end{minipage}
\end{center}
    \vspace{-0.5cm}
    \caption{\textbf{Library of mixing pairs reduces wall-clock overhead for 
    mixing methods but negatively impacts accuracy.} We plot the validation 
    accuracy for models trained with \INp{} as we vary the library size.  
    The teacher is ConvNext-Base-IN22FT1K. See \cref{sec:library} for details.  
    (E=150)}\label{fig:library}
\end{figure*}

We consider an alternative where the pairing is done only with a library of 
selected samples from the training set. The library can be loaded in the memory 
once and reduce the additional cost incurred during the training.
\cref{fig:library} shows the performance as we vary the library size. Even 
a relatively large library does not cover the accuracy drop caused by the 
reduced randomness in the mixing. The reason is that to reduce the cost, we can 
only have one augmentation per sample in the library which reduces the 
randomness from the mixing substantially and negatively affects knowledge 
distillation.

\begin{figure*}[t!]
\begin{center}
    \begin{subfigure}[t]{0.27\textwidth}
        \includegraphics[width=\textwidth]{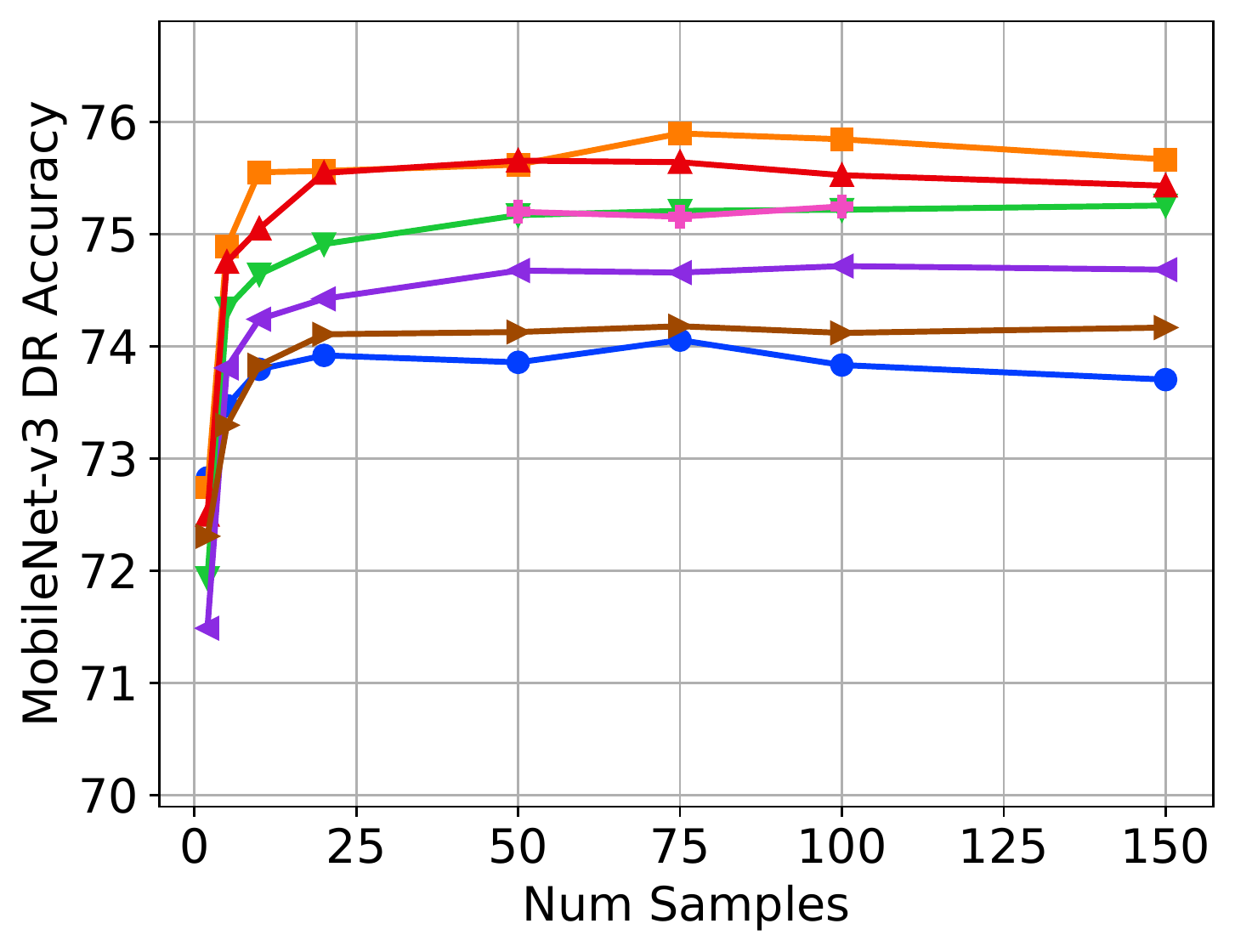}
        \caption{Light-weight CNN 
        (MobileNetV3)}\label{fig:imagenet_MobileNetv3_augs_e150_extra}
    \end{subfigure}
    \begin{subfigure}[t]{0.27\textwidth}
        \includegraphics[width=\textwidth]{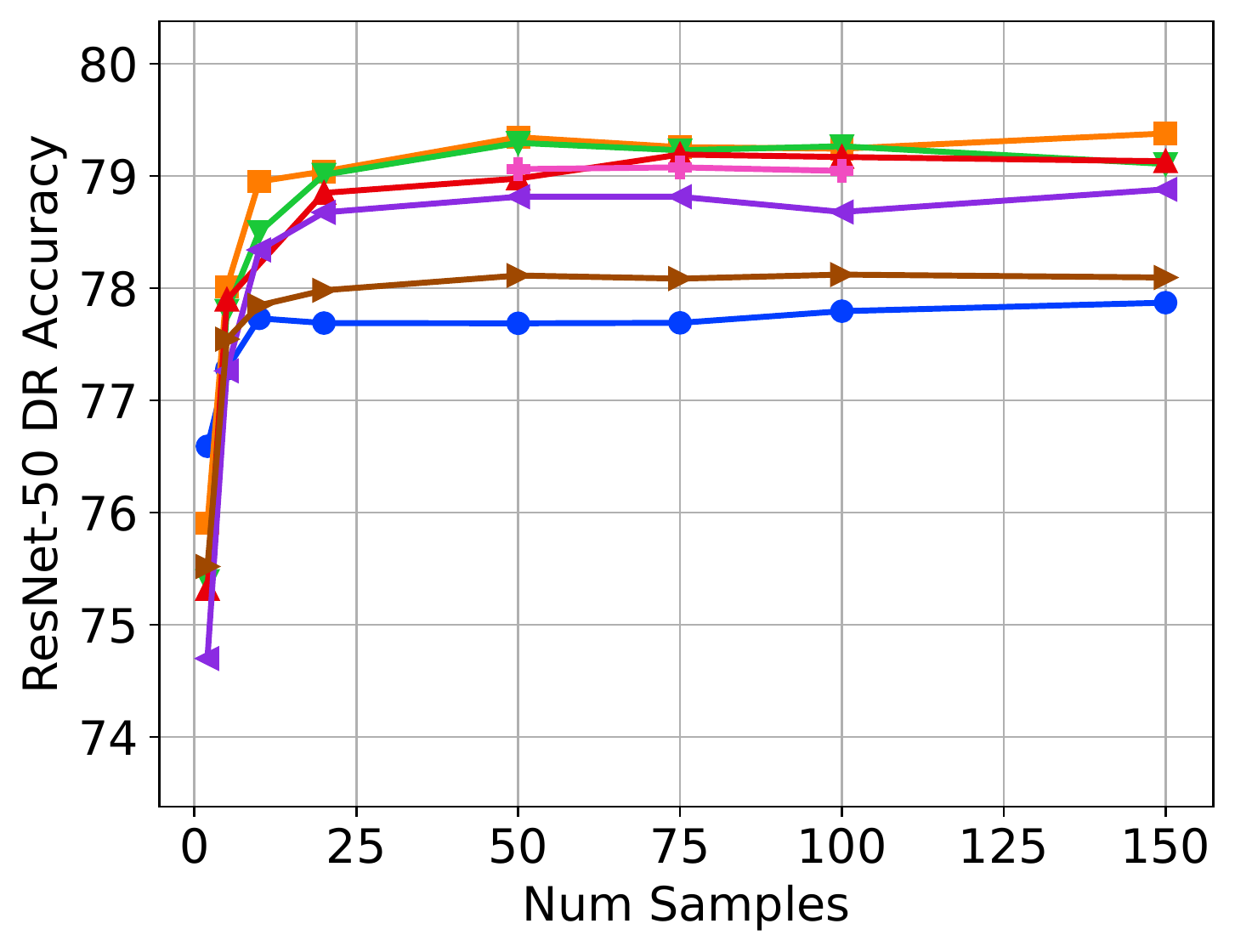}
        \caption{Heavy-weight CNN 
        (ResNet-50)}\label{fig:imagenet_R50_augs_e150_extra}
    \end{subfigure}
    \hfill
    \begin{subfigure}[t]{0.27\textwidth}
        \includegraphics[width=\textwidth]{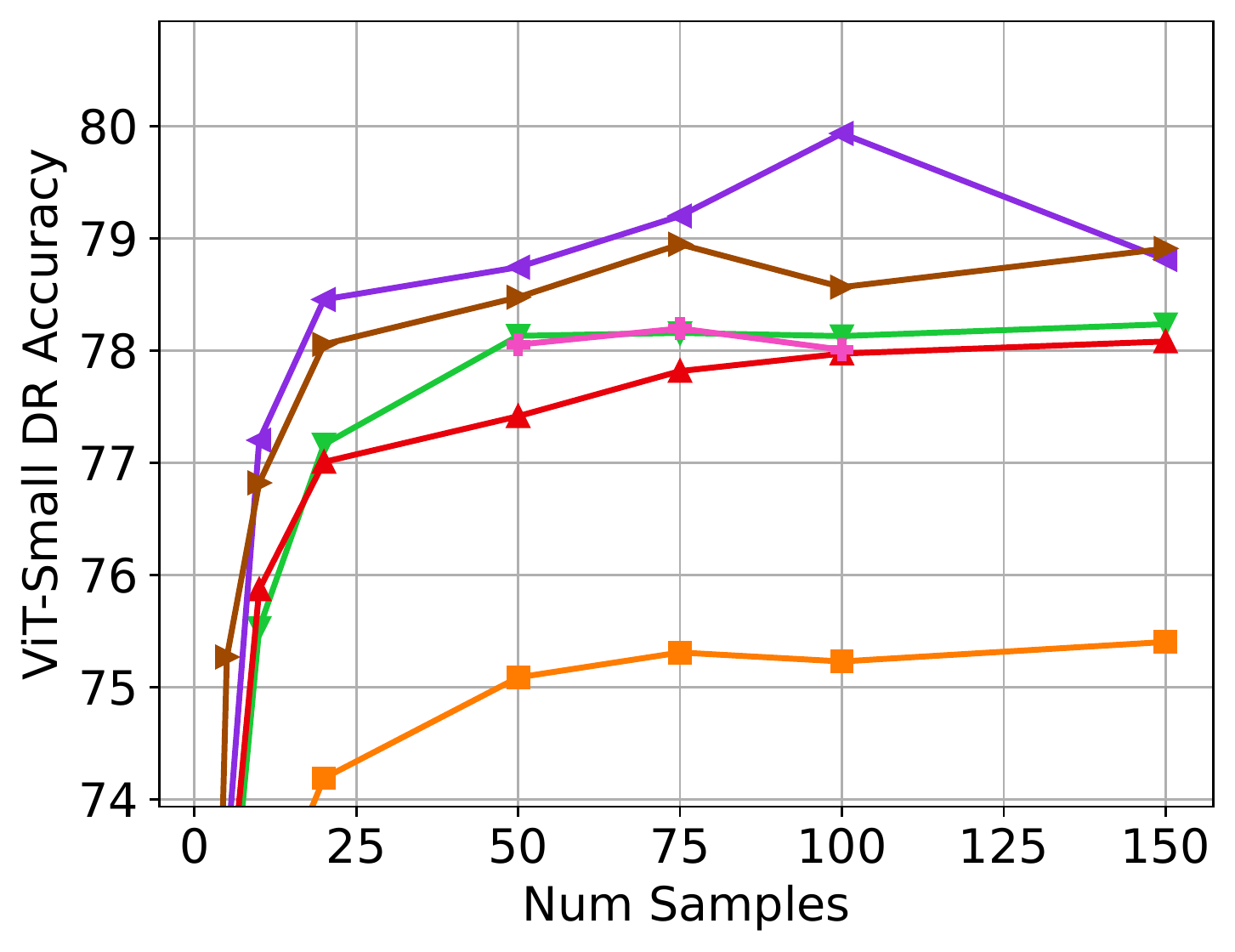}
        \caption{Transformer 
        (ViT-Small)}\label{fig:imagenet_ViTSmall_augs_e150_extra}
    \end{subfigure}
    \hfill
    \begin{minipage}[t]{0.17\linewidth}
        \vspace{-3.5cm}
        \begin{subfigure}[t]{\textwidth}
            \includegraphics[width=\textwidth]{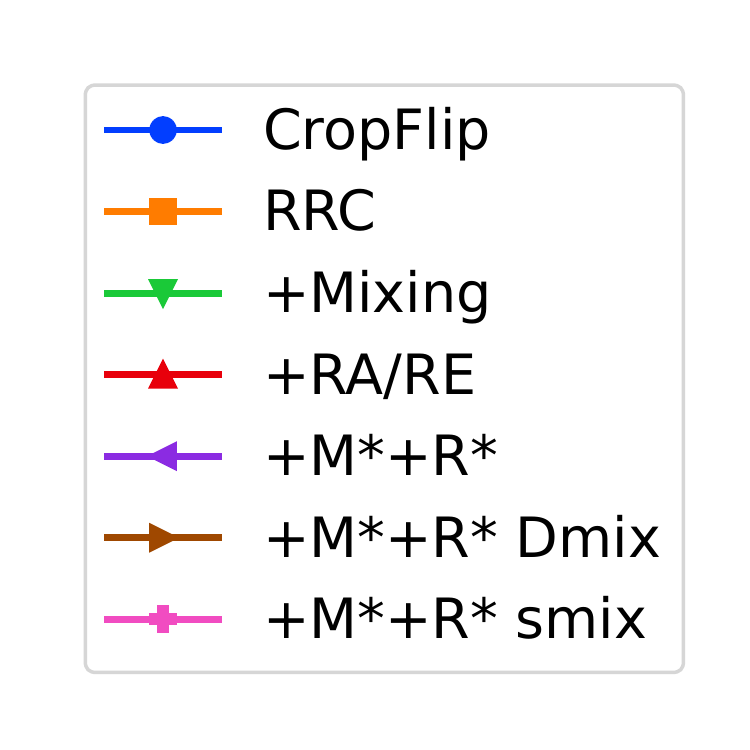}
        \end{subfigure}
    \end{minipage}
\end{center}
    \vspace{-0.5cm}
    \caption{\textbf{Alternative Mixing Augmentations.}
    Accuracy of three representative architectures trained on reinforcements of 
    \IN{}. We use ConvNext-Base-IN22FT1K as the teacher and train for 150 
    epochs. The x-axis is the number of augmentations stored per original 
    sample in the \IN{} training set. Augmentations used are Fixed 
    Resize-RandomCrop and horizontal flip (CropFlip),
    Random-Resize-Crop and horizontal flip (\RRC{}), MixUp and CutMix (Mixing), 
    RandomAugment/RandomErase (RA/RE) and Mixing+RA/RE (M*+R*).  Alterantive 
    mixing augmentations: Double-mix (Dmix) and Self-mix 
    (Smix).}\label{fig:aug_difficulty_extra}
\end{figure*}

We also consider variations of mixing in \cref{fig:aug_difficulty_extra}. We 
consider two variations: Double-mix and Self-mix. In double-mix, for every 
augmented pair, we store two outputs with two sets of mixing coefficients. This 
means for every mini-batch we can load half the mini-batch along with a random 
pair for each sample, perform the stored augmentation on each and get two 
different mixed samples. As a result the overhead is zero.  Second alternative, 
self-mix, mixes every image only by itself. As such, there is no data load 
time, but there is still an extra overhead of preprocessing the input twice.
\cref{fig:aug_difficulty_extra} shows that neither of the considered 
alternatives provide a better tradeoff compared with \RRCRARE{}. Therefore, we use \RRCRARE{} in our paper and call it \INp{}.

\subsection{What is the best curriculum of 
reinforcements?}\label{sec:curriculum}
\begin{figure*}[thb!]
    \centering
    \begin{subfigure}[t]{0.33\textwidth}
        \centering
        \includegraphics[width=\textwidth]{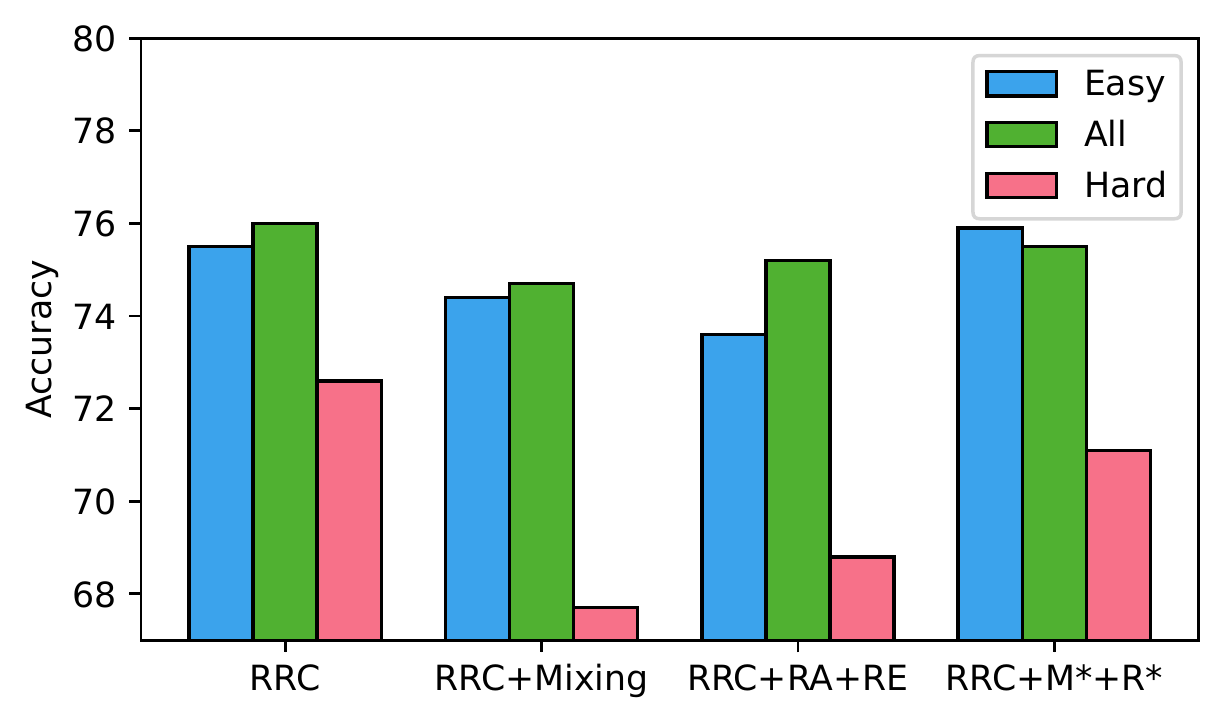}
        \caption{Light-weight CNN 
        (MobileNetV3)}\label{fig:curriculum-RRC_Mixing}
    \end{subfigure}
    \hfill
    \begin{subfigure}[t]{0.33\textwidth}
        \centering
        \includegraphics[width=\textwidth]{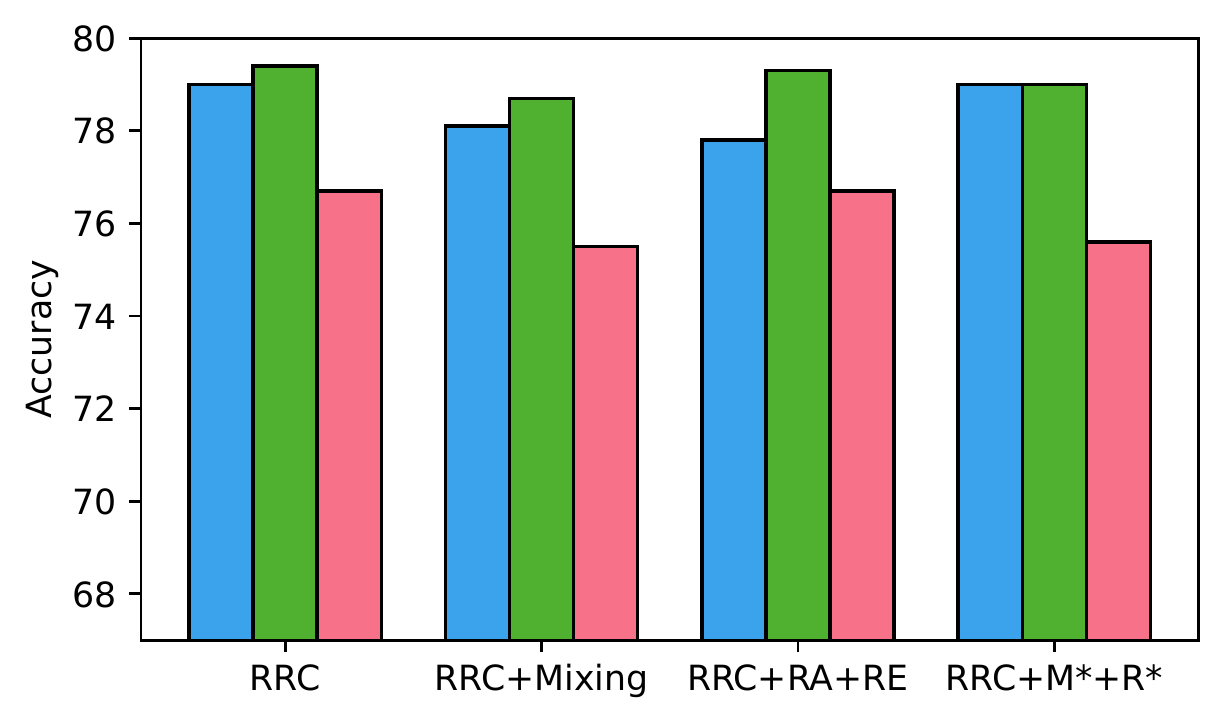}
        \caption{Heavy-weight CNN (ResNet-50)}\label{fig:curriculum-RRC}
    \end{subfigure}
    \hfill
    \begin{subfigure}[t]{0.33\textwidth}
        \centering
        \includegraphics[width=\textwidth]{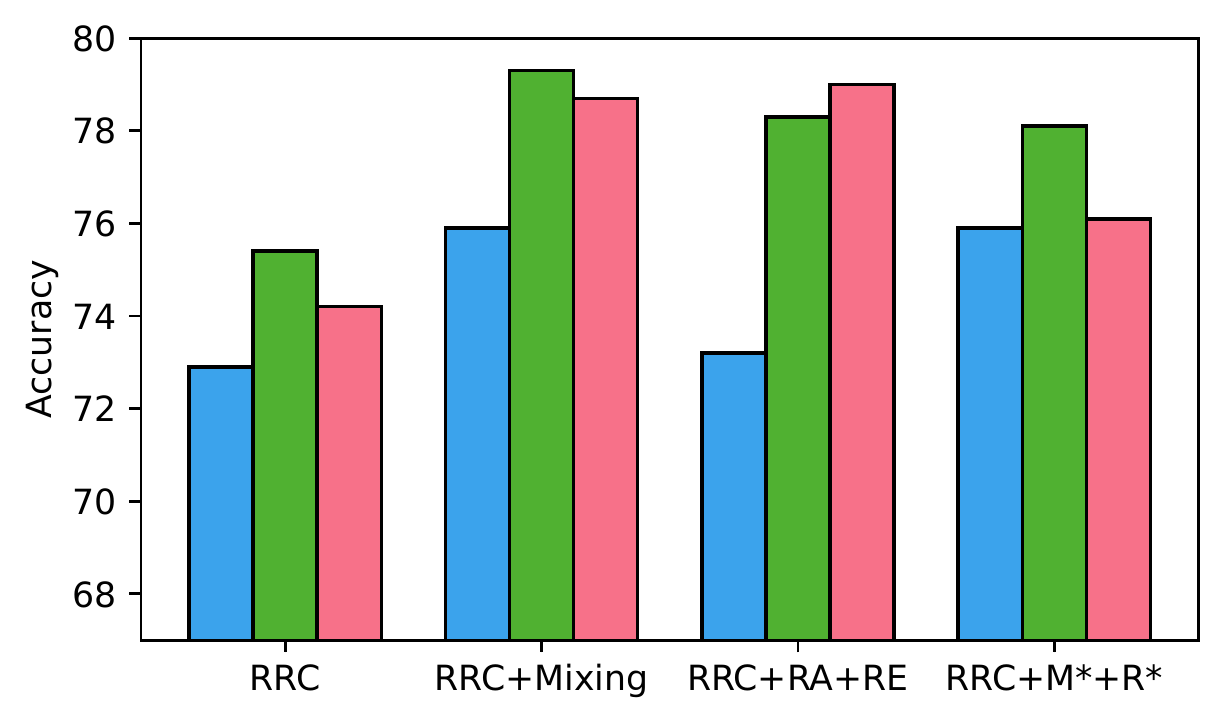}
        \caption{Transformer (ViT-Small)}\label{fig:curriculum-RRC_RA_RE}
    \end{subfigure}
    \vspace{-3mm}
    \caption{
        \textbf{Tradeoff in augmentation difficulty can be further reduced with 
        curriculums.}
        Using random samples of the optimal augmentation is the best (`All').
        If we have access to one reinforcement dataset or we want to only keep 
        a subset of data for efficiency, then it is better to use `easy' 
        curriculums for CNN based architectures and `hard' curriculums for 
        Transformers.  Full table in 
        \cref{sec:curriculum_extra}.}\label{fig:curriculum}
    \vspace{-4mm}
\end{figure*}

The one-time cost of reinforcing a dataset allows us to generate as much useful 
information as we need and store it for future use. An example is various 
metrics that can be used to devise learning curriculums that adapt to specific 
students. In this section we consider a set of initial curriculums we get for 
free with our dataset reinforcement strategy. Specifically, the output of the 
teacher on each sample also incorporates the confidence of the teacher on its 
prediction. We can use the confidence or the entropy of its predictions to make 
curriculums.

Given $\pp\in\R^{c}$ the set of predicted probabilities of the teacher for $c$ 
classes, we define confidence as $\max \pp_j$. For every sample, we order its 
augmentations by the confidence value from $0$ to \#samples. During training, 
at each iteration we only sample from a range of augmentations with indices 
between $[a, b]$, where $0\leq a,b < \text{\#samples}$. We devise curriculums 
by smoothly changing $a,b$ during the training using a cosine function between 
specified values of initial and final values for $a,b$.

\cref{fig:curriculum} shows the performance of various Easy, Hard, and All 
curriculums. Easy curriculums start from $[0, 10]$ (the $10\%$ easiest 
samples), hard samples start from $[90, 100]$ (the $10\%$ hardest samples), and 
All curriculums start from $[0, 100]$ (all the samples). We observe that the 
curriculum provides an alternative knob to control the difficulty of 
reinforcements that we can use adaptively during the training of the student.  
For example, the best performance of the light-weight CNN is with \RRC{} 
combined with the All curriculum, but similar performance can be achieved with 
\RRCRARE{} combined with an Easy curriculum. Similarly, the transformer achieves 
its best performance with \RRCMsRs{} combined with the All curriculum, while 
a similar performance can be achieved with \RRCMixing{} and a Hard curriculum.

In \cref{sec:optimal_augs}, we study various objectives for choosing most 
useful samples during the reinforcement process. We consider storing on the 
most informative samples according to a number of metrics such as entropy, 
loss, and clustering. We make similar observations to the behaviour of 
curriculums that the objectives that increase hardness benefit the transformer 
while the easy objectives benefit the light-weight {CNN}.

\subsection{Additional details of curriculums}\label{sec:curriculum_extra}

\begin{figure}[thb!]
\begin{center}
    \includegraphics[width=0.5\linewidth]{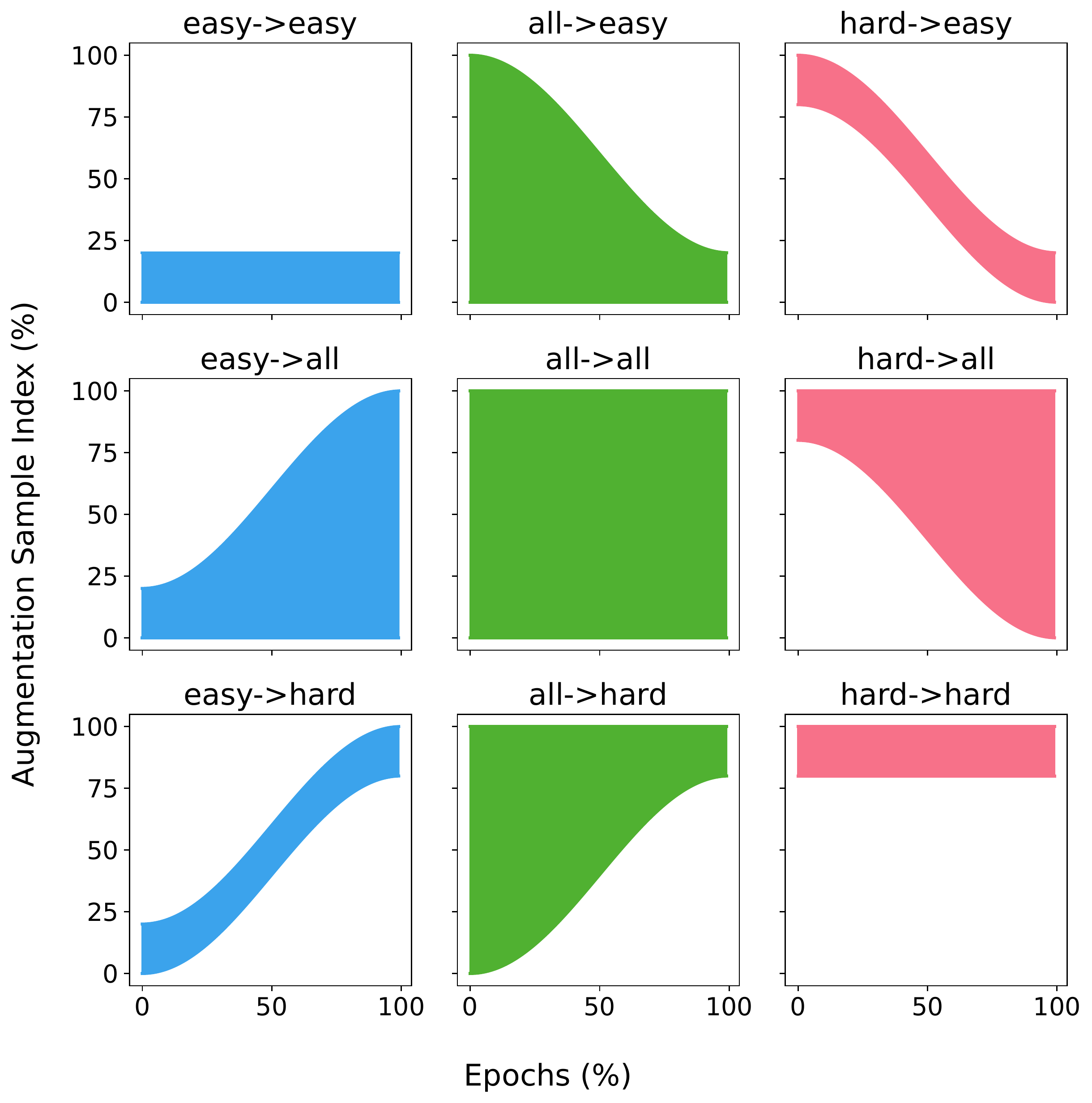}
\end{center}
    \vspace{-0.5cm}
    \caption{\textbf{Illustration of curriculums.} The x-axis shows the 
    percentage of training epochs while the y-axis shows the index of 
    augmentation samples in percentages as we order them from easy to hard by 
    the confidence of the teacher.  Highlighted regions show the subset of 
    indices of reinforcements to uniformly draw from at each epoch.  
    }\label{fig:curriculum_illustration}
    \vspace{-4mm}
\end{figure}

We study reinforcements on curriculums shown in 
\cref{fig:curriculum_illustration}. \Cref{tab:effect_curr} provides the full 
results for the effect of dataset reinforcement curriculums.  We summarized 
these results in \cref{fig:curriculum} where we compared `*$\rightarrow$all' curriculums 
that end with `all' of the data. We observe that the beginning of the 
curriculum has much more impact on the generalization than the end of the 
curriculum. We observe that `all$\rightarrow$*' curriculums perform the best while 
`hard$\rightarrow$*' curriculums perform near optimal for ViT-Small and `easy$\rightarrow$*' performs
best for MobileNetV3-Large. At the same time, we observe clearly that the hard 
and easy curriculums result in significantly worse generalization when used to 
train the opposite architecture, i.e., `easy$\rightarrow$*' for ViT-Small and `hard$\rightarrow$*' 
for MobileNetV3-Large. This result clearly demonstrates the tradeoff in the 
architecture independent generalization controlled by the difficulty of 
reinforcements.

\begin{table*}[thb!]
    \centering
    \resizebox{\columnwidth}{!}{
        \begin{tabular}{lcccccccccccccc}
\toprule[1.5pt]
\multirow{2}{*}{\bfseries Curriculum} & \multicolumn{4}{c}{\bfseries MobileNetV3-Large} & & \multicolumn{4}{c}{\bfseries  ResNet50} & & \multicolumn{4}{c}{\bfseries ViT-Small}\\
\cmidrule[1.25pt]{2-5} \cmidrule[1.25pt]{7-10} \cmidrule[1.25pt]{12-15}
 & RRC & +RA/RE & +Mixing & +M*+R* & & RRC & +RA/RE & +Mixing & +M*+R* & & RRC & +RA/RE & +Mixing & +M*+R* \\ 
 \midrule[1.25pt]
\verb|easy->all| & ${75.5}$ & ${\bm{75.9}}$ & ${73.6}$ & ${74.4}$ & & ${79.0}$ 
    & ${79.0}$ & ${77.8}$ & ${78.1}$ && ${72.9}$ & ${75.9}$ & ${73.2}$ 
    & ${75.9}$\\
\verb|easy->easy| & ${75.2}$ & ${75.7}$ & ${74.3}$ & ${74.5}$ & & ${79.0}$ & ${79.2}$ & ${77.7}$ & ${78.0}$ && ${72.6}$ & ${75.9}$ & ${73.1}$ & ${75.9}$\\
\verb|easy->hard| & ${75.6}$ & ${\bm{75.8}}$ & ${74.1}$ & ${74.6}$ & & ${79.0}$ & ${79.2}$ & ${77.8}$ & ${78.0}$ && ${72.6}$ & ${76.2}$ & ${72.9}$ & ${76.0}$\\
\midrule
\verb|all->all| & ${\bm{76.0}}$ & ${75.5}$ & ${75.2}$ & ${74.7}$ 
    & & ${\bm{79.4}}$ & ${79.0}$ & ${\bm{79.3}}$ & ${78.7}$ && ${75.4}$ 
    & ${78.1}$ & ${78.3}$ & ${\bm{79.3}}$\\
\verb|all->easy| & ${75.7}$ & ${75.5}$ & ${75.2}$ & ${74.6}$ & & ${\bm{79.5}}$ & ${79.2}$ & ${\bm{79.3}}$ & ${78.9}$ && ${75.2}$ & ${78.4}$ & ${78.2}$ & ${\bm{79.2}}$\\
\verb|all->hard| & ${\bm{75.8}}$ & ${75.5}$ & ${75.3}$ & ${74.6}$ & & ${\bm{79.4}}$ & ${\bm{79.3}}$ & ${79.2}$ & ${78.9}$ && ${75.3}$ & ${77.8}$ & ${78.5}$ & ${\bm{79.3}}$\\
\midrule
\verb|hard->all| & ${72.6}$ & ${71.1}$ & ${68.8}$ & ${67.7}$ & & ${76.7}$ & ${75.6}$ & ${76.7}$ & ${75.5}$ && ${74.2}$ & ${76.1}$ & ${79.0}$ & ${78.7}$\\
\verb|hard->easy| & ${72.5}$ & ${71.3}$ & ${68.7}$ & ${67.7}$ & & ${76.7}$ & ${75.9}$ & ${77.0}$ & ${75.3}$ && ${74.2}$ & ${76.2}$ & ${78.8}$ & ${78.6}$\\
\verb|hard->hard| & ${72.2}$ & ${71.4}$ & ${68.9}$ & ${67.7}$ & & ${76.8}$ & ${75.7}$ & ${76.9}$ & ${75.7}$ && ${73.9}$ & ${76.2}$ & ${79.0}$ & ${78.6}$\\
\bottomrule[1.5pt]
\end{tabular}

    }
    \caption{\textbf{The effect of curriculum.} We observe that the beginning 
    of the curriculum has much more impact on the generalization than the end 
    of the curriculum (accuracy within the groups of three rows is similar).  
    Accuracies within 0.2\% of the best accuracy in each column are 
    highlighted.}
    \label{tab:effect_curr}
    \vspace{-4mm}
\end{table*}

\subsection{Optimal augmentation sample selection}\label{sec:optimal_augs}

\begin{figure*}[b!]
\begin{center}
    \begin{subfigure}[t]{0.27\textwidth}
        \includegraphics[width=\textwidth]{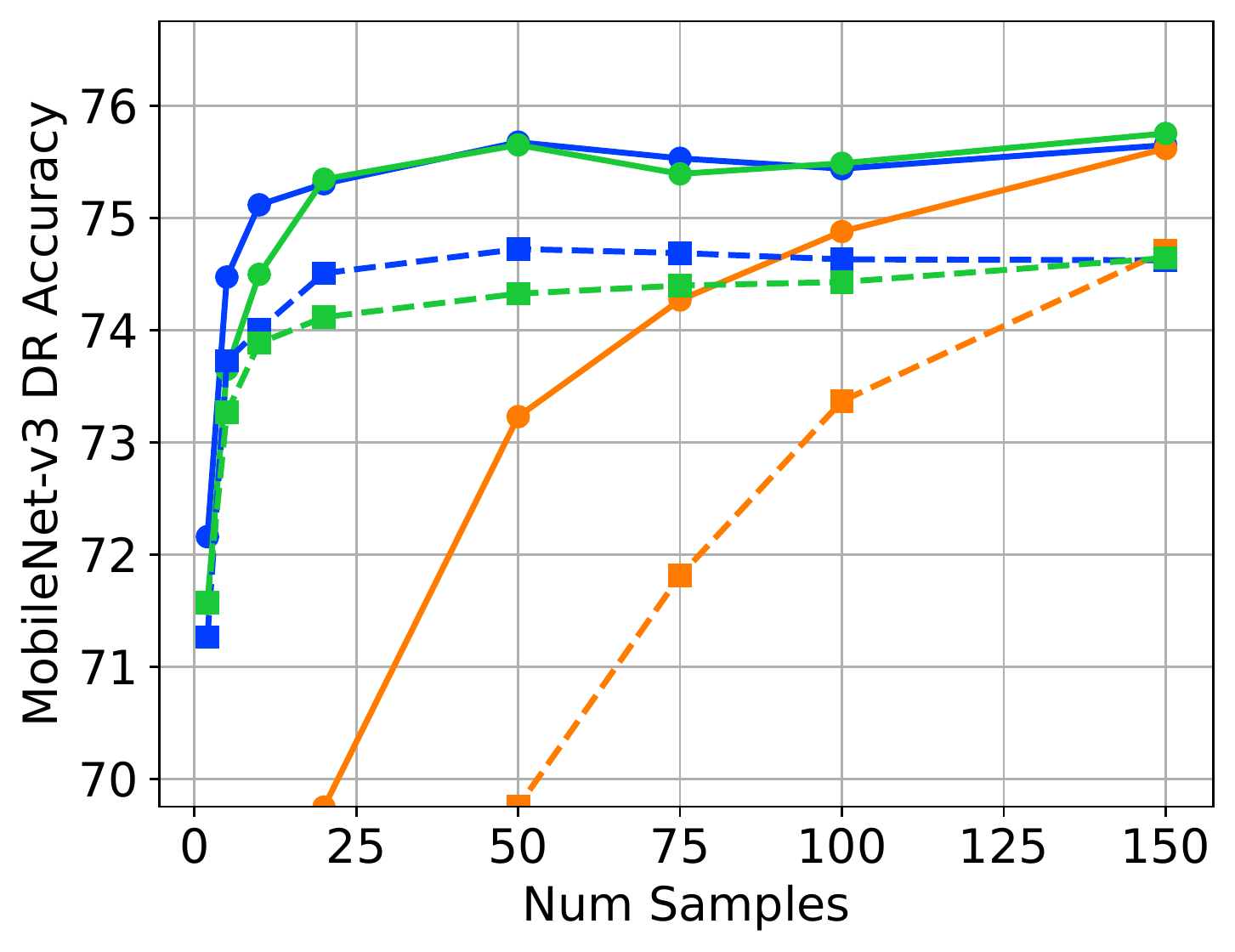}
        \caption{MobileNetV3}\label{fig:imagenet_MobileNetv3_optimal_e150}
    \end{subfigure}
    \hfill
    \begin{subfigure}[t]{0.27\textwidth}
        \includegraphics[width=\textwidth]{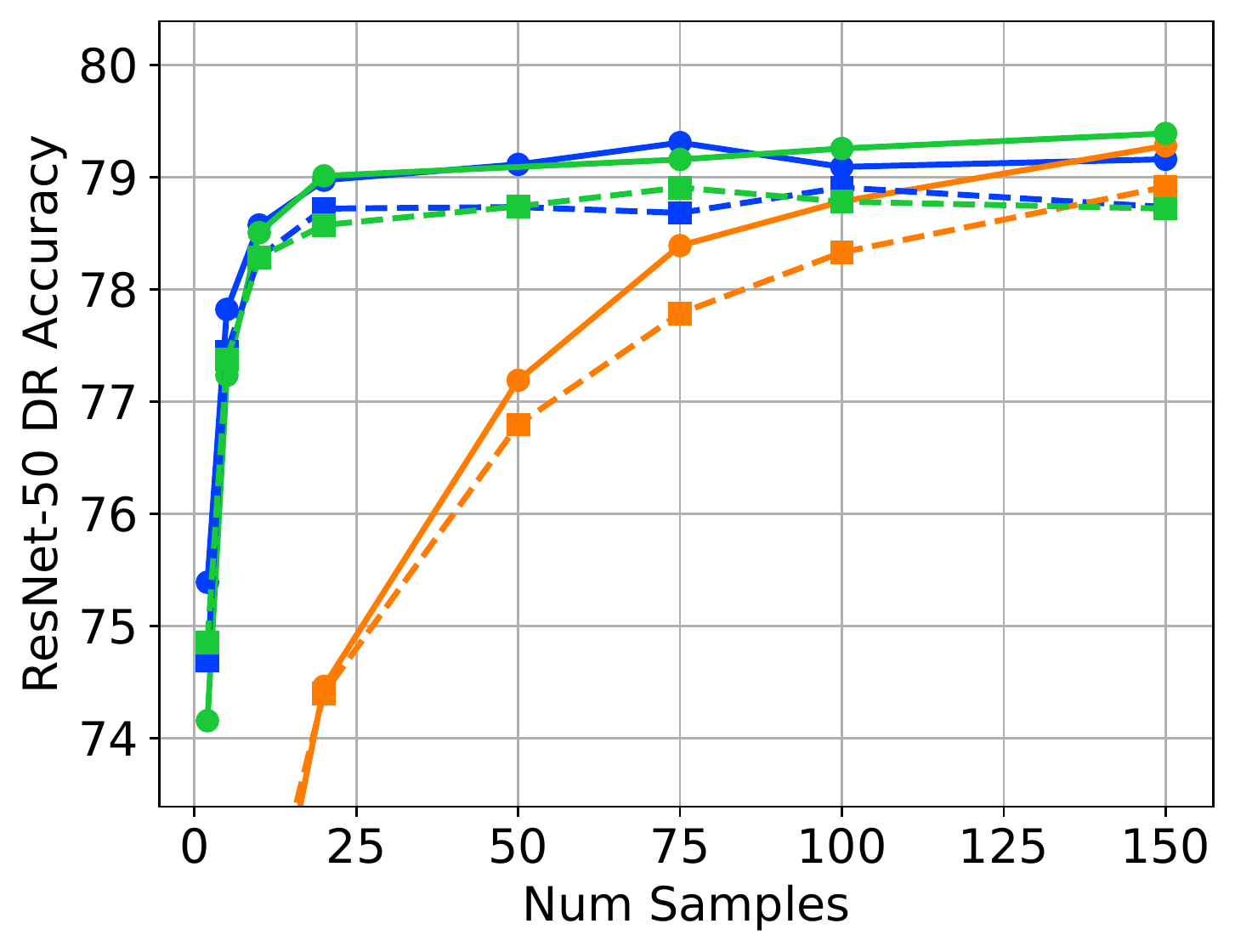}
        \caption{ResNet50}\label{fig:imagenet_R50_optimal_e150}
    \end{subfigure}
    \hfill
    \begin{subfigure}[t]{0.27\textwidth}
        \includegraphics[width=\textwidth]{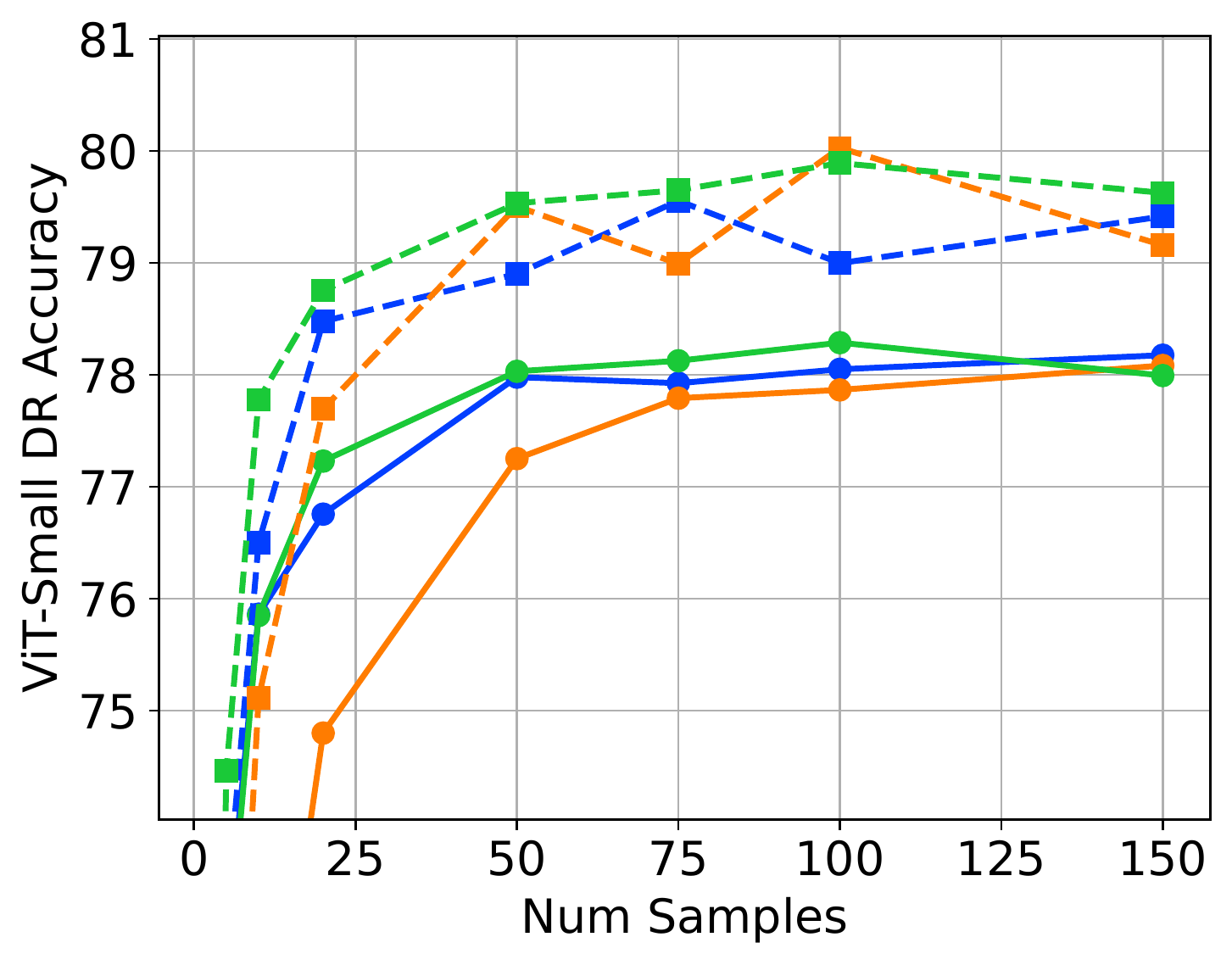}
        \caption{ViT-Small}\label{fig:imagenet_ViTSmall_optimal_e150}
    \end{subfigure}
    \hfill
    \begin{minipage}[t]{0.17\linewidth}
        \vspace{-2.9cm}
        \begin{subfigure}[t]{\textwidth}
            \includegraphics[width=\textwidth]{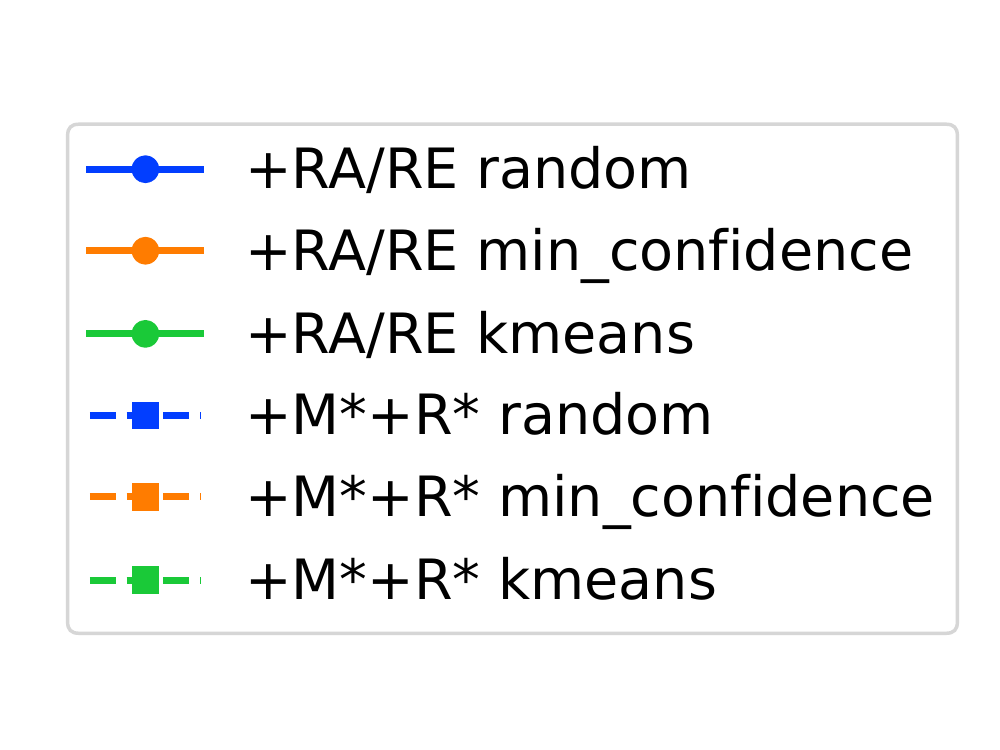}
        \end{subfigure}
    \end{minipage}
\end{center}
    \vspace{-0.5cm}
    \caption{\textbf{Optimal augmentation sample selection.} \INp{} accuracy 
    for varying objectives and number of samples. The teacher is ConvNext 
    (Base-IN22FT1K)  (E=150)}\label{fig:optimal_augs}
\end{figure*}

We discussed that augmentations used to reinforce the dataset are sampled from 
a pool of augmentation operations and that we apply the augmentations with 
a predetermined application probability. The setup of dataset reinforcement 
allows us to optimize for the most informative augmentation samples. For 
example, we can generate a large set of candidate augmentations and choose 
a subset with maximum or minimum values of ad-hoc metrics. We considered 
selecting samples according to metrics such as confidence, entropy, and loss.  
Given $\pp\in\R^{c}$, the set of predicted probabilities of the teacher for $c$ 
classes, we define confidence as $\max \pp_j$, the entropy as $-\sum_j 
\pp_j\log\pp_j$, and the loss as $-\log\pp_{y}$ where $y$ is the ground-truth 
label. To encourage diversity, we also considered selecting samples based on 
the clustering of the predicted probability vectors by performing KMeans on 
$\pp$ vectors of the candidates and selecting one sample per cluster.  
\Cref{fig:optimal_augs} shows the performance of a subset of sample selection 
methods we considered. 

Generally we observe that max-entropy/min-confidence objectives demonstrate 
similar behaviors better than min-entropy/max-confidence. So we only show the 
min-confidence variant. We observe that overall random samples (blue lines) 
provides the best validation accuracy if used with the right augmentations 
(\RRCRARE{} for light-weight CNNs and \RRCMsRs{} for transformers).  Using 
min-confidence (orange) with \RRCMsRs{} (dashed orange), leads to similar 
generalization on transformers while hurting the generalization on CNNs. This 
matches our observations with the complexity of augmentations and curriculums 
that transformers prefer difficult samples. We observe that diversified samples 
using KMeans clustering (green) provide similar behavior to random samples 
(blue) while for transformers provide more consistent improvements at varying 
number of samples (dashed green compared with dashed blue). We identify this 
potential for future work and investigate reinforced datasets with random 
samples in the rest of the paper. Note that the curriculums are 
a generalization of the objective-based metrics that are adaptive to the 
student (See \cref{sec:curriculum} and \cref{sec:curriculum_extra}).

\newpage
\clearpage
\section{Additional pretraining/finetuning/transfer learning 
results}\label{sec:transfer_full}
In \cref{tab:transfer_cls_full} we provide results for various combinations of 
pretraining and fine-tuning on reinforced/non-reinforced datasets. We observe 
that the best results are achieved when both pretraining and fine-tuning are 
done using reinforced datasets. We also observe that the improvement is 
significant compared to when only one of the pretraining/fine-tuning datasets 
is reinforced. The idea of training and fine-tuning on multiple reinforced 
datasets is unique to dataset reinforcement and would be challenging to 
replicate with standard data augmentations or knowledge distillation.

We train models for 100, 400, 1000 epochs on \CIFAR{}, \Food{} and 1000, 
4000, and 10000 epochs on \Flowers{} and report the best accuracy for each 
model. Models pretrained/fine-tuned on non-reinforced datasets tend to overfit 
at longer training while models trained on reinforced datasets benefit from 
longer training.

For future tasks and datasets,  additional task-specific information could be 
considered as reinforcements. For example, an object detection dataset can be 
further reinforced using the teacher's uncertainty on bounding boxes, occlusion 
estimate, and border uncertainty.  Multi-modal models such as CLIP are an 
immediate future work that can provide variety of additional training signal 
based on the relation to an anchor text.

\begin{table}[tbh!]
    \centering
    \resizebox{\textwidth}{!}{
        \begin{tabular}{llcc|cc|cc}
            \toprule[1.5pt]
            \multirow{2}{*}{\textbf{Model}} 
            & \multicolumn{1}{c}{\multirow{2}{*}{\textbf{Pretraining dataset}}} 
            & \multicolumn{6}{c}{\textbf{Fine-tuning dataset}}  \\
            \cmidrule[1.25pt]{3-8}
             &  & \CIFAR{} & \CIFARp{} & \Flowers{} & \Flowersp{} & \Food{} & \Foodp{}\\
             \midrule[1.25pt]
             \multirow{2}{*}{MobileNetV3-Large}
             & None & 80.2 & 83.6 & 68.8 & 87.5 & 85.1 & 88.2\\
             & \IN{} &  84.4 & 87.2 & 92.5 & 94.1 & 86.1 & 89.2 \\
             & \INp{} (Ours) &  86.0 & \textbf{87.5} & 93.7 & \textbf{95.3} & 86.6 &\textbf{89.5} \\
             \midrule
             \multirow{2}{*}{ResNet-50}
             & None & 83.8 & 85.0 & 87.3 & 85.0 & 89.1 & 90.2\\
             & \IN{} & 88.4 & 89.5 &  93.6 & 94.9 & 90.0 & 91.8\\
             & \INp{} (Ours) & 88.8 & \textbf{89.8} & 95.0 & \textbf{96.3} & 90.5 & \textbf{92.1}\\
             \midrule
             \multirow{2}{*}{SwinTransformer-Tiny}
             & None & 35.0 & 82.2 & 78.3 & 72.5 & 89.6 & 90.9\\
             & \IN{} & 90.6 & 90.7 & 96.3 & 96.5 &  92.3 & 92.7\\
             & \INp{} (Ours) & 90.9 & \textbf{91.2} & 96.6 & \textbf{97.0} & \textbf{93.0} & \textbf{92.9}\\
            \bottomrule[1.5pt]
        \end{tabular}
    }
    \vspace{-2mm}
    \caption{\textbf{Pretraining/Finetuning/Transfer learning for fine-grained 
    object classification.}}
    \label{tab:transfer_cls_full}
    \vspace{-4mm}
\end{table}

\section{Full table of calibration results}\label{sec:val_calib_error_full}

In \cref{tab:val_calib_error_full} we provide the full results for 
\cref{fig:val_calib_error}. We see observe that validation ECE of \INp{} 
pretrained models is lower than \IN{} pretrained models.

\begin{table}[htb!]
    \centering
    \resizebox{1.0\textwidth}{!}{
        \begin{tabular}{lcccccccc}
            \toprule[1.5pt]
            \textbf{Model}
            & \textbf{Method}
            & \textbf{Epochs}
            & \textbf{Train ECE}
            & \textbf{Val ECE}
            & \textbf{ECE gap}
            & \textbf{Train Error}
            & \textbf{Val Error}
            & \textbf{Error gap}
            \\
             \midrule[1.25pt]
             \multirow{5}{*}{MobileNetV3-Large}
&\IN{}               &  300    &      0.1503 &    0.0727 &    0.0776 &        0.0934 &      0.2509 &      0.1575 \\
&\INp{}              &  300    &      0.0339 &    0.0309 &    0.0030 &        0.1400 &      0.2298 &      0.0898 \\
&\IN{}               &  1000   &      0.1489 &    0.0608 &    0.0881 &        0.0599 &      0.2491 &      0.1891 \\
&\INp{}              &  1000   &      0.0312 &    0.0323 &    0.0011 &        0.1218 &      0.2206 &      0.0988 \\
&KD                     &  300    &      0.0303 &    0.0297 &    0.0006 &        0.1550 &      0.2358 &      0.0808 \\
\midrule
             \multirow{5}{*}{ResNet-50}
&\IN{}               &  300    &      0.1938 &    0.1513 &    0.0425 &        0.1239 &      0.2122 &      0.0883 \\
&\INp{}              &  300    &      0.0263 &    0.0362 &    0.0098 &        0.1115 &      0.1944 &      0.0829 \\
&\IN{}               &  1000   &      0.1887 &    0.1348 &    0.0539 &        0.0906 &      0.2036 &      0.1130 \\
&\INp{}              &  1000   &      0.0241 &    0.0360 &    0.0119 &        0.0936 &      0.1830 &      0.0894 \\
&KD                     &  300    &      0.0250 &    0.0339 &    0.0089 &        0.1065 &      0.1846 &      0.0781 \\
\midrule
             \multirow{5}{*}{SwinTransformer-Tiny}
&\IN{}               &  300    &      0.1084 &    0.0663 &    0.0421 &        0.0734 &      0.1910 &      0.1176 \\
&\INp{}              &  300    &      0.0201 &    0.0381 &    0.0180 &        0.0818 &      0.1698 &      0.0880 \\
&\IN{}               &  1000   &      0.1042 &    0.0522 &    0.0519 &        0.0421 &      0.1905 &      0.1484 \\
&\INp{}              &  1000   &      0.0195 &    0.0397 &    0.0203 &        0.0743 &      0.1621 &      0.0877 \\
&KD                     &  300    &      0.0206 &    0.0379 &    0.0173 &        0.0958 &      0.1701 &      0.0742 \\
            \bottomrule[1.5pt]
        \end{tabular}
    }
    \vspace{-2mm}
    \caption{Full calibration error and validation error for \cref{fig:val_calib_error}.}
    \label{tab:val_calib_error_full}
    \vspace{-4mm}
\end{table}

\section{Cost of dataset reinforcement}
\label{sec:cost}
In \cref{sec:aug_difficulty_sup}, we observe that similar accuracy to knowledge 
distillation is reached with $\times 3$ fewer samples than the number of target 
epochs. This reduces the reinforcement cost.
\INp{} took 2080 mins to generate using 64xA100 GPUs which is highly 
parallelizable and similar to training ResNet-50 for 300 epochs on 8xA100 GPUs.  
The parallelization is another significant advantage to knowledge distillation 
because samples are reinforced independently while knowledge distillation 
requires following a trajectory on training samples.
For \CIFAR{}, \Flowers{}, and \Food{}, the reinforcement took 90, 40, and 
120 minutes respectively. With pretrained teachers and extrapolating our 
\INp{} observations, we can reinforce any new dataset and the cost is 
performing inference using the teacher on the dataset for
approximately $\times 3$ fewer samples than the maximum intended training epochs.  
This is a one-time cost that is amortized over many uses.

We provide storage cost analysis for \INp{} in 
\cref{tab:imagenet_plus_specs}. Note that for variants with mixing, the storage 
of \RRCRARE{} parameters doubles because each reinforcement consists of 
augmentations for a pair. The proposed \INp{} variant, \RRCRARE{}, does not 
have that doubling cost. Also note that the storage can be further reduced 
using compression methods. For example, \INp{} \RRCRARE{} with the 
compression from Python's Joblib with compression level 3 can be reduced to 
55GBs instead of 61GBs. Even more compression is possible by reducing the 
number of stored logits for the teacher and more aggressive compression 
methods.

The storage cost for \CIFARp{}, \Flowersp{}, and \Foodp{} uses the same set 
of formula given the number of samples that amounts to approximately 4.8, 1.0, 
and 7.3GBs in basic compressed form as in \cref{tab:imagenet_plus_specs}.
We have not explored reducing the size of these datasets significantly as it is 
not a significant overhead for small datasets. For larger datasets such as 
\IN{}, the reinforcement overhead is much smaller relative to the original 
dataset size because the bulk of the dataset is taken by the inputs while our 
reinforcements only store the outputs.

We provide the breakdown of training time on MobileNetV3-Large, ResNet-50, and 
SwinTransformer-Tiny in \cref{tab:train_time}.
Except for mixing augmentations, reapplying all augmentations has zero overhead 
compared to standard training with the same augmentations.  For mixing 
augmentations, our current implementation has approximately $30\%$ time 
overhead because of the extra load time of mixing pairs stored with each 
reinforced sample. This overhead only translate to extra wall-clock for very 
small models where the bottleneck is on the CPU rather than {GPU}.
We discuss efficient alternatives in \cref{sec:library}.  Our balanced 
solution, \RRCRARE{}, does not use mixing and has zero overhead.

\begin{table}[htb!]
    \centering
    \resizebox{0.45\textwidth}{!}{
        \begin{tabular}{llccc}
            \toprule[1.5pt]
            \multirow{2}{*}{\textbf{Model}} & \multirow{2}{*}{\textbf{Dataset}} & \multicolumn{3}{c}{\textbf{Training Epochs}}  \\
            \cmidrule[1.25pt]{3-5}
            & & \textbf{150} & \textbf{300} & \textbf{1000} \\
             \midrule[1.25pt]
             \multirow{2}{*}{MobileNetV3-Large} & \IN{} & $1.00\times$ & $1.00\times$ & $1.00\times$ \\
             & \INp{} (Ours) & $1.13\times$ & $1.12\times$ & $1.12\times$ \\
             \midrule
             \multirow{2}{*}{ResNet-50} & \IN{} & $1.00\times$ & $1.00\times$ & $1.00\times$ \\
             & \INp{} (Ours) & $1.04 \times$ & $1.02 \times$ & $0.97\times$ \\
             \midrule
             \multirow{2}{*}{SwinTransformer-Tiny} & \IN{} & $1.00\times$ & $1.00\times$ & $1.00\times$ \\
              & \INp{} (Ours) & $0.99\times$ & $0.99\times$ & $0.99\times$ \\
            \bottomrule[1.5pt]
        \end{tabular}
    }
    \vspace{-2mm}
    \caption{\textbf{Training time for different models using \INp{} is 
    similar to \IN{} dataset}. Full results in 
    \cref{sec:imagenet_plus_full}.  }
    \label{tab:train_time}
    \vspace{-4mm}
\end{table}

\newpage
\clearpage
\section{Hyperparameters and implementation details}\label{sec:hparams}
We follow \cite{mehta2022cvnets, wightman2021resnet} and use  state-of-the-art 
recipes, including optimizers, hyperparameters, and learning. The details are 
provided in \cref{tab:train_hparams}. Because of resource limitations, we train 
EfficientNet-B3/B4 with KD using batch size $512$. Overall, we use the same 
hyperparameters on \IN{} and \INp{} with the exception of the data 
augmentations that are removed from the training on \INp{} based on our 
observations in \cref{sec:invariance}. For KD, we use the KL loss with 
temperature $1.0$ (no mixing with the cross-entropy loss) and shrink the weight 
decay by $10\times$.

In \cref{tab:train_hparams_cvnets} we provide hyperparameters for training with CVNets. For higher resolution and variable resolution training, we use the same metadata in \INp{} to create a random crop then resize it to the target resolution instead of the base resolution of 224.
In \cref{tab:det_seg_hparams} we provide hyperparameters for training Detection/Segmentation models.
In \cref{tab:transfer_hparams} we provide hyperparameters for transfer learning on \CIFAR{}/\Flowers{}/\Food{} datasets.

\begin{table}[h!]
    \centering
    \resizebox{\columnwidth}{!}{
        \begin{tabular}{ll cccccc cccc}
            \toprule[1.5pt]
           \multirow{2}{*}{\textbf{Model}} & \multirow{2}{*}{\bfseries Training Method} & \multicolumn{6}{c}{\bfseries Optimization Hyperparams} & \multicolumn{4}{c}{\bfseries Data augmentation methods} \\ 
           \cmidrule[1.25pt]{3-8}
           \cmidrule[1.25pt]{9-12}
            & & \textbf{Optimizer} & \textbf{Batch Size} & \textbf{LR} & \textbf{Warmup} & \textbf{Weight Decay} & \textbf{Label Smoothing}
            & \textbf{RandAugment} & \textbf{Random Erase} ($p$) & \textbf{MixUp} ($\alpha$) & \textbf{CutMix} ($\alpha$)\\
           \midrule[1.25pt]
            \multirow{3}{*}{\bfseries MobileNetV1}
            & \IN{}  & SGD+Mom=0.9 & 1024 & 0.8 & 3 & $4.0\mathrm{e}{-5}$ & \cmark & \xmark & \xmark & \xmark & \xmark \\
            & \INp{} & SGD+Mom=0.9 & 1024 & 0.8 & 3 & $4.0\mathrm{e}{-5}$ & \cmark & \xmark & \xmark & \xmark & \xmark \\
            & KD        & SGD+Mom=0.9 & 1024 & 0.8 & 3 & $4.0\mathrm{e}{-6}$ & \xmark & \xmark & \xmark & \xmark & \xmark \\
            \midrule
            \multirow{3}{*}{\bfseries MobileNetV2}
            & \IN{}  & SGD+Mom=0.9 & 1024 & 0.4 & 3 & $4.0\mathrm{e}{-5}$ & \cmark & \xmark & \xmark & \xmark & \xmark \\
            & \INp{} & SGD+Mom=0.9 & 1024 & 0.4 & 3 & $4.0\mathrm{e}{-5}$ & \cmark & \xmark & \xmark & \xmark & \xmark \\
            & KD        & SGD+Mom=0.9 & 1024 & 0.4 & 3 & $4.0\mathrm{e}{-6}$ & \xmark & \xmark & \xmark & \xmark & \xmark \\
            \midrule
            \multirow{3}{*}{\bfseries MobileNetV3}
            & \IN{}  & SGD+Mom=0.9 & 1024 & 0.4 & 3 & $4.0\mathrm{e}{-5}$ & \cmark & \xmark & \xmark & \xmark & \xmark \\
            & \INp{} & SGD+Mom=0.9 & 1024 & 0.4 & 3 & $4.0\mathrm{e}{-5}$ & \cmark & \xmark & \xmark & \xmark & \xmark \\
            & KD        & SGD+Mom=0.9 & 1024 & 0.4 & 3 & $4.0\mathrm{e}{-6}$ & \xmark & \xmark & \xmark & \xmark & \xmark \\
            \midrule
            \multirow{3}{*}{\bfseries ResNet}
            & \IN{}  & SGD+Mom=0.9 & 1024 & 0.4 & 5 & $1.0\mathrm{e}{-4}$ & \cmark & \cmark & \cmark (0.25) & \cmark (0.2) & \cmark (1.0) \\
            & \INp{} & SGD+Mom=0.9 & 1024 & 0.4 & 5 & $1.0\mathrm{e}{-4}$ & \cmark & \xmark & \xmark & \xmark & \xmark \\
            & KD        & SGD+Mom=0.9 & 1024 & 0.4 & 5 & $1.0\mathrm{e}{-5}$ & \xmark & \cmark & \cmark (0.25) & \cmark (1.0) & \cmark (1.0) \\
            \midrule
            \multirow{3}{*}{\bfseries EfficientNet-B2}
            & \IN{}  & SGD+Mom=0.9 & 1024 & 0.4 & 5 & $4.0\mathrm{e}{-5}$ & \cmark & \cmark & \cmark (0.25) & \cmark (0.2) & \cmark (1.0) \\
            & \INp{} & SGD+Mom=0.9 & 1024 & 0.4 & 5 & $4.0\mathrm{e}{-5}$ & \cmark & \xmark & \xmark & \xmark & \xmark \\
            & KD        & SGD+Mom=0.9 & 1024 & 0.4 & 5 & $4.0\mathrm{e}{-6}$ & \xmark & \cmark & \cmark (0.25) & \cmark (1.0) & \cmark (1.0) \\
            \midrule
            \multirow{3}{*}{\bfseries EfficientNet-B3}
            & \IN{}  & SGD+Mom=0.9 & 1024 & 0.4 & 5 & $4.0\mathrm{e}{-5}$ & \cmark & \cmark & \cmark (0.25) & \cmark (0.2) & \cmark (1.0) \\
            & \INp{} & SGD+Mom=0.9 & 1024 & 0.4 & 5 & $4.0\mathrm{e}{-5}$ & \cmark & \xmark & \xmark & \xmark & \xmark \\
            & KD        & SGD+Mom=0.9 & 512 & 0.2 & 5 & $4.0\mathrm{e}{-6}$ & \xmark & \cmark & \cmark (0.25) & \cmark (1.0) & \cmark (1.0) \\
            \midrule
            \multirow{3}{*}{\bfseries EfficientNet-B4}
            & \IN{}  & SGD+Mom=0.9 & 1024 & 0.4 & 5 & $4.0\mathrm{e}{-5}$ & \cmark & \cmark & \cmark (0.25) & \cmark (0.2) & \cmark (1.0) \\
            & \INp{} & SGD+Mom=0.9 & 1024 & 0.4 & 5 & $4.0\mathrm{e}{-5}$ & \cmark & \xmark & \xmark & \xmark & \xmark \\
            & KD        & SGD+Mom=0.9 & 512 & 0.4 & 5 & $4.0\mathrm{e}{-6}$ & \xmark & \cmark & \cmark (0.25) & \cmark (1.0) & \cmark (1.0) \\
            \midrule
            \multirow{3}{*}{\bfseries ViT}
            & \IN{}  & AdamW (0.9, 0.999) & 1024 & 0.001 & 5 & 0.05 & \cmark & \cmark & \cmark (0.25) & \cmark (0.2) & \cmark (1.0) \\
            & \INp{} & AdamW (0.9, 0.999) & 1024 & 0.001 & 5 & 0.05 & \cmark & \xmark & \xmark & \xmark & \xmark \\
            & KD        & AdamW (0.9, 0.999) & 1024 & 0.001 & 5 & 0.005 & \xmark & \cmark & \cmark (0.25) & \cmark (1.0) & \cmark (1.0) \\
            \midrule
            \multirow{3}{*}{\bfseries SwinTransformer}
            & \IN{}  & AdamW (0.9, 0.999) & 1024 & 0.001 & 5 & 0.05 & \cmark & \cmark & \cmark (0.25) & \cmark (0.2) & \cmark (1.0) \\
            & \INp{} & AdamW (0.9, 0.999) & 1024 & 0.001 & 5 & 0.05 & \cmark & \xmark & \xmark & \xmark & \xmark \\
            & KD        & AdamW (0.9, 0.999) & 1024 & 0.001 & 5 & 0.005 & \xmark & \cmark & \cmark (0.25) & \cmark (1.0) & \cmark (1.0) \\
           \bottomrule[1.5pt]
        \end{tabular}
    }
    \caption{\textbf{Hyperparameters used for training different models.} We use cosine learning rate schedule to zero.}
    \label{tab:train_hparams}
\end{table}

\begin{table}[h!]
    \centering
    \resizebox{\columnwidth}{!}{
        \begin{tabular}{ll cccccccc c}
            \toprule[1.5pt]
           \multirow{2}{*}{\textbf{Model}} & \multirow{2}{*}{\bfseries Training Method} & \multicolumn{8}{c}{\bfseries Optimization Hyperparams} & \multirow{2}{*}{\bfseries Data augmentation methods} \\ 
           \cmidrule[1.25pt]{3-10}
            & & \textbf{Optimizer} & \textbf{Batch Size} & \textbf{LR} & \textbf{Warmup} & \textbf{Weight Decay} & \textbf{Mixed Precision} & \textbf{Resolution} & \textbf{Grad. Clip} & \\
           \midrule[1.25pt]
            \multirow{2}{*}{\bfseries MobileNetV1}
            & \IN{}  & SGD+Mom=0.9 & 1024 & 0.8 & 3 & $4.0\mathrm{e}{-5}$ & \cmark & 224 & \xmark & LS+RRC+HF\\
            & \INp{} & SGD+Mom=0.9 & 1024 & 0.8 & 3 & $4.0\mathrm{e}{-5}$ & \cmark & 224 & \xmark & \xmark\\
            \midrule
            \multirow{2}{*}{\bfseries MobileNetV2}
            & \IN{}  & SGD+Mom=0.9 & 1024 & 0.4 & 3 & $4.0\mathrm{e}{-5}$ & \cmark & 224 & \xmark & LS+RRC+HF\\
            & \INp{} & SGD+Mom=0.9 & 1024 & 0.4 & 3 & $4.0\mathrm{e}{-5}$ & \cmark & 224 & \xmark & \xmark\\
            \midrule
            \multirow{2}{*}{\bfseries MobileNetV3}
            & \IN{}  & SGD+Mom=0.9 & 2048 & 0.4 & 3 & $4.0\mathrm{e}{-5}$ & \cmark & 224 & \xmark & LS+RRC+HF\\
            & \INp{} & SGD+Mom=0.9 & 2048 & 0.4 & 3 & $4.0\mathrm{e}{-5}$ & \cmark & 224 & \xmark & \xmark\\
            \midrule
            \multirow{2}{*}{\bfseries MobileNetViT}
            & \IN{}  & AdamW (0.9, 0.999) & 1024 & 0.002 & 20 & $0.01$ & \cmark & VBS(160, 320, 256) & \xmark & LS+RRC+HF\\
            & \INp{} & AdamW (0.9, 0.999) & 1024 & 0.002 & 20 & $0.01$ & \cmark & VBS(160, 320, 256) & \xmark & \xmark\\
            \midrule
            \multirow{2}{*}{\bfseries ResNet}
            & \IN{}  & SGD+Mom=0.9 & 1024 & 0.4 & 5 & $1.0\mathrm{e}{-4}$ & \cmark & 224 & \xmark & LS+RRC+HF+RA+RE+MU+CM\\
            & \INp{} & SGD+Mom=0.9 & 1024 & 0.4 & 5 & $1.0\mathrm{e}{-4}$ & \cmark & 224 & \xmark & \xmark\\
            \midrule
            \multirow{2}{*}{\bfseries EfficientNet-B2}
            & \IN{}  & SGD+Mom=0.9 & 2048 & 0.8 & 3 & $4.0\mathrm{e}{-5}$ & \cmark & VBS(144, 432, 288) & \xmark & LS+RRC+HF+RA+RE+MU+CM\\
            & \INp{} & SGD+Mom=0.9 & 2048 & 0.8 & 3 & $4.0\mathrm{e}{-5}$ & \cmark & VBS(144, 432, 288) & \xmark & \xmark\\
            \midrule
            \multirow{2}{*}{\bfseries EfficientNet-B3}
            & \IN{}  & SGD+Mom=0.9 & 2048 & 0.8 & 3 & $4.0\mathrm{e}{-5}$ & \cmark & VBS(150, 450, 300) & \xmark & LS+RRC+HF+RA+RE+MU+CM\\
            & \INp{} & SGD+Mom=0.9 & 2048 & 0.8 & 3 & $4.0\mathrm{e}{-5}$ & \cmark & VBS(150, 450, 300) & \xmark & \xmark\\
            \midrule
            \multirow{2}{*}{\bfseries EfficientNet-B4}
            & \IN{}  & SGD+Mom=0.9 & 2048 & 0.8 & 3 & $4.0\mathrm{e}{-5}$ & \cmark & VBS(190, 570, 380) & \xmark & LS+RRC+HF+RA+RE+MU+CM\\
            & \INp{} & SGD+Mom=0.9 & 2048 & 0.8 & 3 & $4.0\mathrm{e}{-5}$ & \cmark & VBS(190, 570, 380) & \xmark & \xmark\\
            \midrule
            \multirow{2}{*}{\bfseries ViT-Tiny}
            & \IN{}  & AdamW (0.9, 0.999) & 2048 & 0.002 & 10 & 0.05 & \cmark & 224 & \cmark (1.0) & LS+RRC+HF+RA+RE+MU+CM\\
            & \INp{} & AdamW (0.9, 0.999) & 2048 & 0.002 & 10 & 0.05 & \cmark & 224 & \cmark (1.0) & \xmark\\
            \midrule
            \multirow{2}{*}{\bfseries ViT-Small/Base}
            & \IN{}  & AdamW (0.9, 0.999) & 2048 & 0.002 & 10 & 0.2 & \cmark & 224 & \cmark (1.0) & LS+RRC+HF+RA+RE+MU+CM\\
            & \INp{} & AdamW (0.9, 0.999) & 2048 & 0.002 & 10 & 0.2 & \cmark & 224 & \cmark (1.0) & \xmark\\
            \midrule
            \multirow{2}{*}{\bfseries ViT-Base $\uparrow$384}
            & \IN{}  & AdamW (0.9, 0.999) & 2048 & 0.002 & 20 & 0.2 & \cmark & VBS(192, 576, 384) & \cmark (1.0) & LS+RRC+HF+RA+RE+MU+CM\\
            & \INp{} & AdamW (0.9, 0.999) & 2048 & 0.002 & 20 & 0.2 & \cmark & VBS(192, 576, 384) & \cmark (1.0) & \xmark\\
            \midrule
            \multirow{2}{*}{\bfseries SwinTransformer}
            & \IN{}  & AdamW (0.9, 0.999) & 1024 & 0.001 & 20 & 0.05 & \cmark & 224 & \cmark (5.0) & LS+RRC+HF+RA+RE+MU+CM\\
            & \INp{} & AdamW (0.9, 0.999) & 1024 & 0.001 & 20 & 0.05 & \cmark & 224 & \cmark (5.0) & \xmark\\
            \midrule
            \multirow{2}{*}{\bfseries SwinTransformer-Base $\uparrow$384}
            & \IN{}  & AdamW (0.9, 0.999) & 1024 & 0.001 & 20 & 0.05 & \cmark & VBS(192, 576, 384) & \cmark (5.0) & LS+RRC+HF+RA+RE+MU+CM\\
            & \INp{} & AdamW (0.9, 0.999) & 1024 & 0.001 & 20 & 0.05 & \cmark & VBS(192, 576, 384) & \cmark (5.0) & \xmark\\
           \bottomrule[1.5pt]
        \end{tabular}
    }
    \caption{\textbf{Hyperparameters used for training different models in 
    CVNets.} LS: Label Smoothing with 0.1, RRC: Random-Resize-Crop, HF: Horizontal Flip, VBS(min-res, max-res, crop-size): Variable Batch Sampler with variable resolution. RA: RandAugment, RE: Random Erase with 0.25, MU: MixUp with alpha
0.2, CM: CutMix with alpha 0.1. We use cosine-learning rate schedule to 0.}
    \label{tab:train_hparams_cvnets}
\end{table}

\begin{table}[h!]
    \centering
    \resizebox{\columnwidth}{!}{
        \begin{tabular}{ll cccccccccc c}
            \toprule[1.5pt]
           \multirow{2}{*}{\textbf{Model}} & \multirow{2}{*}{\bfseries Training Method} & \multicolumn{10}{c}{\bfseries Optimization Hyperparams} & \multirow{2}{*}{\bfseries Data augmentation methods} \\ 
           \cmidrule[1.25pt]{3-12}
            & & \textbf{Optimizer} & \textbf{Epochs} & \textbf{Batch Size} & \textbf{LR} & \textbf{BackBone LR Mul.} & \textbf{Warmup iter.} & \textbf{Weight Decay} & \textbf{Mixed Precision} & \textbf{Resolution} & \textbf{Grad. Clip} & \\
           \midrule[1.25pt]
            \multirow{2}{*}{\bfseries MobileNetV3-Large}
            & Detection    & SGD+Mom=0.9 & 36 & 64 & multi-step-lr(0.1, [24, 33])  & 0.1 & 500 & $4.0\mathrm{e}{-5}$ & \xmark & VBS(512, 1280, 1024) & \xmark & \xmark\\
            & Segmentation & SGD+Mom=0.9 & 50 & 16 & cosine-lr(0.02, 0.0001)       & 0.1 & 0   & $1.0\mathrm{e}{-4}$ & \xmark & 512 & \xmark & RC+RSSR+RR+PD+RG\\
            \midrule
            \multirow{2}{*}{\bfseries ResNet-50}
            & Detection    & SGD+Mom=0.9 & 100 & 64 & multi-step-lr(0.1, [60, 84]) & 0.1 & 500 & $4.0\mathrm{e}{-5}$ & \xmark & VBS(512, 1280, 1024) & \xmark & \xmark\\
            & Segmentation & SGD+Mom=0.9 & 50  & 16 & cosine-lr(0.02, 0.0001)      & 0.1 & 500 & $4.0\mathrm{e}{-5}$ & \xmark & 512 & \xmark & RC+RSSR+RR+PD+RG\\
            \midrule
            \multirow{2}{*}{\bfseries SwinTransformer-Tiny}
            & Detection    & SGD+Mom=0.9 & 100 & 64 & multi-step-lr($6.0\mathrm{e}{-4}$, [60, 84])        & 1.0 & 500 & 0.05 & \xmark & VBS(512, 1280, 1024) & \xmark & \xmark\\
            & Segmentation & SGD+Mom=0.9 & 50  & 16 & cosine-lr($6.0\mathrm{e}{-4}$, $1.0\mathrm{e}{-6}$) & 0.1 & 500 & 0.05 & \xmark & 512 & \xmark & RC+RSSR+RR+PD+RG\\
           \bottomrule[1.5pt]
        \end{tabular}
    }
    \caption{\textbf{Hyperparameters of detection/segmentation using CVNets.} RC: Random Crop, RSSR: Random Short-Size Resize, RR: Random Rotate by maximum 10 degrees angle. VBS(min-res, max-res, crop-size): Variable Batch Sampler with variable resolution. PD: Photometric Distortion, RG: Random Gaussian noise}
    \label{tab:det_seg_hparams}
\end{table}

\begin{table}[h!]
    \centering
    \resizebox{\columnwidth}{!}{
        \begin{tabular}{ll ccccc c}
            \toprule[1.5pt]
           \multirow{2}{*}{\textbf{Model}} & \multirow{2}{*}{\bfseries Pretrained} & \multicolumn{5}{c}{\bfseries Optimization Hyperparams} & \multirow{2}{*}{\bfseries Data augmentation methods} \\ 
           \cmidrule[1.25pt]{3-7}
            & & \textbf{Optimizer} & \textbf{Batch Size} & \textbf{LR} & \textbf{Warmup} & \textbf{Weight Decay} & \\
           \midrule[1.25pt]
            \multirow{3}{*}{\bfseries MobileNetV3-Large}
            & \xmark & SGD+Mom=0.9 & 256 & 0.2   & 0 & $5.0\mathrm{e}{-4}$ & \xmark \\
            & \cmark & SGD+Mom=0.9 & 256 & 0.002 & 0 & $5.0\mathrm{e}{-4}$ & \xmark \\
            & \cmark+ & SGD+Mom=0.9 & 256 & 0.002 & 0 & $5.0\mathrm{e}{-4}$ & \xmark \\
            \midrule
            \multirow{3}{*}{\bfseries ResNet-50}
            & \xmark & SGD+Mom=0.9 & 256 & 0.2   & 0 & $5.0\mathrm{e}{-4}$ & RA+MU+CM\\
            & \cmark & SGD+Mom=0.9 & 256 & 0.002 & 0 & $5.0\mathrm{e}{-4}$ & RA+MU+CM\\
            & \cmark+ & SGD+Mom=0.9 & 256 & 0.002 & 0 & $5.0\mathrm{e}{-4}$ & \xmark\\
            \midrule
            \multirow{3}{*}{\bfseries SwinTransformer-Tiny}
            & \xmark & AdamW (0.9, 0.999) & 256 & 0.0001 & 5 & 0.05 & RA+MU+CM\\
            & \cmark & AdamW (0.9, 0.999) & 256 & 0.00001 & 5 & 0.05 & RA+MU+CM\\
            & \cmark+ & AdamW (0.9, 0.999) & 256 & 0.00001 & 5 & 0.05 & \xmark\\
           \bottomrule[1.5pt]
        \end{tabular}
    }
    \caption{\textbf{Hyperparameters used for \CIFAR{}/\Flowers{}/\Food{}.} We use cosine learning rate schedule to zero. We resize the inputs for all datasets to 224 including \CIFAR{} where we pad the input by 16. We also use label smoothing.}
    \label{tab:transfer_hparams}
\end{table}

\newpage
\clearpage

\section{CLIP, ViT, and Mixed Architecture Teachers}\label{sec:clip_vit_mix}

In this section, we evaluate the effectiveness of CLIP-pretrained models 
fine-tuned on \IN{} as teachers.  We evaluate various ensembles teachers 
mixed with non-CILP pretrained teachers and a variety of ViT-based models. We 
provide the model names in \cref{tab:imagenet_plus_clip_vit_mixed_names}.
\Cref{tab:imagenet_plus_clip_vit_mixed} shows the accuracy of various student 
models trained on reinforced datasets with our selection of ensembles. We 
observe 1) Ensembles are consistently better teachers 2) CLIP-pretrained 
teachers are at best on-par with the IG-ResNext ensemble 3) ViT-based teachers 
are not good teachers for CNN-based models, regardless of their training 
method.

\begin{table*}[htb!]
\centering
\resizebox{0.95\textwidth}{!}{
    \begin{tabular}{l|cccc}
    \toprule[1.5pt]
    Teacher Name & \multicolumn{4}{c}{\textbf{Timm name of Ensemble Member}} \\
    \cmidrule[1.25pt]{2-5}
    & 1 & 2 & 3 & 4\\
    \midrule
    CLIP
    & \verb|vit_large_patch14_clip_224.openai_ft_in12k_in1k|
    & \verb|vit_large_patch14_clip_224.openai_ft_in1k|
    & \verb|vit_base_patch16_clip_224.openai_ft_in12k_in1k|
    & \verb|vit_base_patch16_clip_224.openai_ft_in1k|
    \\
    ViT
    & \verb|vit_base_patch16_224|
    & \verb|vit_base_patch8_224|
    & \verb|vit_large_patch16_224|
    & \verb|vit_small_patch32_224|
    \\
    Mixed (RCVDx4).
    & \verb|ig_resnext101_32x48d|
    & \verb|convnext_xlarge_in22ft1k|
    & \verb|volo_d5_224|
    & \verb|deit3_huge_patch14_224|
    \\
    Mixed (RCCVx4).
    & \verb|ig_resnext101_32x48d|
    & \verb|convnext_xlarge_in22ft1k|
    & \verb|vit_large_patch14_clip_224.openai_ft_in1k|
    & \verb|vit_base_patch16_224|
    \\
    \bottomrule
\end{tabular}

}
\caption{CLIP, ViT, Mixed architecture teacher ensemble names.}
\label{tab:imagenet_plus_clip_vit_mixed_names}
\end{table*}

\begin{table*}[thb!]
    \centering
    \begin{subtable}[b]{.95\textwidth}
        \centering
        \resizebox{\columnwidth}{!}{
            \begin{tabular}{l|cc|cc|ccc|cc}
\toprule
\multirow{2}{*}{\textbf{Model}} & \multicolumn{2}{c}{\textbf{Prev.}} & \multicolumn{2}{c}{\textbf{Mixed Archs}} &\multicolumn{3}{c}{\textbf{CLIP}} & \multicolumn{2}{c}{\textbf{ViT}}\\
\cmidrule[1.25pt]{2-10}
& \textbf{IN} & \textbf{IN+} & \textbf{IN+-RCVDx4} & \textbf{IN+-RCCVx4} & \textbf{IN+-CLIPx1} & \textbf{IN+-CLIPx2} & \textbf{IN+-CLIPx4} & \textbf{IN+-ViTx1} & \textbf{IN+-ViTx4}\\
\midrule[1.25pt]
\verb|MobileNetV3-Large| & ${74.7}$ & $\bm{76.2}_{\scriptscriptstyle +1.6}$ & ${75.9}_{\scriptscriptstyle +1.2}$ & ${75.9}_{\scriptscriptstyle +1.2}$ & ${75.5}_{\scriptscriptstyle +0.8}$ & ${75.5}_{\scriptscriptstyle +0.8}$ & ${75.5}_{\scriptscriptstyle +0.8}$ & ${74.3}_{\scriptscriptstyle -0.4}$ & ${74.0}_{\scriptscriptstyle -0.6}$ \\
\midrule
\verb|ResNet-50| & ${77.4}$ & $\bm{79.6}_{\scriptscriptstyle +2.3}$ & $\bm{79.5}_{\scriptscriptstyle +2.1}$ & $\bm{79.4}_{\scriptscriptstyle +2.0}$ & ${79.2}_{\scriptscriptstyle +1.8}$ & ${79.2}_{\scriptscriptstyle +1.8}$ & ${79.3}_{\scriptscriptstyle +2.0}$ & ${77.8}_{\scriptscriptstyle +0.4}$ & ${78.0}_{\scriptscriptstyle +0.6}$ \\
\midrule
\verb|Swin-Tiny| & ${79.9}$ & $\bm{82.0}_{\scriptscriptstyle +2.1}$ & $\bm{81.9}_{\scriptscriptstyle +2.0}$ & $\bm{82.0}_{\scriptscriptstyle +2.1}$ & ${81.6}_{\scriptscriptstyle +1.7}$ & ${81.6}_{\scriptscriptstyle +1.7}$ & $\bm{81.8}_{\scriptscriptstyle +1.9}$ & ${80.0}_{\scriptscriptstyle +0.0}$ & ${80.2}_{\scriptscriptstyle +0.3}$ \\
\bottomrule[1.5pt]
\end{tabular}

        }
        \caption{150 epochs}
        \label{tab:imagenet_plus_clip_vit_mixed_e150}
    \end{subtable}
    \vfill
    \begin{subtable}[b]{.95\columnwidth}
        \centering
        \resizebox{\columnwidth}{!}{
            \begin{tabular}{l|cc|cc|ccc|cc}
\toprule
\multirow{2}{*}{\textbf{Model}} & \multicolumn{2}{c}{\textbf{Prev.}} & \multicolumn{2}{c}{\textbf{Mixed Archs}} &\multicolumn{3}{c}{\textbf{CLIP}} & \multicolumn{2}{c}{\textbf{ViT}}\\
\cmidrule[1.25pt]{2-10}
& \textbf{IN} & \textbf{IN+} & \textbf{IN+-RCVDx4} & \textbf{IN+-RCCVx4} & \textbf{IN+-CLIPx1} & \textbf{IN+-CLIPx2} & \textbf{IN+-CLIPx4} & \textbf{IN+-ViTx1} & \textbf{IN+-ViTx4}\\
\midrule[1.25pt]
\verb|MobileNetV3-Large| & ${74.9}$ & $\bm{77.0}_{\scriptscriptstyle +2.1}$ & ${76.6}_{\scriptscriptstyle +1.7}$ & ${76.7}_{\scriptscriptstyle +1.7}$ & ${76.2}_{\scriptscriptstyle +1.3}$ & ${76.3}_{\scriptscriptstyle +1.4}$ & ${76.4}_{\scriptscriptstyle +1.5}$ & ${75.1}_{\scriptscriptstyle +0.2}$ & ${75.0}_{\scriptscriptstyle +0.1}$ \\
\midrule
\verb|ResNet-50| & ${78.8}$ & $\bm{80.6}_{\scriptscriptstyle +1.8}$ & $\bm{80.6}_{\scriptscriptstyle +1.9}$ & $\bm{80.4}_{\scriptscriptstyle +1.7}$ & ${80.0}_{\scriptscriptstyle +1.2}$ & ${80.1}_{\scriptscriptstyle +1.3}$ & ${80.3}_{\scriptscriptstyle +1.5}$ & ${78.5}_{\scriptscriptstyle -0.3}$ & ${78.6}_{\scriptscriptstyle -0.1}$ \\
\midrule
\verb|Swin-Tiny| & ${80.9}$ & $\bm{83.0}_{\scriptscriptstyle +2.1}$ & $\bm{82.9}_{\scriptscriptstyle +2.0}$ & $\bm{82.9}_{\scriptscriptstyle +2.0}$ & ${82.5}_{\scriptscriptstyle +1.6}$ & ${82.6}_{\scriptscriptstyle +1.7}$ & $\bm{82.9}_{\scriptscriptstyle +2.0}$ & ${80.7}_{\scriptscriptstyle -0.2}$ & ${81.0}_{\scriptscriptstyle +0.1}$ \\
\bottomrule[1.5pt]
\end{tabular}

        }
        \caption{300 epochs}
        \label{tab:imagenet_plus_clip_vit_mixed_e300}
    \end{subtable}
    \vfill
    \begin{subtable}[b]{.95\columnwidth}
        \centering
        \resizebox{\columnwidth}{!}{
            \begin{tabular}{l|cc|cc|ccc|cc}
\toprule
\multirow{2}{*}{\textbf{Model}} & \multicolumn{2}{c}{\textbf{Prev.}} & \multicolumn{2}{c}{\textbf{Mixed Archs}} &\multicolumn{3}{c}{\textbf{CLIP}} & \multicolumn{2}{c}{\textbf{ViT}}\\
\cmidrule[1.25pt]{2-10}
& \textbf{IN} & \textbf{IN+} & \textbf{IN+-RCVDx4} & \textbf{IN+-RCCVx4} & \textbf{IN+-CLIPx1} & \textbf{IN+-CLIPx2} & \textbf{IN+-CLIPx4} & \textbf{IN+-ViTx1} & \textbf{IN+-ViTx4}\\
\midrule[1.25pt]
\verb|MobileNetV3-Large| & ${75.1}$ & $\bm{77.9}_{\scriptscriptstyle +2.9}$ & $\bm{77.7}_{\scriptscriptstyle +2.6}$ & ${77.4}_{\scriptscriptstyle +2.3}$ & ${77.2}_{\scriptscriptstyle +2.1}$ & ${77.0}_{\scriptscriptstyle +1.9}$ & ${77.2}_{\scriptscriptstyle +2.1}$ & ${76.0}_{\scriptscriptstyle +0.9}$ & ${75.8}_{\scriptscriptstyle +0.7}$ \\
\midrule
\verb|ResNet-50| & ${79.6}$ & $\bm{81.7}_{\scriptscriptstyle +2.1}$ & ${81.4}_{\scriptscriptstyle +1.7}$ & $\bm{81.5}_{\scriptscriptstyle +1.8}$ & ${81.1}_{\scriptscriptstyle +1.5}$ & ${81.0}_{\scriptscriptstyle +1.4}$ & ${81.1}_{\scriptscriptstyle +1.4}$ & ${79.3}_{\scriptscriptstyle -0.3}$ & ${79.6}_{\scriptscriptstyle -0.1}$ \\
\midrule
\verb|Swin-Tiny| & ${80.9}$ & $\bm{83.8}_{\scriptscriptstyle +2.8}$ & $\bm{83.7}_{\scriptscriptstyle +2.8}$ & $\bm{83.8}_{\scriptscriptstyle +2.8}$ & ${83.5}_{\scriptscriptstyle +2.5}$ & $\bm{83.6}_{\scriptscriptstyle +2.6}$ & $\bm{83.7}_{\scriptscriptstyle +2.7}$ & ${81.3}_{\scriptscriptstyle +0.3}$ & ${81.7}_{\scriptscriptstyle +0.8}$ \\
\bottomrule[1.5pt]
\end{tabular}

        }
        \caption{1000 epochs}
        \label{tab:imagenet_plus_clip_vit_mixed_e1000}
    \end{subtable}
    \caption{\textbf{CLIP, ViT, Mixed architecture teachers.} Subscripts show 
    the improvement on top of the \IN{} accuracy.  We highlight the best 
    accuracy on each row from our proposed datasets and any number that is 
    within $0.2$ of the best.
    }
    \label{tab:imagenet_plus_clip_vit_mixed}
\end{table*}

\end{document}